%% file: main.tex
\documentclass[letterpaper]{article} 

\input{preamble}

\title{Neural Representations Reveal Distinct Modes of Class Fitting in Residual Convolutional Networks}
\author {
    Michał Jamroż,\textsuperscript{\rm 1}
    Marcin Kurdziel\textsuperscript{\rm 1}
}
\affiliations {
    \textsuperscript{\rm 1} AGH University of Science and Technology\\
    Krakow, Poland\\
    mijamroz@agh.edu.pl, kurdziel@agh.edu.pl
}

\begin{document}

\maketitle

\begin{abstract}

We leverage probabilistic models of neural representations to investigate how residual networks fit classes. To this
end, we estimate class-conditional density models for representations learned by deep ResNets. We then use these models
to characterize distributions of representations across learned classes. Surprisingly, we find that classes in the
investigated models are not fitted in an uniform way. On the contrary: we uncover two groups of classes that are fitted
with markedly different distributions of representations. These distinct modes of class-fitting are evident only in the
deeper layers of the investigated models, indicating that they are not related to low-level image features. We show that
the uncovered structure in neural representations correlate with memorization of training examples and adversarial
robustness. Finally, we compare class-conditional distributions of neural representations between memorized and typical
examples. This allows us to uncover where in the network structure class labels arise for memorized and standard
inputs.

\end{abstract}

\section{Introduction} \label{sec:introduction}

\input{introduction}

\section{Probabilistic model for class representations} \label{sec:DPGMM}

\input{dpgmm}

\section{Memorization in neural networks} \label{sec:memorized_examples}

\input{memorized_examples}

\section{Experimental setup} \label{sec:experimental_setup}

\input{experimental_setup}

\section{How residual networks fit classes} \label{sec:stability}

\input{class_fitting}

\section{Class representations correlate with memorization and adversarial robustness}
\label{sec:memorization_robustness}

\input{memorization_robustness}

\section{Where neural network fit classes} \label{sec:class_divergences}

\input{class_divergences}

\section{Related work} \label{sec:related_work}

\input{related_work}

\section{Conclusions} \label{sec:conclusions}

\input{conclusions}

\section*{Acknowledgments}
Research presented in this paper was supported by funds assigned to AGH University of Science and Technology by the
Polish Ministry of Education and Science. This research was supported in part by PL-Grid Infrastructure.

\input{main.bbl}
\input{technical_appendix}

\end{document}

%% file: preamble.tex
\usepackage{aaai23}  
\usepackage{times}  
\usepackage{helvet}  
\usepackage{courier}  
\usepackage[hyphens]{url}  
\usepackage{graphicx} 
\usepackage{natbib}  
\usepackage{caption} 
\usepackage{algorithm}
\usepackage{algorithmic}

\usepackage{newfloat}
\usepackage{listings}
\DeclareCaptionStyle{ruled}{labelfont=normalfont,labelsep=colon,strut=off} 
\floatstyle{ruled}
\newfloat{listing}{tb}{lst}{}
\floatname{listing}{Listing}

\usepackage{booktabs}
\usepackage{amsfonts}
\usepackage{amssymb}
\usepackage{amsmath}
\usepackage{nicefrac}
\usepackage{microtype}
\usepackage{xcolor}

\usepackage[page]{appendix}
\usepackage{array}
\usepackage{bm}
\usepackage{ltablex}
\usepackage{makecell}
\usepackage{rotating}
\usepackage{subfigure}
\usepackage{tabularx}

\newcommand{\dkl}[2]{D_\mathrm{KL}\left(#1\; \left| \right| \; #2\right)}

\newcommand{\ds}{\mathcal{D}}

\newcommand{\dsass}{\bm{c}}

\DeclareMathOperator*{\E}{\mathbb{E}}
\DeclareMathOperator*{\argmax}{arg\,max}

%% file: introduction.tex
Neural networks are vastly over-parametrized models. This flexible parameter space makes it quite difficult to answer
basic questions about internal representations learned by these models. For a long time it was not even clear whether
two identical networks trained on the same task learn similar sets of representations. That said, the picture begin to
change in recent years with the introduction of new tools to analyse internal representations in neural networks.
First,~\citet{Raghu2017} and \citet{Morcos2018} proposed canonical correlation analysis-based similarity measures for
neural representations. Subsequently,~\citet{Kornblith2019} proposed a kernel-based similarity score and found
significant representational similarity between networks trained to solve closely related tasks. Machinery for analysis
of neural representations was then extended by~\citet{Jamroz2020}, who proposed a non-parametric density model for
features learned by neural networks. They used this model to show that memorizing networks learn more complex 
representations than networks that can exploit patterns~in~data.

So far neural representations were studied from the perspective of features learned by network layers. In this work we
extend this line of research to investigate \emph{class representations} learned by neural networks. To this end, we
estimate class-conditional distributions of neural representations and use them as proxies to the representations of
classes learned by the network. This allows us to make several important contributions. First, we find that ResNets do
not fit classes in an uniform way. On the contrary, we observe two distinct modes of class fitting evident in the deeper
layers of these networks. We characterize class-conditional distributions of neural representations in the two groups of
classes and propose a likely explanation for the observed differences. Next, we demonstrate that the uncovered structure
in class representations translates to observable differences in memorization of input examples. To this end, we leverage
tractable memorization measures recently proposed by~\citet{Feldman2020b}. We also demonstrate that the two groups of
classes differ in adversarial robustness. Finally, we leverage class-conditional distributions of neural representations
to uncover where in the network structure classes are fit and to compare this process for memorized and typical inputs.

While this work focuses on convolutional residual networks, in the appendix we also report initial results for
non-convolutional models with skip connections, namely Vision Transformers~\citep{Dosovitskiy2021} and
MLP-Mixers~\citep{Tolstikhin2021}. We provide the source code for the experiments, as well as our density and
memorization estimates.\footnote{\url{https://github.com/mjamroz90/dnn-class-fitting}}

%% file: dpgmm.tex
Our goal in this work is to characterize representations of classes in neural networks. While the class
membership information is typically available for a large set of observations, namely the training set, combining it
into representations of classes is a non-trivial task. The main hurdle here comes from the stochastic nature of neural
representations. Specifically, a neural representation inferred for some input $\mathbf{x}$ can be seen as an outcome of
sampling $\mathbf{x}$ from the data distribution. Clearly, a reasonable notion of a \emph{class representation} should
capture the outcome of this sampling. We therefore propose to use class-conditional distributions of inputs'
representations as proxies to neural representations of classes. Concretely, we fit tractable density models to sets of
neural representations and then use these models to characterize distributions of representations in classes.

We capture the distributions of neural representations with the hierarchical Bayesian model that was recently used
by~\citet{Jamroz2020} to investigate networks that fit random labels. \citeauthor{Jamroz2020} used this model to
characterize representations of kernels in convolutional layers. A neural representation in their work was therefore
defined as a vector of kernel responses over a fixed sequence of inputs, averaged across spatial dimensions. Our goal,
however, is not to investigate features learned by convolutional kernels, but to characterize---in distributional
settings---representations of inputs from the learned classes. We therefore adopt a different construction for neural
representations. Specifically, in modern variable-resolution convolutional models input to the classification head is
frequently constructed by global average pooling of the output feature maps. We follow a similar construction for
representations of inputs in our class-conditional density models. In particular, consider a network layer $l$ with a
sequence of convolutional kernels: $k^l_1, k^l_2, \ldots, k^l_d$---where $d$ is the layer width---and let
$\mathbf{x} \in C$ be an input observation from class $C$. We construct the representation of $\mathbf{x}$ at layer $l$
as the vector of respective kernel activations averaged across spatial dimensions:
\begin{equation} \label{eq:input_representation}
  \begin{split}
    nn_l\left(\mathbf{x}\right) =
      & \left[ \mathrm{avg\_pool}(k^l_1(\mathbf{x})), \mathrm{avg\_pool}(k^l_2(\mathbf{x})), \ldots, \right. \\
      & \ \left. \mathrm{avg\_pool}(k^l_d(\mathbf{x})) \right].
  \end{split}
\end{equation}
Activation vectors for all inputs $\mathbf{x} \in C$ collectively form a set of observations that is then
explained with the density model. Because all input observations are sampled from $C$, the model estimates the
class-conditional distribution for $C$.

\subsection{Density model}

The hierarchical Bayesian model used by \citet{Jamroz2020} explains a set of observations $\mathbf{x} \in \ds$---e.g. a
set of neural representations of some network inputs---with a mixture of multivariate normal distributions with an
unknown number of components:
\begin{equation} \label{eq:dpgmm}
  \begin{split}
    \alpha        & \sim Gamma(1,1), \\
    G \mid \alpha & \sim DP(NIW(\bm{\theta}_0), \alpha), \\
    \bm{\mu}_k, \bm{\Sigma}_k & \sim G, \\
    \bm{x} \mid \bm{\mu}_k, \bm{\Sigma}_k &\sim \mathcal{N}(\bm{\mu}_k, \bm{\Sigma}_k).
  \end{split}
\end{equation}
The prior over component parameters is constructed as a Dirichlet Process ($DP)$ with Normal-Inverse-Wishart~($NIW$)
base distribution. A mixture of this form is consistent in total variation for a large family of continuous
distributions~\cite{Ghosal2017} and admits an efficient \emph{collapsed Gibbs sampler} (CGS) for the posterior over
component parameters~\citep{NealDpMCMM}. Shortly, CGS constructs a Markov chain over assignments of observations to
components: $\left(\dsass_1,  \dsass_2,  \dsass_3, \ldots\right)$. Under the model in~Eq.~\eqref{eq:dpgmm}, the
conditional posterior predictive distribution~$p(\bm{x^*} \mid \ds, \dsass_t)$ over a new observation $\bm{x^*}$ 
given the set of assignments $\dsass_t$ has a closed-form solution~\citep{Jamroz2020}. This conditional
predictive distribution can be leveraged to construct a tractable Monte Carlo approximation to the full posterior
predictive distribution:
\begin{equation} \label{eq:mcmc_post_predictive}
  p(\bm{x^*} \mid \ds) \approx \frac{1}{n} \sum_{t=1}^n p(\bm{x^*} \mid \ds, \dsass_t).
\end{equation}
The average here is taken over the assignments sampled by CGS, i.e. over the Markov chain steps. 

The closed-form solution for the predictive distribution conditioned on component assignments~$p(\bm{x^*} \mid \ds, \dsass_t)$
gives a simple sampling strategy for the full posterior predictive: sample a component assignment $\dsass_t$ from the
Markov chain and then sample $\bm{x^*}$ from the induced conditional distribution~$p(\bm{x^*} \mid \ds, \dsass_t)$. This
sampling scheme can be used to compare the posterior predictive distribution with another
distribution~$q\left(\bm{x^*} \right)$ that has a tractable density. Specifically, the samples can be used to
construct a Monte Carlo approximation to the relative entropy---or Kullback-Leibler (KL)
divergence---from~$q \left(\bm{x^*} \right)$ to~$p(\bm{x^*} \mid \ds)$:
\begin{equation} \label{eq:mcmc_dkl}
  \begin{split}
  & \dkl{p\left( \bm{x^*} \mid \ds \right)}{q\left( \bm{x^*} \right)} = \\
  & \quad = \E_{\bm{x^*} \sim p\left( \bm{x^*} \mid \ds \right)}
            \left[ \log p\left( \bm{x^*} \mid \ds \right) -  \log q\left( \bm{x^*} \right) \right] \\
  & \quad \approx \frac{1}{nm} \sum_{t=1}^n\sum_{s=1}^m
            \left[ \log p\left(\bm{x^*}_{st} \mid \ds, \dsass_t\right) - \log q\left(\bm{x^*}_{st}\right) \right],
  \end{split}
\end{equation}
where $\bm{x^*}_{st}$ are sampled from $p\left( \bm{x^*} \mid \ds, \dsass_t \right)$. As proposed by
\citeauthor{Jamroz2020}, the estimate in~Eq.~\eqref{eq:mcmc_dkl} can be used to quantify the complexity
of neural representations. In this case the reference distribution~$q\left( \bm{x^*} \right)$ is chosen to be the
maximum entropy distribution that explains only the location and the scale of $\ds$, i.e. the diagonal Gaussian with the
mean and the variances estimated from~$\ds$. 

In this work we use the divergence from the maximum entropy distribution to characterize the complexity of
class-conditional distributions of neural representations. We also use Eq.~\eqref{eq:mcmc_dkl} to compare posterior
predictive distributions estimated for different classes. Specifically, let~$p\left(\bm{x^*} \mid \mathcal{F} \right)$
be the posterior predictive distribution under model in Eq.~\eqref{eq:dpgmm} induced by a set of neural
representations~$\mathcal{F}$. Because the density in this posterior predictive is tractable
(Eq.~\eqref{eq:mcmc_post_predictive}), we can use Eq.~\eqref{eq:mcmc_dkl} to estimate the KL divergence from
$p\left(\bm{x^*} \mid \mathcal{F} \right)$ to~$p(\bm{x^*} \mid \ds)$. To this end, we simply set the reference
distribution~$q\left( \bm{x^*} \right)$ to be the posterior predictive $p\left(\bm{x^*} \mid \mathcal{F} \right)$.

Standard CGS exhibits relatively slow mixing. To speed-up the estimation of predictive distributions, in this work we
sample posterior parameters using a variant of this algorithm with blocked Gibbs sampling~\citep{BlockCGSJensen}. See
Appendix for more details.

%% file: memorized_examples.tex
Recently,~\citet{Feldman2020a} proposed a theoretical framework that explains the role of input memorization when
fitting a long-tailed data distribution, i.e. a population of input examples characterized by many infrequent
components. They consider a case where some of the examples in an $n$-element train set come from sub-populations with
frequency below~$1/n$. Such examples are statistically indistinguishable from noise or mislabelled data. Consequently,
their fitting requires memorization. Importantly, \citeauthor{Feldman2020a} showed that memorization in these settings
is in fact necessary to minimize the generalization error. They also proposed a measure for the degree of input
memorization.  Specifically, given a learning algorithm~$\mathcal{A}$ and an example $\left(\mathbf{x}, t \right)$,
\citeauthor{Feldman2020a} defines the degree of memorization of $\left(\mathbf{x}, t\right)$ as the probability of
predicting $t$ for input $\mathbf{x}$ when~\mbox{$\left(\mathbf{x}, t\right)$} is observed during training, relative to
the probability of the correct prediction when $\left(\mathbf{x}, t \right)$ is absent from the training data. A score
of $1.0$ therefore corresponds to an example that is memorized in every training run, whereas a score of $0.0$
corresponds to an example with no memorization. This definition assumes that $\mathcal{A}$ is a stochastic learning
algorithm, for example stochastic gradient descent. In a follow-up work \citet{Feldman2020b} proposed a
computationally-efficient approximation to this memorization score and used it to estimated memorization in two popular
image classification benchmarks, namely CIFAR-100 and ImageNet. They found that in both cases convolutional neural
networks memorize a non-trivial part of the train set.

\citeauthor{Feldman2020b} work opens an avenue for investigating mechanisms behind memorization in deep networks. In
particular, previous research focused on models where memorization of specific examples was induced artificially, e.g.
by corrupting the labels or the network input~\citep{Zhang2017, Arpit2017}. Memorization estimates can, however, be used
to investigate memorization in neural networks that learn from uncorrupted data. We use this opportunity to uncover the
relationship between representations of classes in neural networks and the degree of input memorization. We also
leverage class-conditional density models of neural representations to investigate where in the network structure
classes arise for typical and memorized examples.

%% file: experimental_setup.tex
We conduct our experiments on two popular image datasets: CIFAR-100~\citep{CIFAR} and
Mini-ImageNet~\citep{MiniImageNet}. For CIFAR-100 we use the memorization scores published
by~\citet{Feldman2020b}.\footnote{Available at: \url{https://pluskid.github.io/influence-memorization}} To avoid
potential discrepancies between these estimates and our experimental conditions, we replicate important aspects of their
training setup. Concretely, we investigate representations learned by a ResNet50 network that was trained by closely
following hyper-parameters reported in~\citep{Feldman2020b}. To confirm our findings, we also repeat these experiments
with another residual convolutional network, namely ResNet18. In this case we estimated the necessary memorization
scores by following the Algorithm~$1$ in~\citep{Feldman2020b}. To this end, we trained $2000$ network instances
(\emph{trials} in Algorithm~1 therein) in order to estimate the required probabilities of correct prediction. For the
Mini-ImageNet experiments we collected images from the provided training, validation and test subsets, and then split
them randomly into $50,000$ training and~$10,000$ test examples. We then estimated memorization scores for the training
images. These experiments were also carried out on the ResNet50 and ResNet18 architectures. However, to lessen the
computational burden associated with estimation of memorization, we modified the original architectures by introducing a
3-pixel stride in the first convolutional layer. All other experimental details and training hyper-parameters match
those used with the CIFAR-100 dataset. We provide memorization scores estimated for the CIFAR-100 and Mini-ImageNet
datasets alongside this paper. Finally, it is worth noting that~\citeauthor{Feldman2020b} published memorization scores
also for the ImageNet dataset. However, due to the cost of estimating class-conditional density models and related
quantities on the ImageNet scale, we ultimately decided to focus on the Mini-ImageNet dataset.

ResNet architecture has four \emph{stages} that correspond to progressively smaller spatial dimensions. We extract
neural activations immediately after the summation operation at the end of each stage (See Appendix for details). In
each case we collect activations for the entire train set and use them to calculate neural
representations~(Eq.~\ref{eq:input_representation}). This gives us four sets of neural representations spaced evenly
across the network depth. Importantly, our main findings come from the analysis of the representations in the last
stage, i.e. the input to the classification head. We use representations from the intermediate stages only to
show how class representations form across the network depth. While in principle we could analyse representations
from each residual block, this would have no bearing on the our main findings and would significantly increase the
computational cost of experiments.

To obtain representations of classes we estimate the mixture model in Eq.~\eqref{eq:dpgmm} independently for each
combination of the class label and the network stage. One difficulty in modelling ResNet representations comes from the
width of this network: neural representations have up to $2048$ dimensions. Estimating Gaussian mixture models in this
many dimensions is expensive. To avoid this issue, we reduce the dimensionality of the collected representations via
Singular Value Decomposition~(SVD). In each case we retain enough dimensions to preserve most of the variance in the
data set (See Appendix for details). It is worth noting that SVD preprocessing was used before in investigations of
neural representations, e.g. by \citet{Raghu2017} and by \citet{Jamroz2020}. Importantly, \citeauthor{Raghu2017} found
that intrinsic dimensionality of neural representations---in their case the number of SVD directions needed to match the
performance of a complete network---is much smaller than the number of neurons in the corresponding network layer, which
justify the SVD pre-processing step.

To estimate the posterior distributions in the density models~(Eq.~\eqref{eq:dpgmm}) we perform, in each case, $400$
block collapsed Gibbs sampler steps (i.e. passes over the dataset). We use a block size of $b=4$ observations for both
datasets. Afterwards, we discard initial~$320$ Gibbs steps and include every~4th of the remaining steps in the Monte
Carlo estimates~(Eqs.~\eqref{eq:mcmc_post_predictive} and~\eqref{eq:mcmc_dkl}). Following~\citet{Jamroz2020}, we put a
data-derived, weakly-informative prior on the parameters of Gaussian components. To this end, we adopt the values of
prior hyper-parameters ($\bm{\theta}_0$ in~Eq.~\eqref{eq:dpgmm}) used in that work.

%% file: class_fitting.tex
\begin{figure*}[t]
  \centering
  \begin{tabular}{cc}
    ResNet18 & ResNet50 \\
    \includegraphics[width=0.230\linewidth]{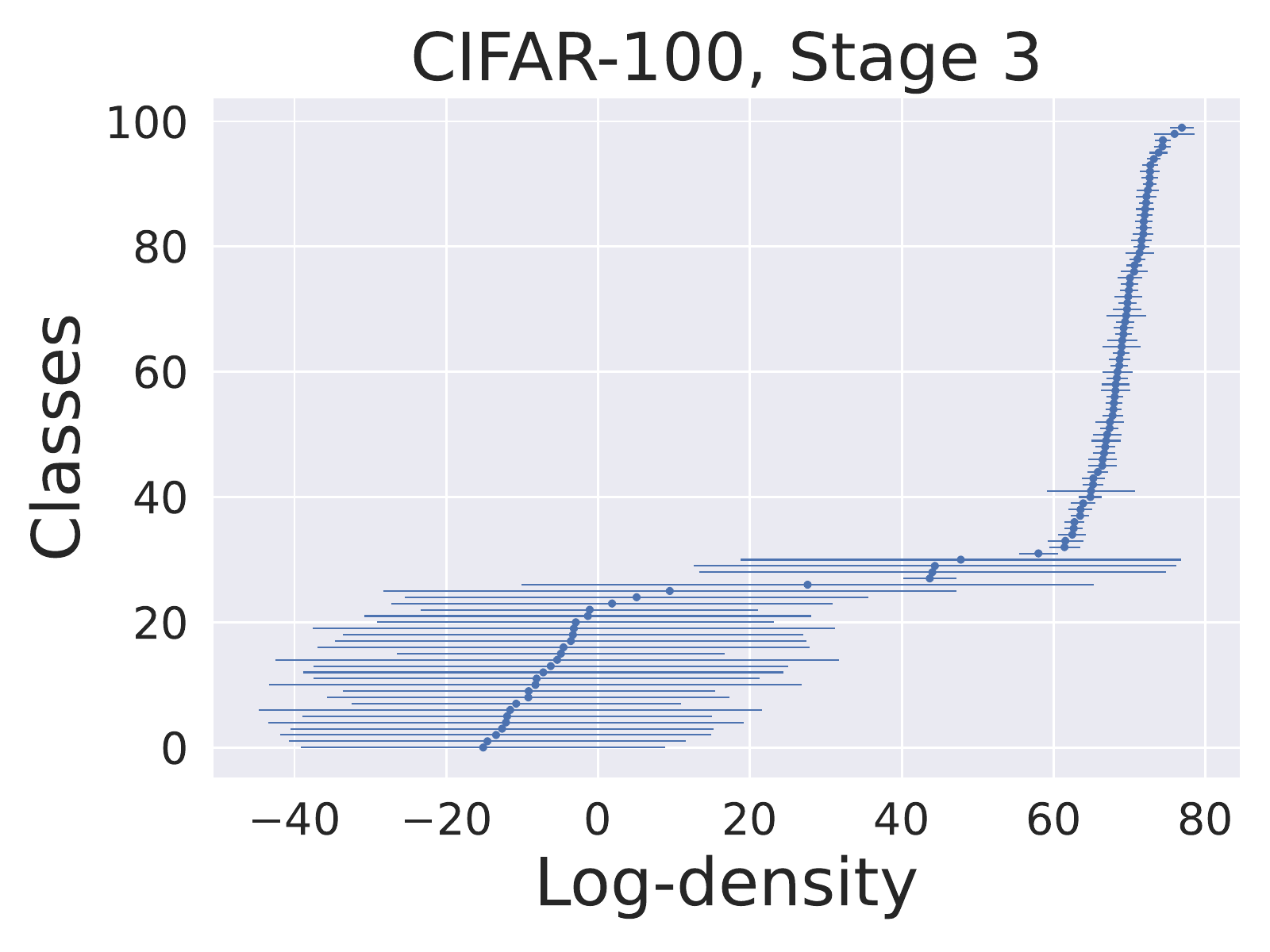}
    \includegraphics[width=0.230\linewidth]{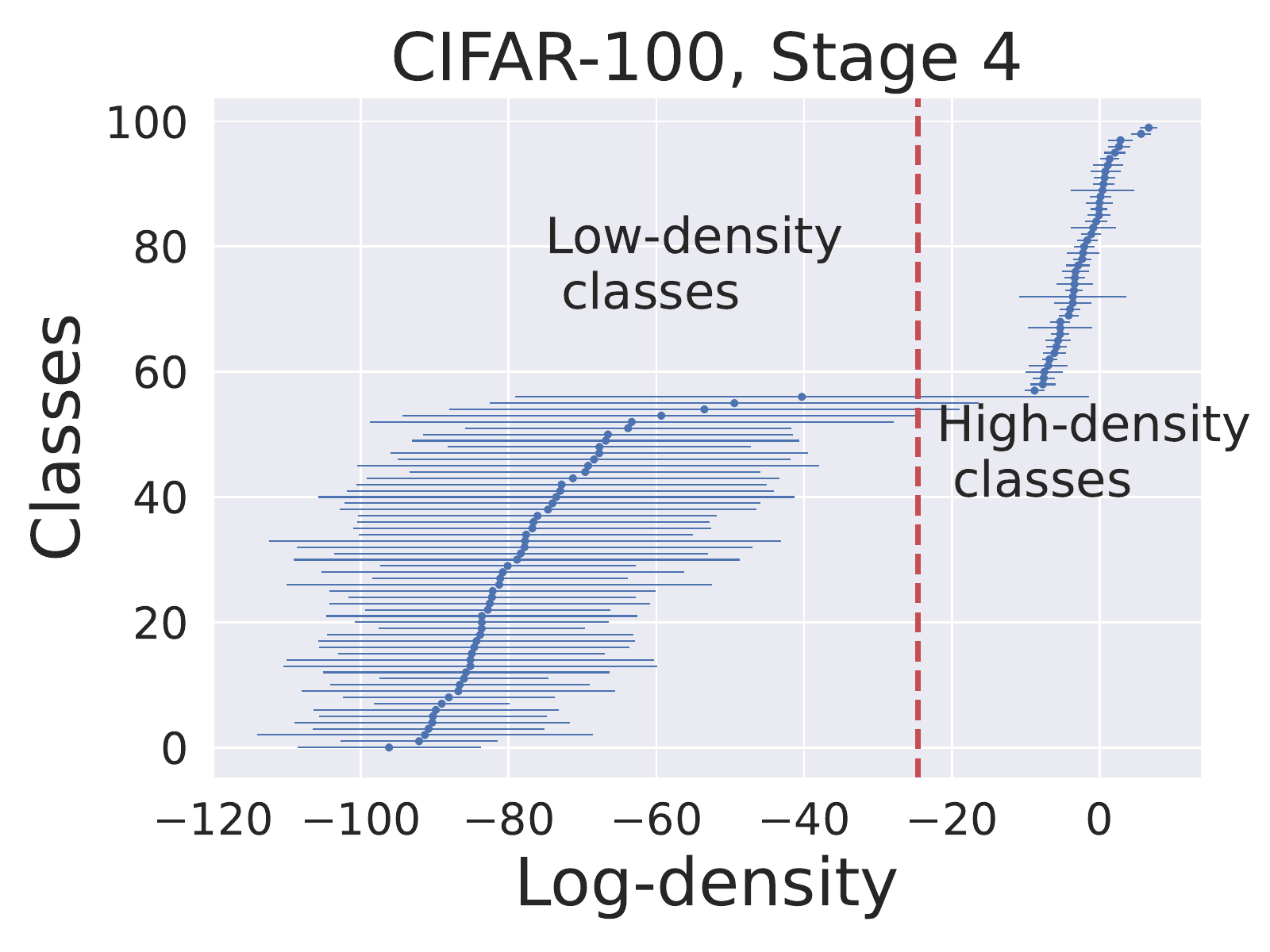}
    &
    \includegraphics[width=0.230\linewidth]{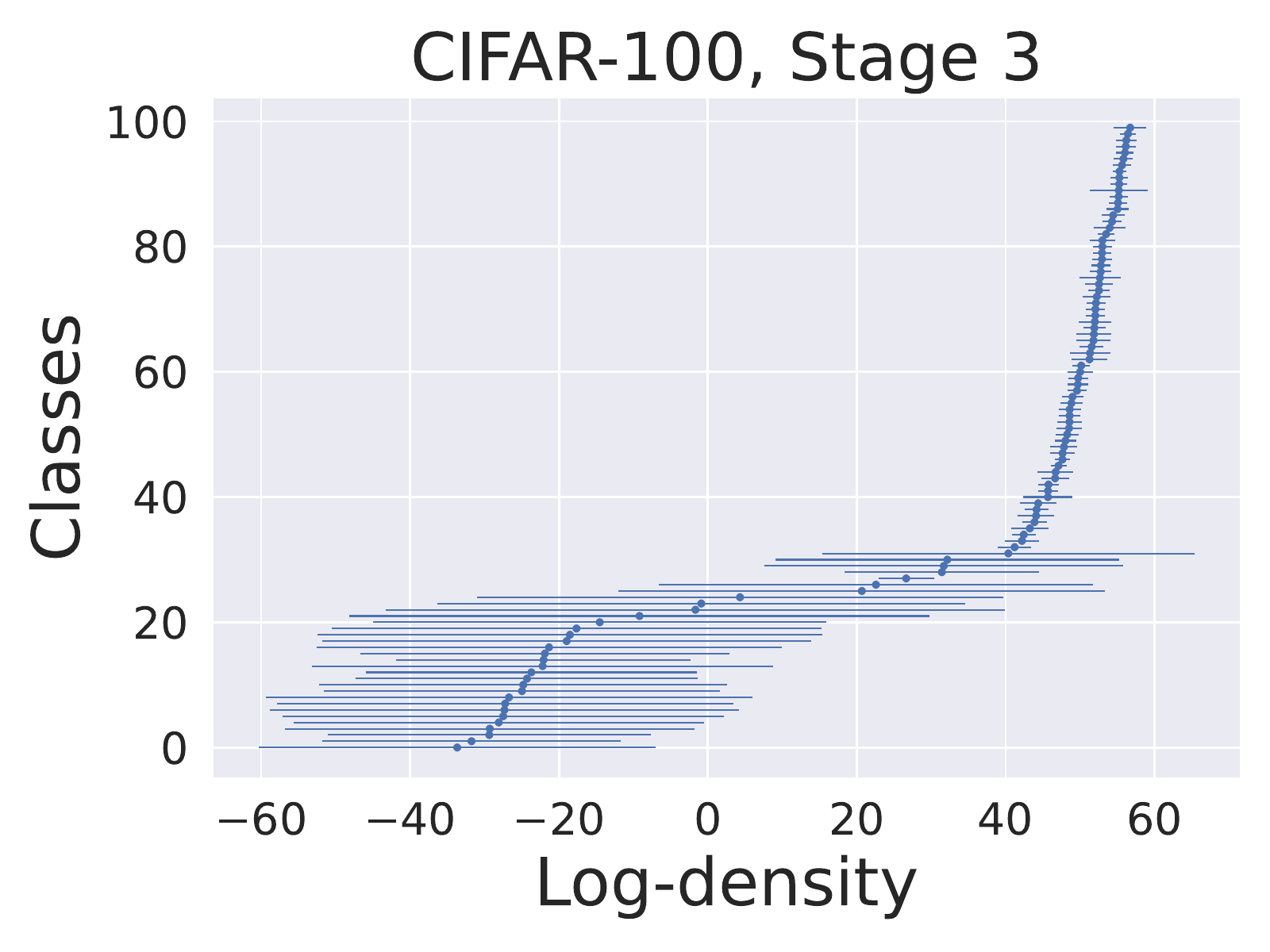}
    \includegraphics[width=0.230\linewidth]{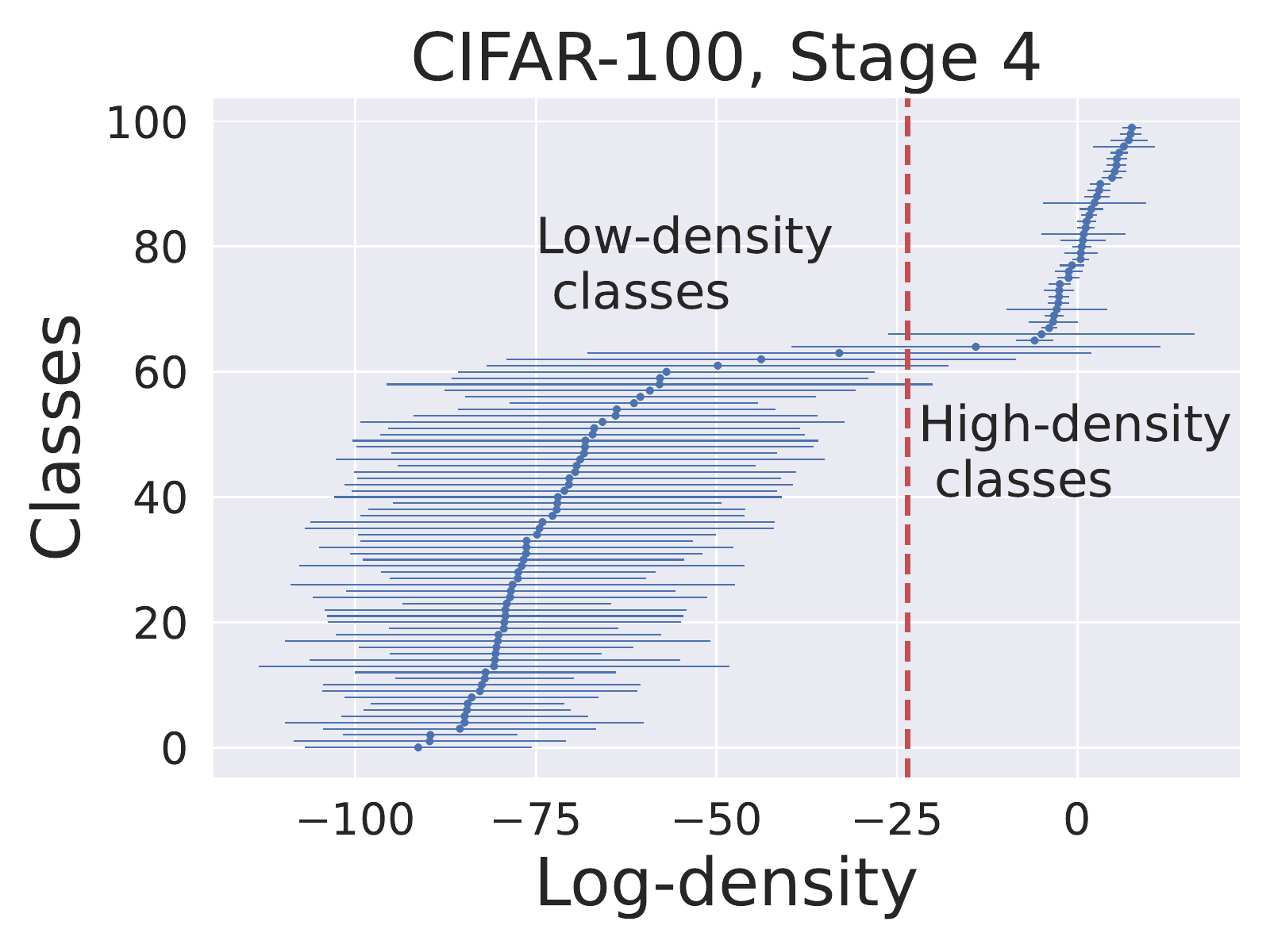}
    \\
    \includegraphics[width=0.230\linewidth]{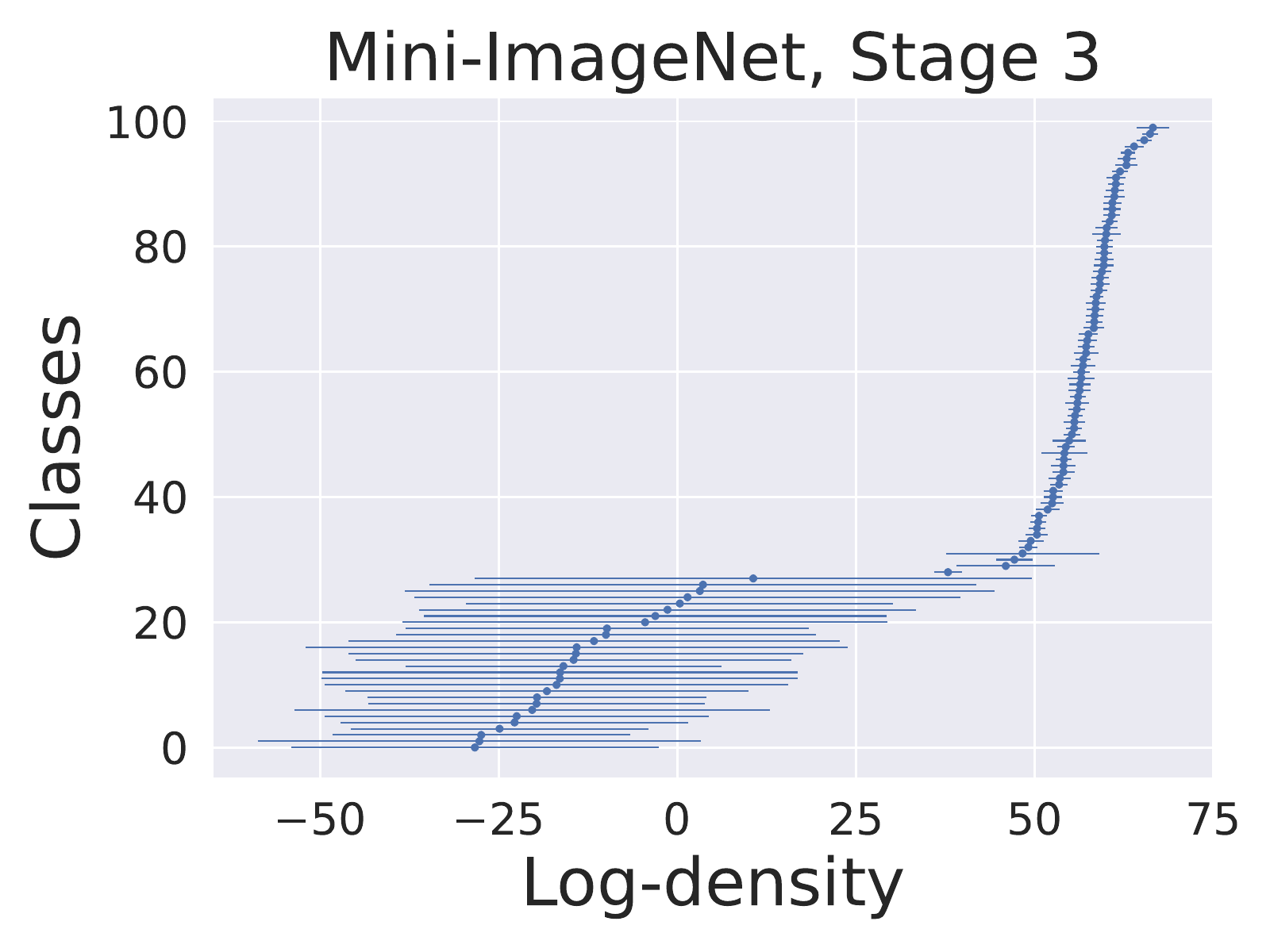}
    \includegraphics[width=0.230\linewidth]{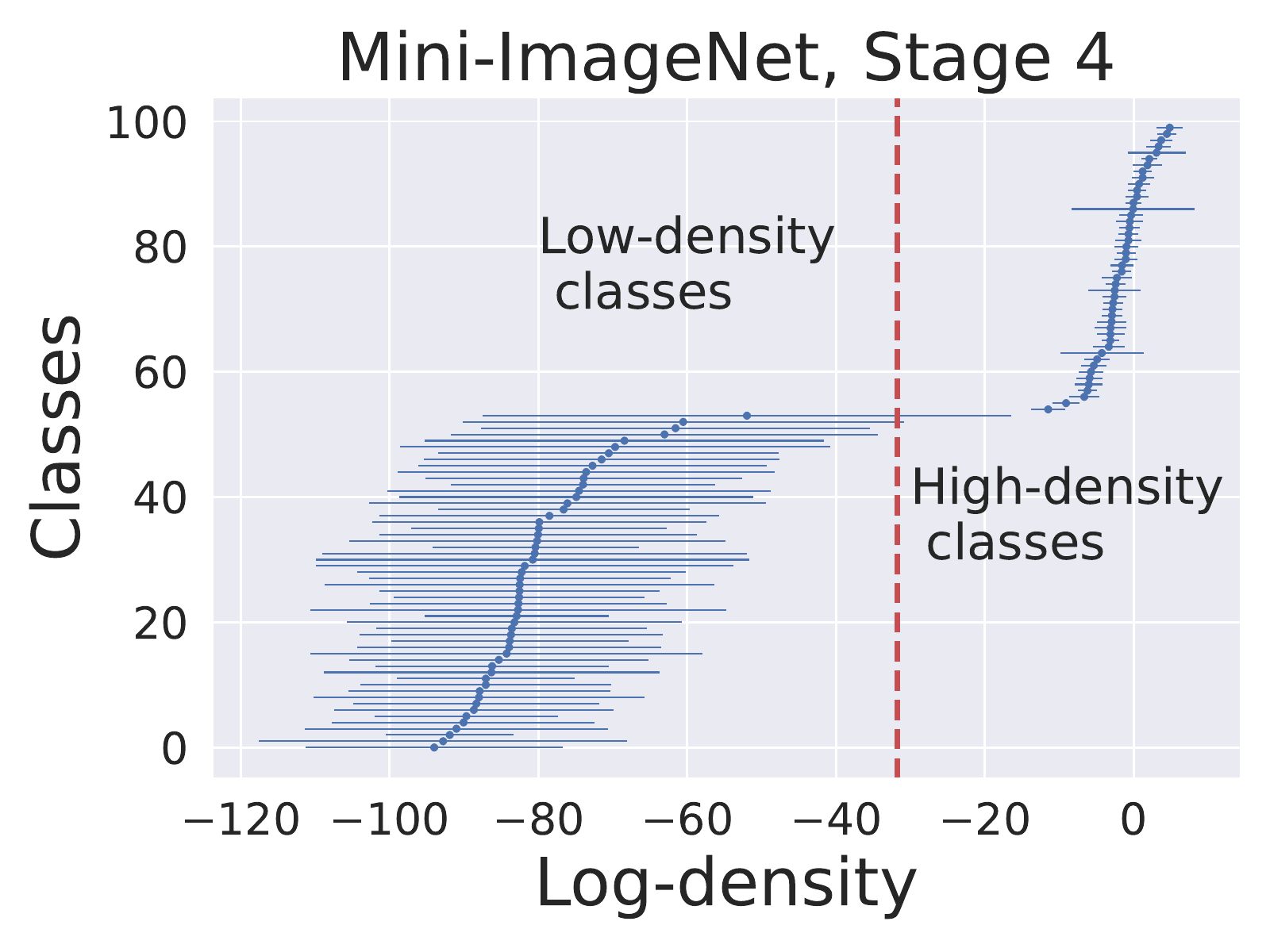}
    &
    \includegraphics[width=0.230\linewidth]{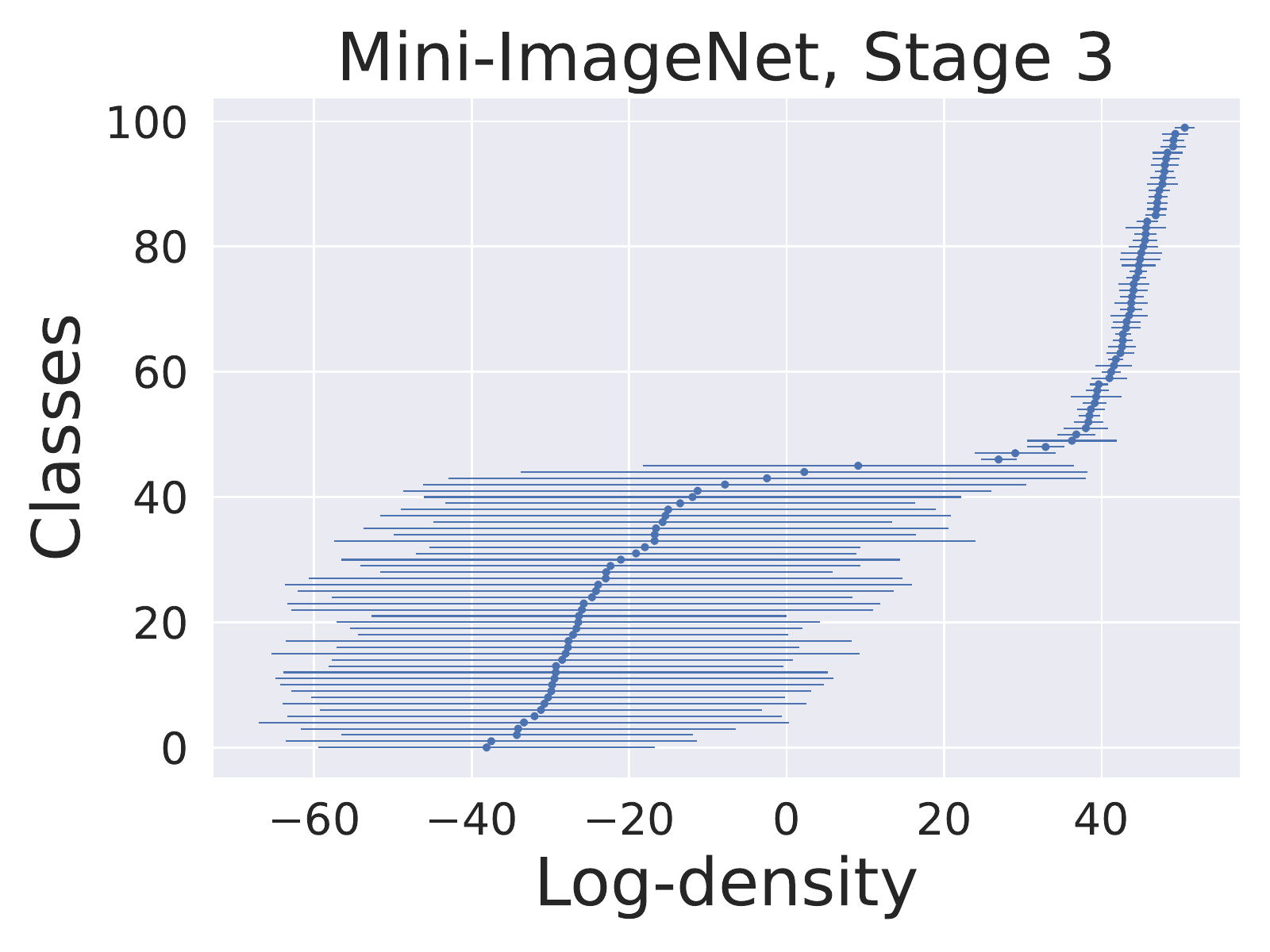}
    \includegraphics[width=0.230\linewidth]{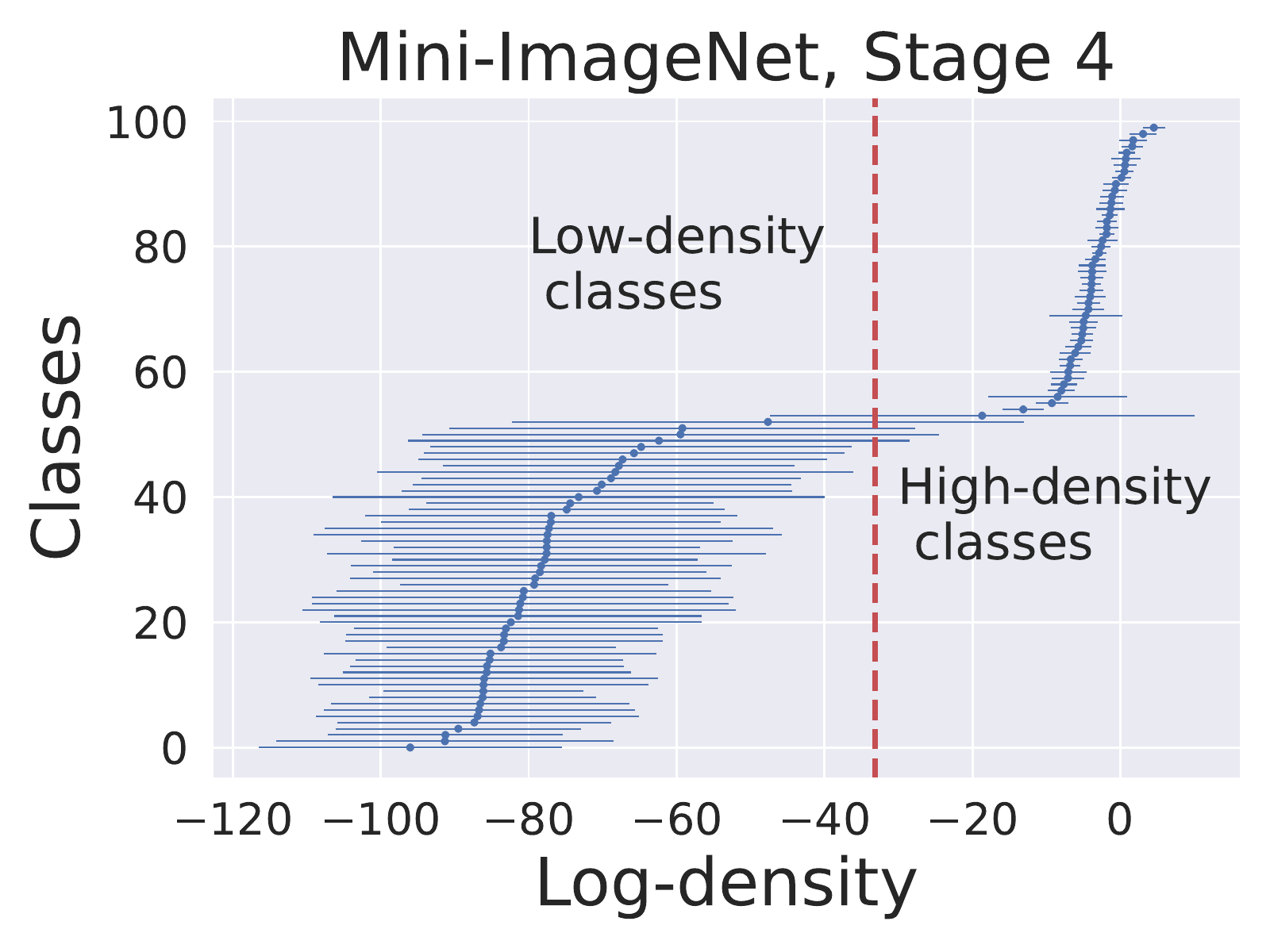}
  \end{tabular}
  \caption{Mean and standard deviation of class-conditional log-densities estimated for inputs from each class. Classes
           on the vertical axis are sorted according to the mean class-conditional density.}
  \label{fig:log_cc_densities}
\end{figure*}

We begin our analysis by characterizing representations of classes learned from the two image datasets. First, for each
class $C$ and each network stage we calculate mean and standard deviation of class-conditional
log-densities~(Eq.~\ref{eq:mcmc_post_predictive}) of examples~$\mathbf{x} \in C$. Results for the third and fourth stage
are reported in Fig.~\ref{fig:log_cc_densities}. Results for the first two stages are in the Appendix. Already, we see
an unexpected structure: class representations in the last two stages display an almost phase transition-like change in
class-conditional log-densities. That is, we observe two groups of classes that cluster around vastly different density
values. Equally striking, classes in these groups typically differ by an order of magnitude in the standard deviation
of the estimated class-conditional log-densities. Specifically, estimates for examples from the classes clustered around
the higher class-conditional density typically have an order of magnitude lower variance. These observations agree
between the two image datasets and the two architectures used in the experiment. Importantly, we did not found a similar
structure in the representations from the first two stages, indicating that the observed phenomenon is not related to
low-level features of the inputs (See Appendix). We also replicated this experiment using a plain (i.e. non-residual)
convolutional network. This time we did not observe distinct modes of class fitting in any of the network layers (See
Appendix). This implies that the observed structure is not simply a product of the datasets used in the experiments, but
is related to the residual architecture. In summary, we observe two distinct modes of class fitting in the last two
stages of ResNet models.

We now focus on the highest-level class representations, i.e. representations constructed by the last stage of the
network. First, we split classes into two groups according to the estimated mean log-density values, namely
\emph{low-} and \emph{high-density} classes (marked on Figure~\ref{fig:log_cc_densities}, stage 4). To this end, we sort
classes according to their mean log-densities and find interval $\left(C_i, C_{i+1}\right)$ in this sequence with the
largest mean log-density difference. We use center of this interval to assign classes to the two groups. Note that we
group entire classes, not individual input examples.

\begin{figure*}[!htb]
  \centering
  \begin{tabular}{cc}
    ResNet18 & ResNet50 \\
    \includegraphics[width=0.44\linewidth]{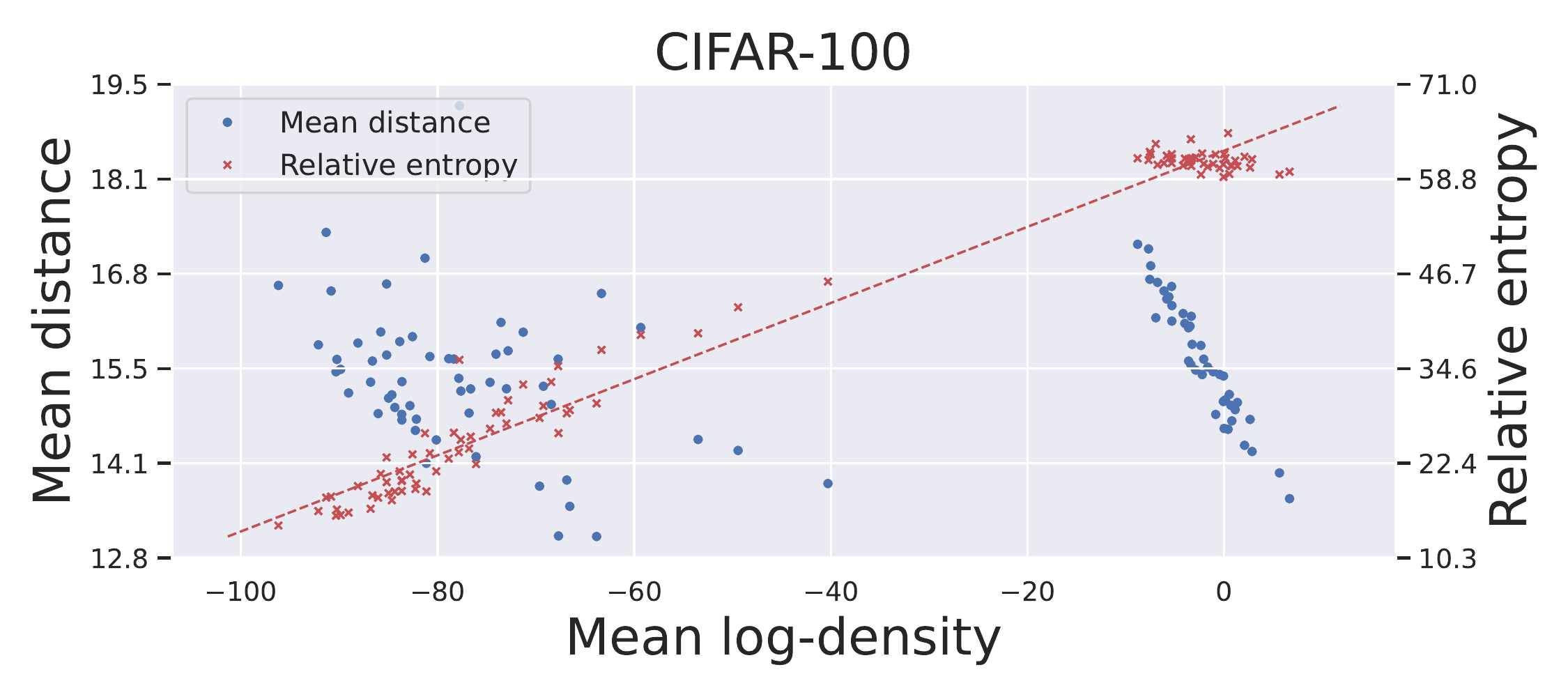} &
    \includegraphics[width=0.44\linewidth]{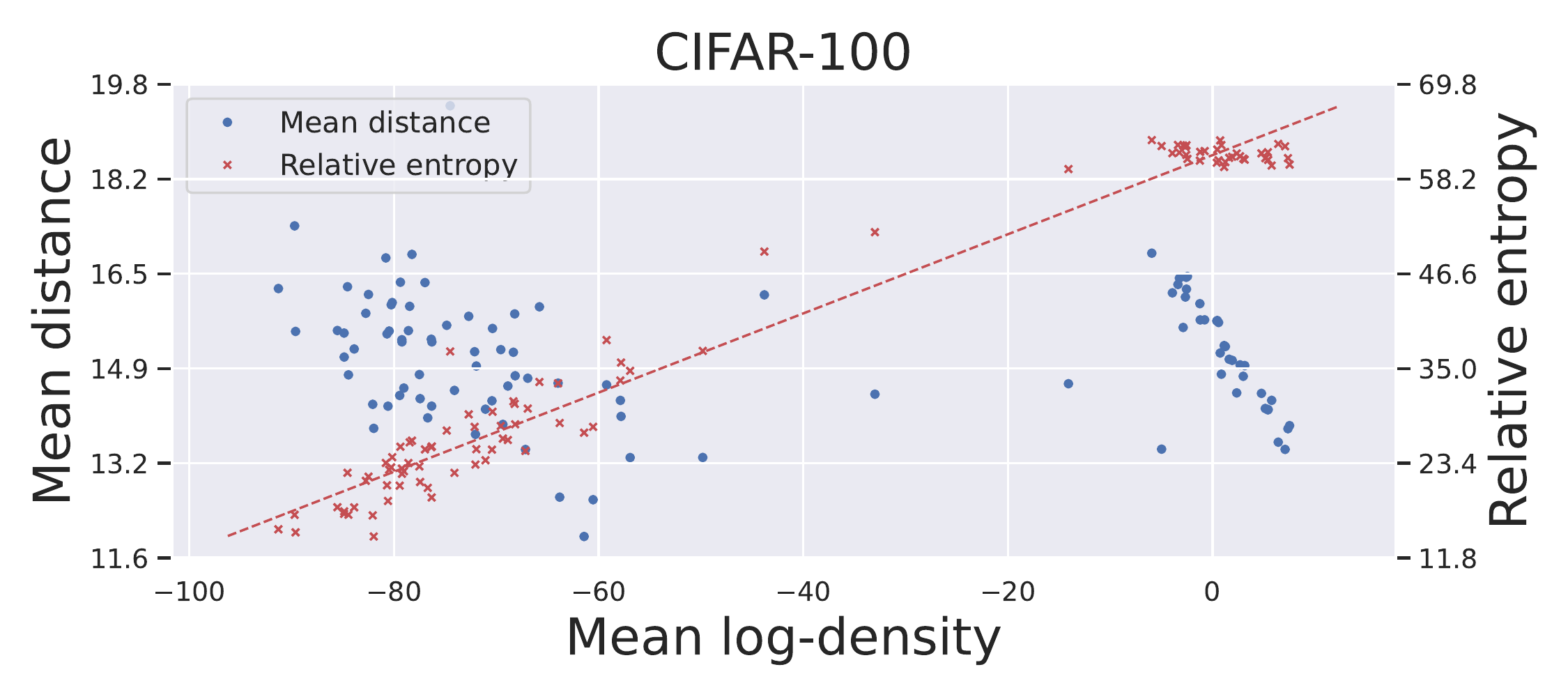} \\
    \includegraphics[width=0.44\linewidth]{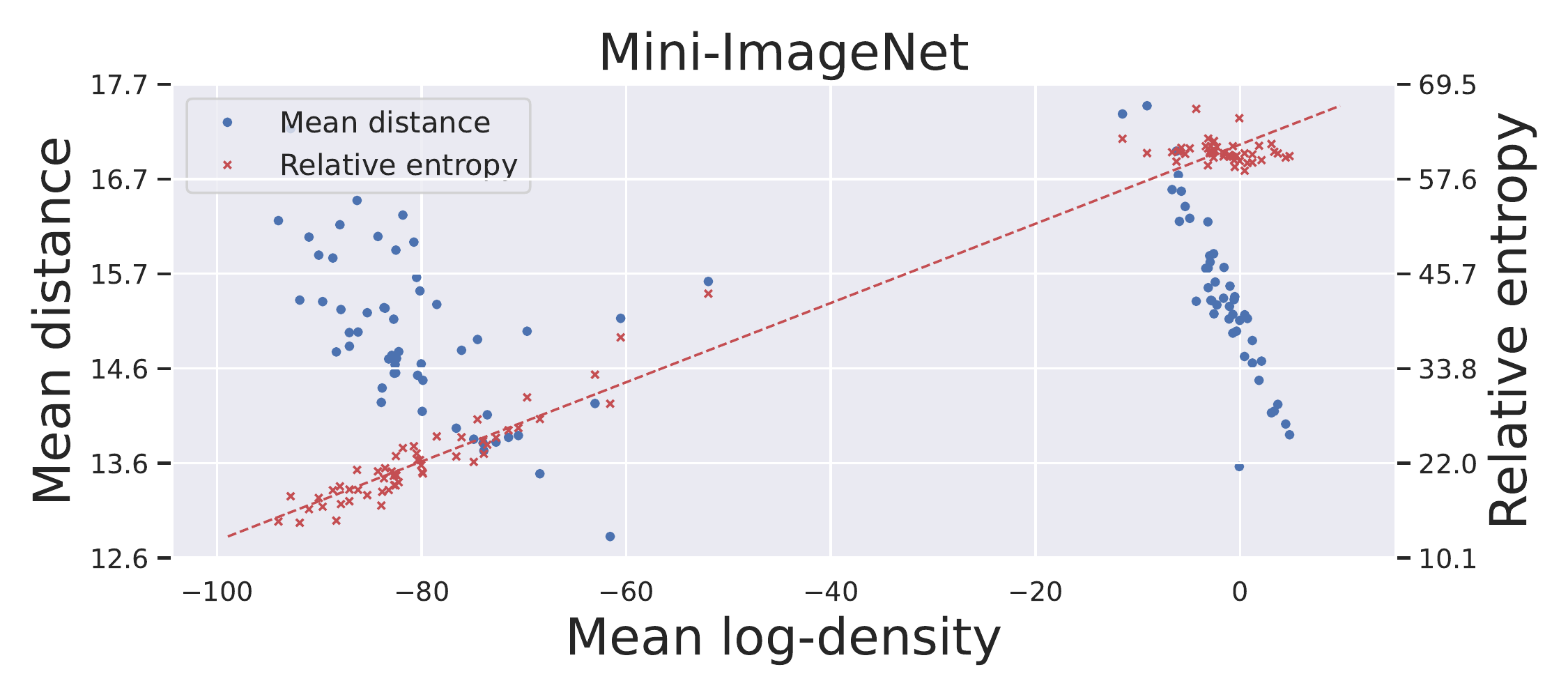} &
    \includegraphics[width=0.44\linewidth]{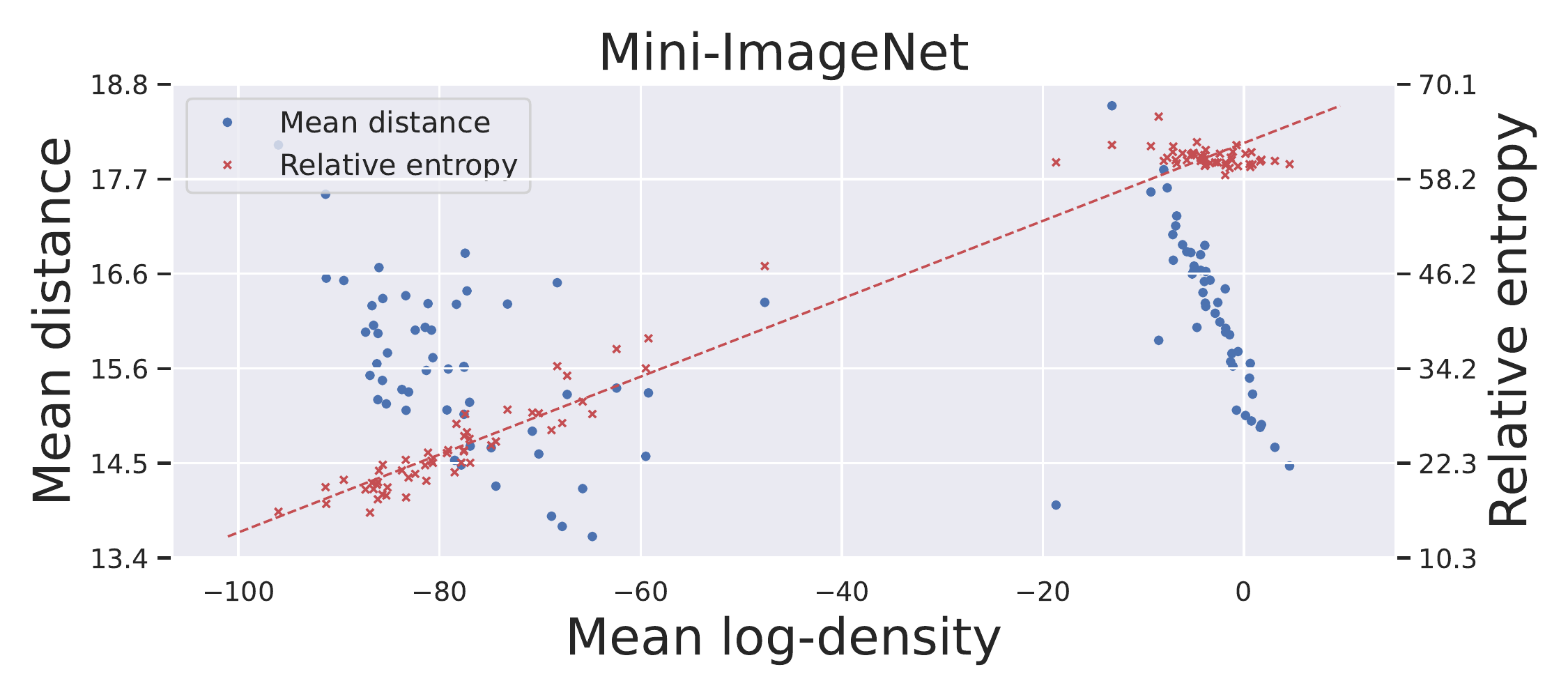}
  \end{tabular}
  \caption{Relative entropies of class-conditional distributions of neural representations and mean intra-class
           distances between representations. Each class corresponds to one blue and one red mark. Dashed line:
           least-squares fit to relative entropies.}
  \label{fig:entropy_vs_density}
\end{figure*}

The most straightforward explanation for the observed differences between the uncovered groups could be that the
high-density classes simply have more spatially compact distributions of neural representations. In other words,
examples from these classes could have more similar high-level neural representations. Our results show that this simple
explanation is incorrect. Specifically, for each class~$C$ we calculated the mean distance
$\|\mathrm{svd}\left(nn_l\left(\mathbf{x}\right)\right) - \mathrm{svd}\left(nn_l\left(\mathbf{y}\right)\right)\|$
between neural representations of examples $\mathbf{x}, \mathbf{y} \in C$ (Eq.~\eqref{eq:input_representation}, after
dimensionality reduction with SVD). We also estimated the complexity of its class representation, namely the relative
entropy of its posterior predictive distribution $p\left(\bm{x^*} \mid \bm{x^*} \in C\right)$ from the reference maximum
entropy distribution $q\left( \bm{x^*} \right)$ (Eq.~\eqref{eq:mcmc_dkl}, and the paragraph below).  Results are
reported in Fig.~\ref{fig:entropy_vs_density}. Clearly, the high-density classes are \emph{not} more spatially compact
at the neural representation level than the low-density classes. However, their posterior predictive distributions have
vastly larger complexity than the posterior predictive distributions of the low-density classes. In other words,
representations of the high-density classes are vastly more non-Gaussian. Together with the high average
class-conditional density of their examples, these results point to a different explanation: in representations of
high-density classes, the probability mass is concentrated in a number of compact but spatially separated modes. This
suggests that at the neural representation level high-density classes are formed by a collection of compact
components.\footnote{One could ask in this place: is this larger number of components observed in the posterior over
parameters of model~\eqref{eq:dpgmm}? We do observe larger number of components sampled by CGS for high-density classes.
However, we do not report these numbers, as---for technical reasons---estimation of component counts is fragile. See
Appendix for details.}

%% file: memorization_robustness.tex
\begin{figure*}[tb]
  \centering
  \begin{tabular}{cc}
    ResNet18 & ResNet50 \\
    \includegraphics[width=0.48\linewidth]{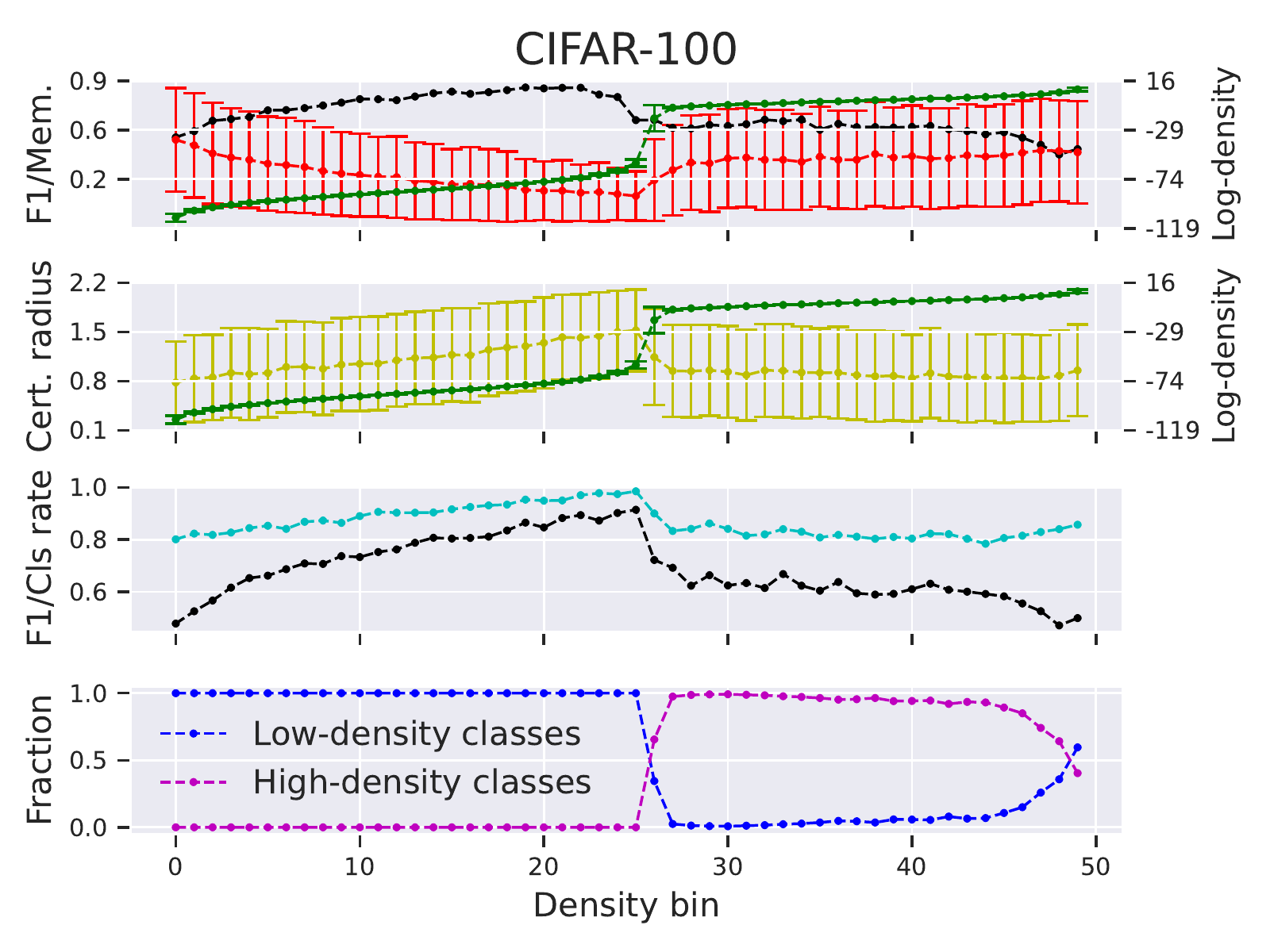} &
    \includegraphics[width=0.48\linewidth]{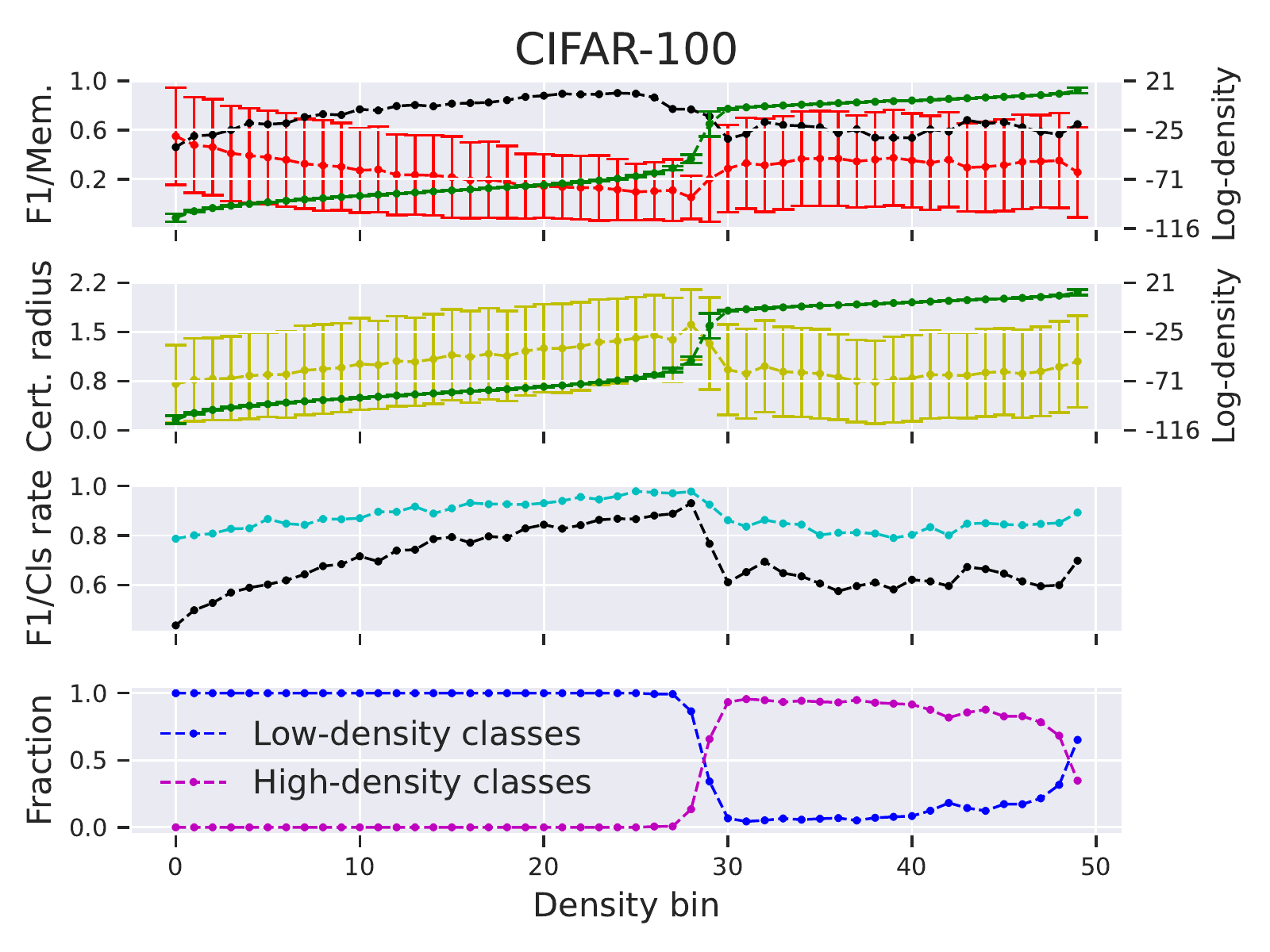} \\
    \includegraphics[width=0.48\linewidth]{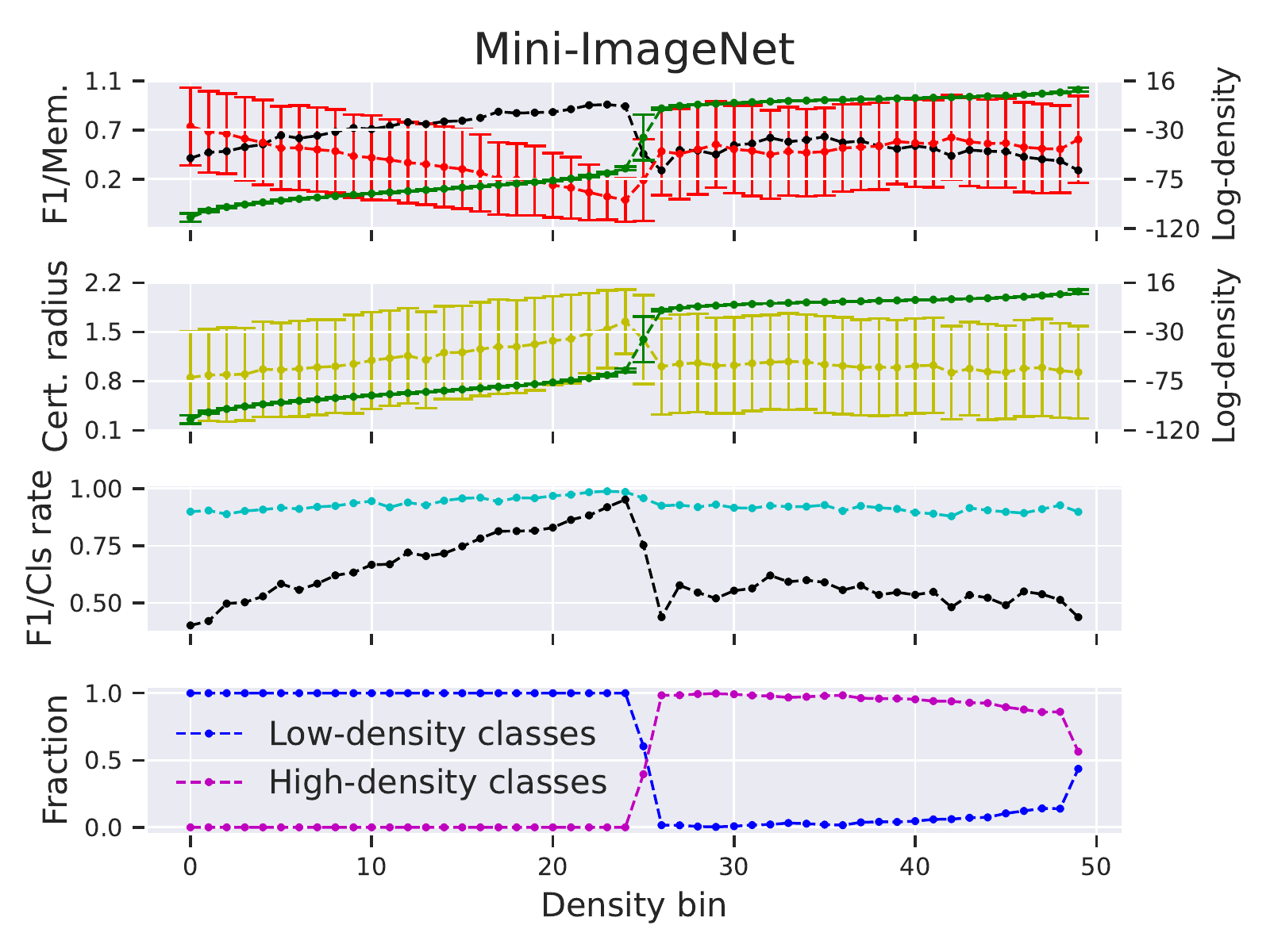} &
    \includegraphics[width=0.48\linewidth]{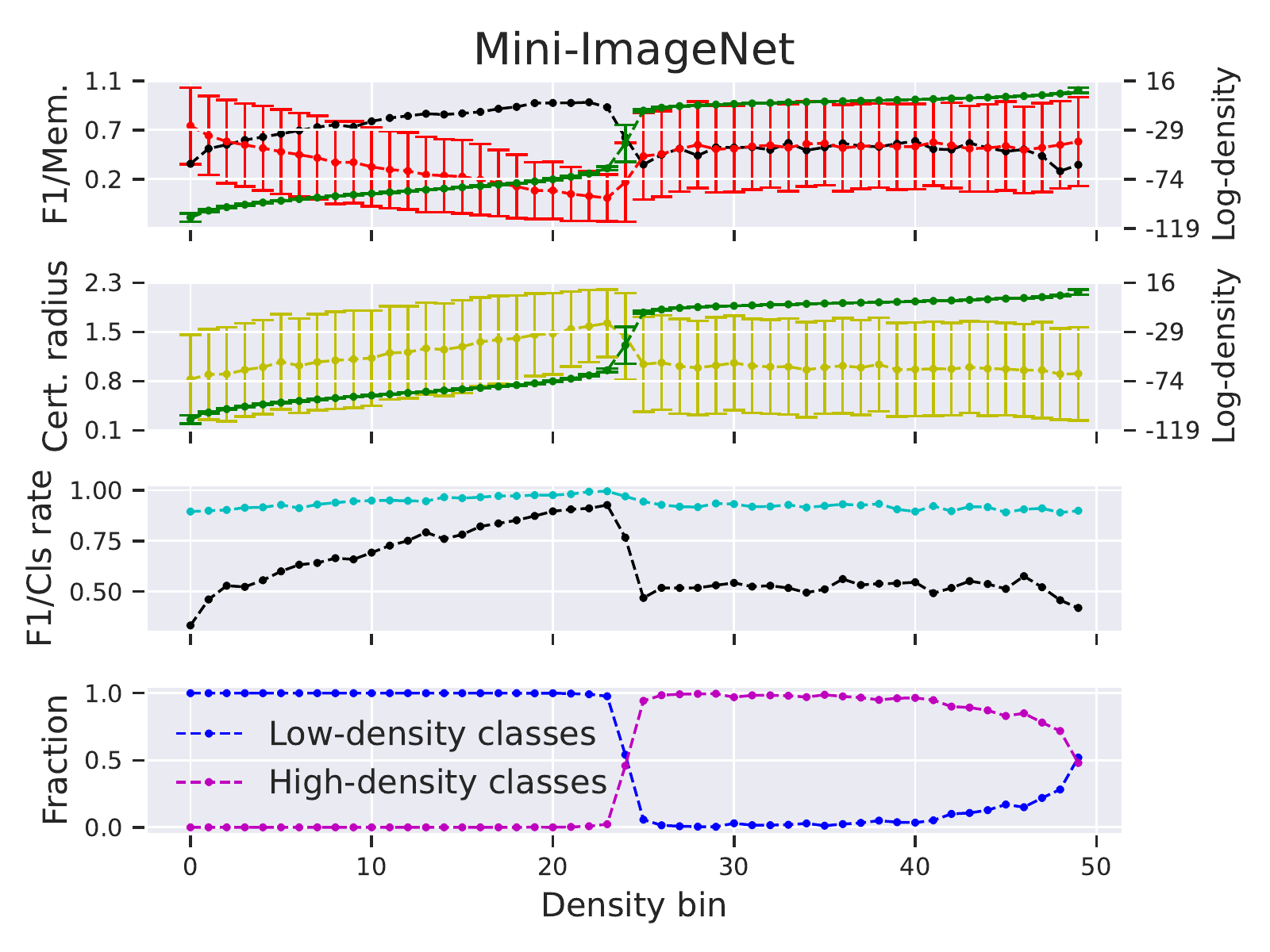}
  \end{tabular}
  \includegraphics[width=0.75\linewidth]{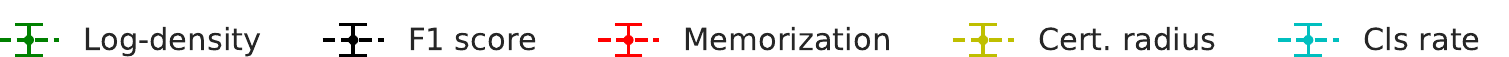}

  \caption{Relationship between class conditional log-density, input memorization and adversarial robustness.
           Low/high-density fraction: fraction of points in a bin from the low-/high-density classes.}
  \label{fig:mem_arobst_vs_density}
\end{figure*}

Our results so far uncover an unexpected structure in the representations of learned classes. An immediate question
that follows from this observation is: does this structure correlate with some phenomena observed in neural networks? We
identify two such phenomena: memorization of input examples and adversarial robustness.

To demonstrate the relationship between input memorization and class representations, we sorted the training examples
according to their estimated class-conditional log-density and then split this ordering into 50 equally sized bins. We
then calculated for each bin the mean and the standard deviation of memorization scores and class-conditional
log-densities of examples assigned to it. We also calculated the ratio of examples belonging to the low- and the
high-density classes. Finally, for each bin we trained a separate ResNet model without using the training examples
assigned to that bin. We then calculated the F-score of this model on the examples from the bin. Results are reported
in Fig.~\ref{fig:mem_arobst_vs_density}.

Even though memorization scores exhibit large variance, averaging across density bins uncovers a difference in their
distribution between the low- and the high-density classes. More precisely, up to the transition point between the low-
and the high-density classes, the degree of memorization of an input examples decreases---on average---with increasing
class-conditional log-density. This agrees with an intuition that increasing class-conditional density translates to an
input example that is increasingly similar at the neural representation level to many other class members. However,
the trend changes abruptly with the transition to the high-density classes. The transition is marked by an abrupt
increase in average memorization of input examples, which subsequently remains largely independent of the
class-conditional density. These observations are corroborated with the calculated F-scores, which are in close
agreement with mean memorization scores.

At first glance, results in Fig.~\ref{fig:mem_arobst_vs_density} may seem counter-intuitive. However, we argue that they
corroborate the long-tail hypothesis put forward by~\citet{Feldman2020a}. Concretely, our results indicate that
distributions of neural representations in the high-density classes have many distinct components. That is, the
high-density classes are formed from smaller subpopulations of examples with distinct neural representations. The
frequency of each such subpopulation in the training data will be below the class frequency. Feldman's hypothesis
suggests that in order to minimize the generalization error, the learning model may need to memorize some of the
examples from these subpopulations. And indeed, we observe an increase in the degree of memorization when the
distribution of input examples switches from the low- to the high-density classes.

Compact and spatially separated components in the high-density classes should---intuitively---be less robust to an
adversarial attack. In particular, a relatively small input perturbation may move the representation of the attacked
example outside of its component. This could be verified by comparing low- and high-density classes w.r.t the robustness
against a selection of adversarial attacks. However, given the vast number of attacks proposed so far, we to opt to
explore this hypothesis in an attack-agnostic way. To this end, we evaluate the performance of a provable adversarial
defense in function of the class-conditional density. Concretely, we take the training examples split into density bins
and for each bin train a smoothed classifier proposed by~\citet{Cohen2019} without using the examples from the bin. We
use ResNet networks as the base (smoothed) models. We then evaluate the \textsc{Certify} procedure \citep{Cohen2019} for
the examples from the held-out bin (See Appendix for details). The procedure either abstains from prediction or returns
a certified radius $r$ and the predicted class. Importantly,~\citeauthor{Cohen2019} proved that if not abstaining,
\textsc{Certify} returns with high-probability the class that will be predicted by the smoothed classifier for inputs no
further than $r$ from the certified example. Note, however, that the smoothed classifier need not agree with the base
classifier. In this sense it trades classification accuracy for provable robustness.

In Fig.~\ref{fig:mem_arobst_vs_density} we report per-bin certification radius (mean and standard deviation), fraction
of inputs for which \textsc{Certify} did not abstain (classification rate) and F-score of the smoothed classifier.
Results confirm that high-density classes correlate with lower adversarial robustness: transition from the low- to the
high-density regime coincide with abrupt decrease in certified radii and F-scores of the smoothed classifier.
\textsc{Certify} is also slightly more likely to abstain in high-density classes.

%% file: class_divergences.tex
\begin{figure*}[htb]
  \centering
  \begin{tabular}{cc}
  	ResNet18 & ResNet50 \\
    \includegraphics[width=0.23\textwidth]{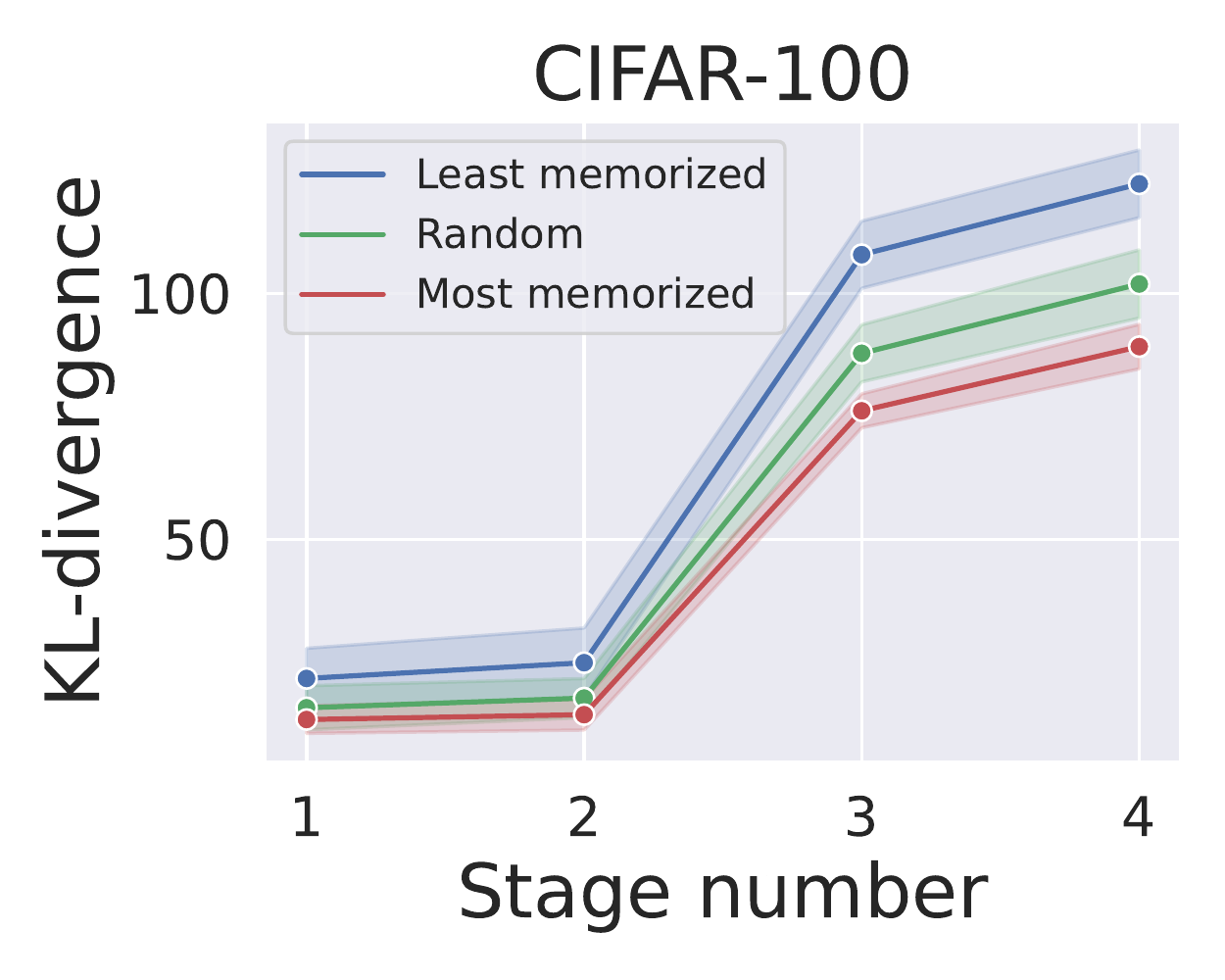}
    \includegraphics[width=0.23\textwidth]{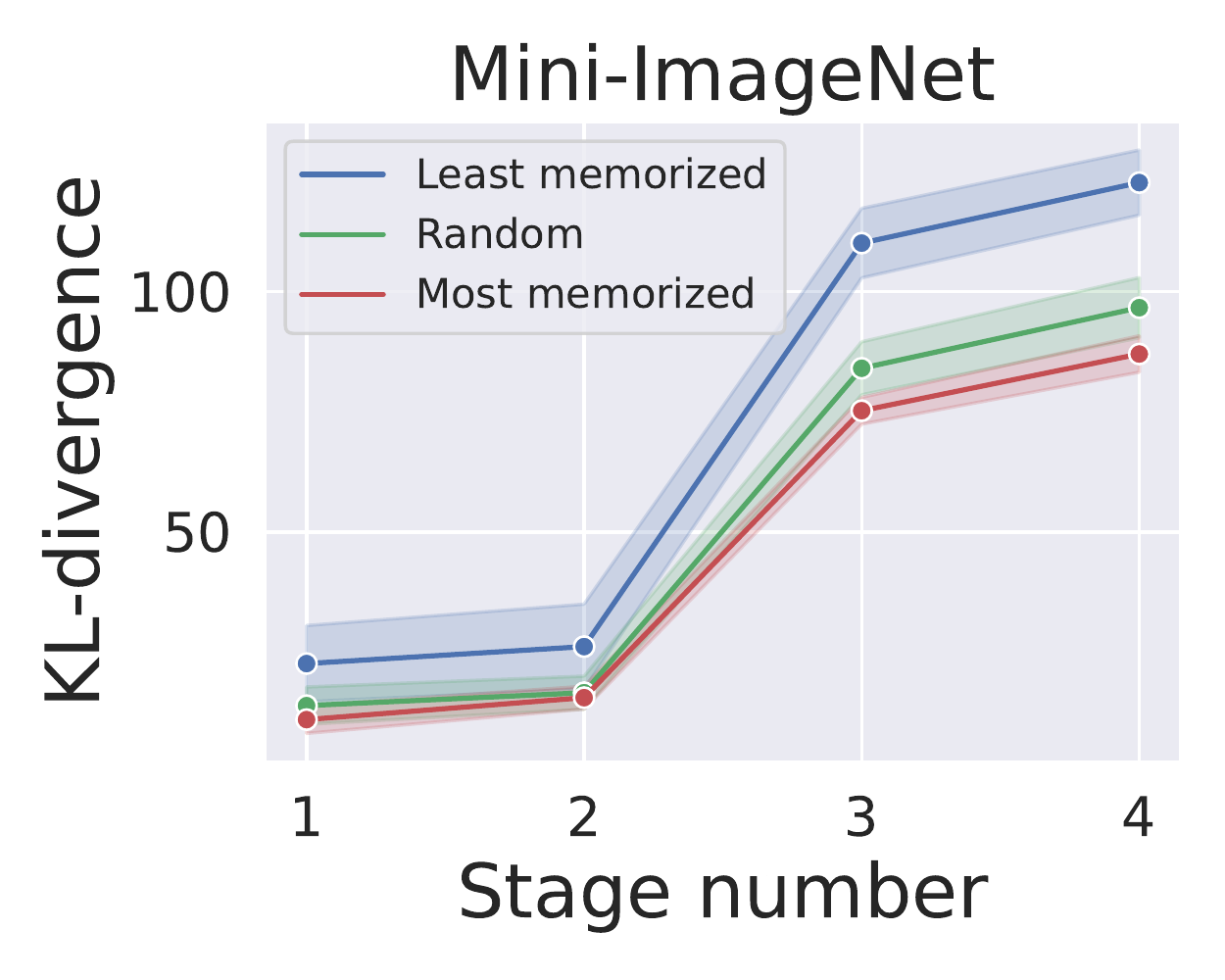} &
    \includegraphics[width=0.23\textwidth]{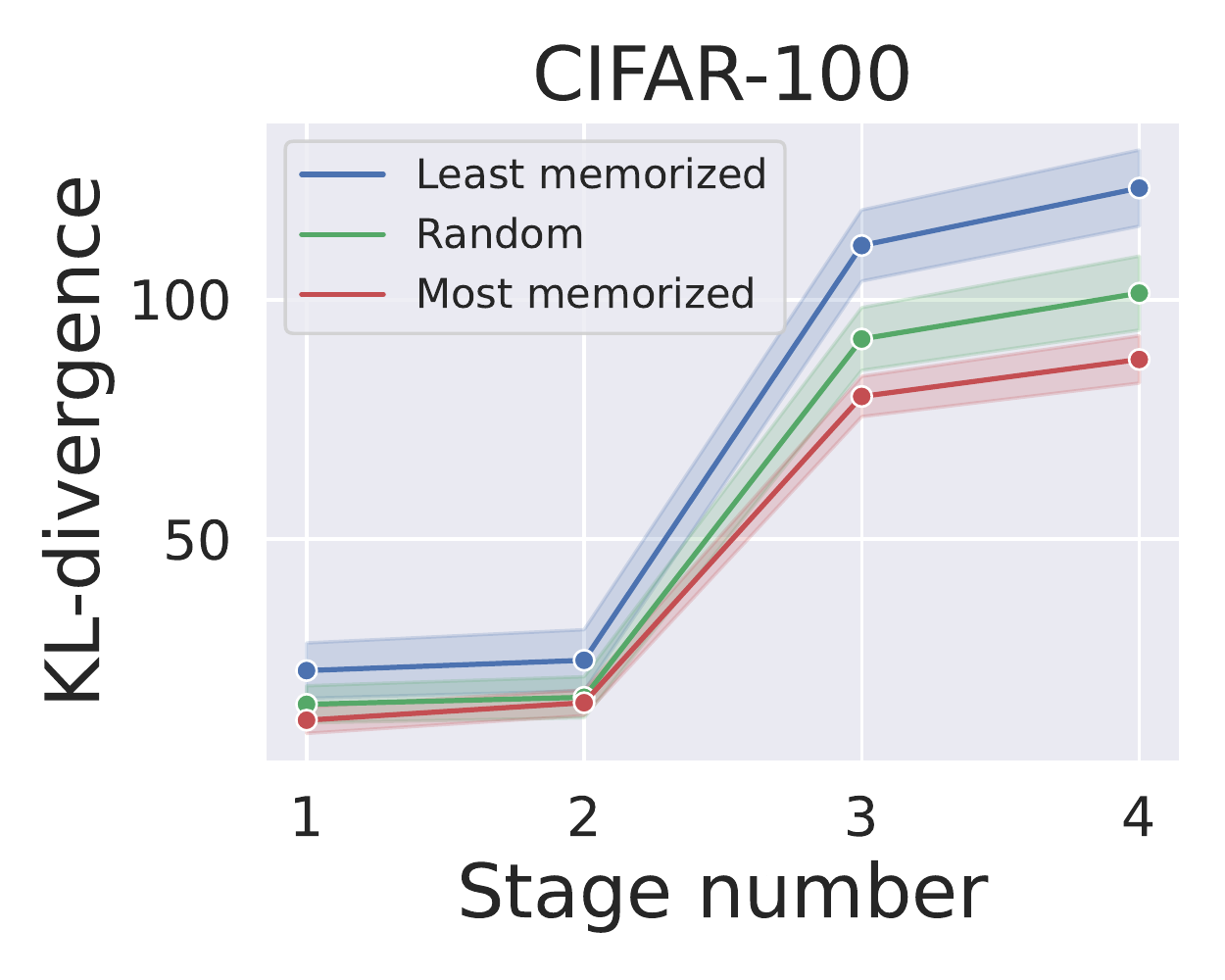}
    \includegraphics[width=0.23\textwidth]{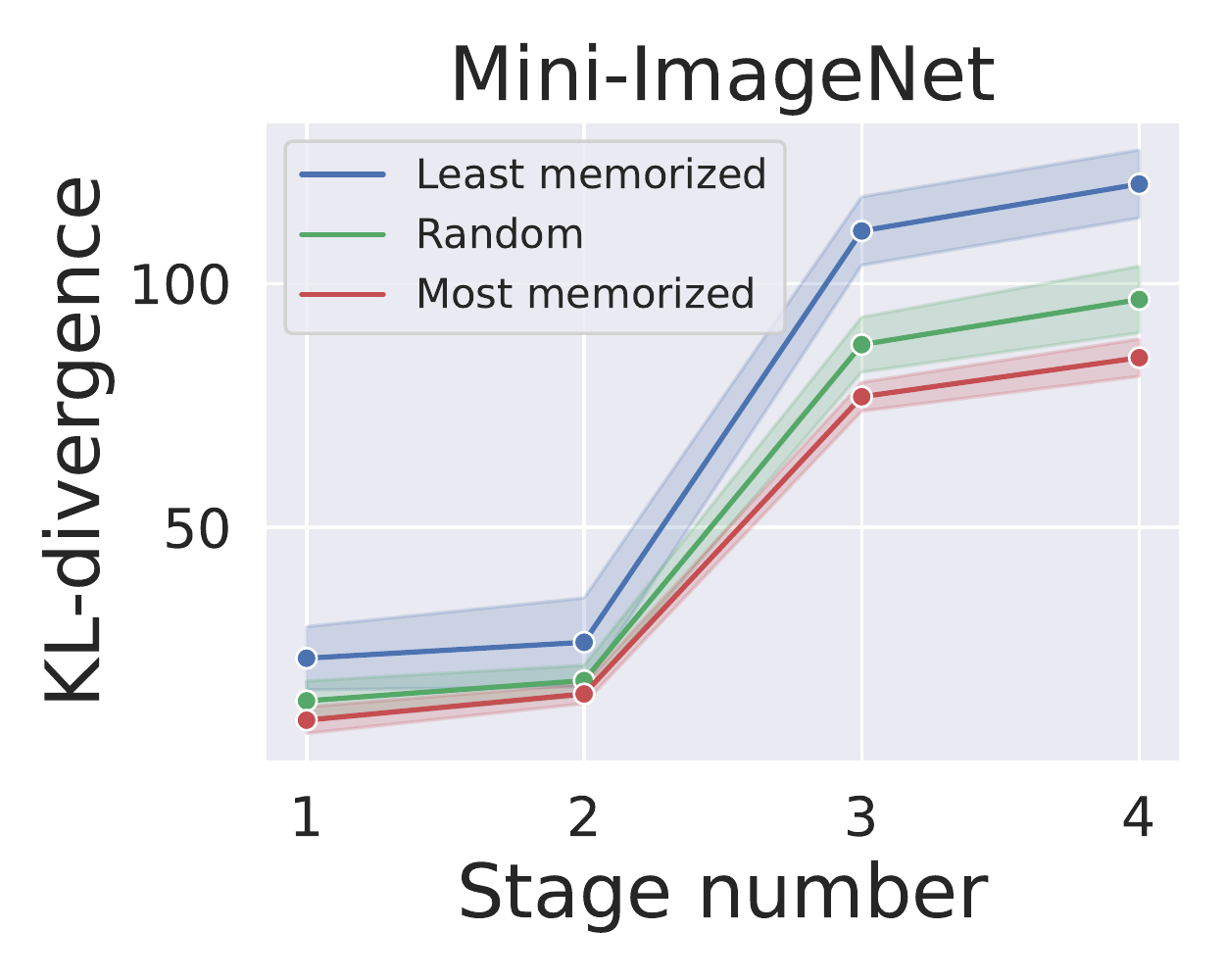}
  \end{tabular}
  \caption{Kullback-Leibler divergences between class-conditional predictive densities.}
  \label{fig:dkls_between_classes}
\end{figure*}

We also use class-conditional distributions of neural representations to uncover where in the network architecture
classes are fit and to show to what extent memorization influences this process. To this end, we selected from each
dataset examples with memorization score above~$0.9$. Additionally, we selected approx. equal number of examples with
the lowest memorization estimates and an equal number of randomly chosen examples. We estimated the class-conditional
posterior predictive distributions separately for the three chosen subsets of examples. Next, we estimated the KL
divergences between predictive densities of every class pair~(Eq.~\eqref{eq:mcmc_dkl}). Note that for a pair of
classes~$(P, Q)$ we estimate the KL divergence from
$p\left(nn_l\left( \mathbf{x} \right) \mid \mathbf{x} \in P \right)$ to
$p\left(nn_l\left( \mathbf{x} \right) \mid \mathbf{x} \in Q \right)$ and the divergence from
$p\left(nn_l\left( \mathbf{x} \right) \mid \mathbf{x} \in Q \right)$ to
$p\left(nn_l\left( \mathbf{x} \right) \mid \mathbf{x} \in P \right)$. The asymmetry of the KL divergence therefore
does not alter the conclusions of this experiment. To recover meaningful predictive densities, we restrict this analysis
to classes that have at least $100$ representatives in all three subsets.

Figure~\ref{fig:dkls_between_classes} reports mean estimated between-class KL divergences, together with one standard
deviation intervals. Additional results are reported in the Appendix. Clearly, input memorization is not fixed to some
specific parts of the network: class-fitting progresses similarly for memorized and typical examples, although memorized
examples induce slightly less distinct class representations. Importantly, class representations are formed mostly
between the second and the third stage of the ResNet model. This agrees with the observation that distinct groups of
classes are evident in the third and the fourth network stage, but not in the first two~stags.

%% file: related_work.tex
Several recent works investigate representations learned by neural networks. The first algorithm to uncover similarities
between representations in these models was proposed by \citet{Raghu2017}. In crux, they use SVD to remove nuisance
factors from network activations, and then uncover shared representations via canonical correlation analysis.
Subsequently, \citet{Morcos2018} refined this approach with a novel method for combining canonical directions, while
\citet{Kornblith2019} proposed a Hilbert-Schmidt Independence Criterion-based metric for comparing neural
representations. Next, \citet{Jamroz2020} proposed probabilistic models for representations learned by kernels in
convolutional networks. They showed that networks that memorize random labels learn significantly more complex
representations than generalizing networks. Note that while we use a similar set of  of probabilistic tools, our goal is
not to characterize distributions of features learned by network units. Instead, we use class-conditional probabilistic
models to capture a notion of class representations in neural networks.

A growing line of research touch on the memorization of input examples by neural networks. \citet{Zhang2017} demonstrate
that neural networks can easily fit a dataset with randomly permuted labels, often despite explicit regularization
during training. \citet{Arpit2017} compare networks that fit corrupted labels and networks trained on uncorrupted data
with respect to the way they fit input examples. \citet{MixupZhang} proposes a new augmentation method that encourages
the model to interpolate linearly between training examples. Their results suggest that such augmentation can prevent
neural networks from memorizing random datasets and make them less prone to adversarial attacks. Effects of memorization
were also covered by~\citet{ExploringGenNey}, who derived several metrics for the degree of memorization by bounding the
Lipschitz constant in the model's transformation. Finally, \citet{Feldman2020a, Feldman2021, Feldman2020b} propose an
alternative perspective on memorization.  They consider a learning task where data is assumed to follow a long-tailed
distribution with many infrequent components. \citeauthor{Feldman2020a} shows that in this scenario minimization of the
generalization error requires memorizing some of the examples.

\citet{Szegedy2013} demonstrated vulnerability of neural networks to an adversarial attack. Subsequently,
\citet{AdvGoodFellow2015} proposed a simple method for constructing adversarial examples.  These results prompted
significant research efforts on adversarial attacks, defence strategies and certification of models' predictions---see
\citet{Huang2020survey} for a recent survey.

%% file: conclusions.tex
In this work we used Bayesian mixture models with unknown number of components to investigate representations of classes
in residual convolutional networks. Our main finding is that classes in investigated neural models are fit in two
distinct ways. Namely, we uncover a group of classes in which high-level neural representations appear to form compact
and spatially separated components. We showed that examples from these classes are memorized to a higher degree than
examples with intermediate class-conditional log-density estimates. We also showed that these classes are less robust to
an adversarial attack. Finally, we used class-conditional density models to uncover where in the network structure class
representations are formed for typical and memorized examples.

Our findings gives further experimental support for the perspective on memorization proposed in~\citep{Feldman2020a,
Feldman2020b}. In particular, \citet{Feldman2020a} argues that memorization of examples from infrequent components is
necessary to minimize generalization error in learning tasks where data comes from a long-tailed distribution. We do
observe increased memorization in classes that appear to be formed from distinct---at the neural representation
level---subpopulations of examples. While we do not claim to have a theoretical explanation for why representations of
these classes are so strikingly different than representations of low-density classes, we point out that current
convolutional architectures evolved largely through experimental optimization of performance on several image
classification benchmarks. \citet{Feldman2020b} identify cases in these benchmarks where memorization of certain inputs
improve predictions on evaluation sets. Therefore, it may simply be that convolutional networks used today were
unwittingly optimized to fit classes the way we observe in this work.

%% file: technical_appendix.tex
\renewcommand{\theequation}{A.\arabic{equation}}
\renewcommand{\thefigure}{A.\arabic{figure}}
\renewcommand\appendixpagename{\centering Technical Appendix\\[2em]}

\onecolumn
\appendixpage

\section{Block collapsed Gibbs sampler}

\input{appendix_block_cgs}

\section{ResNet representations}

\input{appendix_activations}

\section{Evaluation of adversarial robustness}

\input{appendix_adv_robustness}

\section{Distinct modes of class fitting in residual convolutional networks}

\input{appendix_class_fitting}

\section{Class representations in Vision Transformers and MLP-Mixers}

\input{appendix_vit_mixer}

\section{Where neural network fit classes}

\input{appendix_class_divergence}

\section{Computational cost of experiments}

All neural networks used in this work were trained on NVIDIA Tesla V100 GPUs, with one training run taking under 2 hour
on a single GPU. Density models were estimated on 24-core Intel Haswell nodes equipped with 128 GB of RAM. A typical
collapsed Gibbs sampler run in these settings takes between~1 and~3 days, depending on the input dimensionality.
Computing infrastructure used in this work runs under CentOS Linux release 7.9. Versions of all required libraries and
software packages are provided as an environment definition alongside the source code for replicating the experiments.

%% file: appendix_block_cgs.tex
Standard collapsed Gibbs sampler used by~\citet{Jamroz2020}, also called Plain CGS~\citep{NealDpMCMM, BlockCGSJensen},
constructs a Markov chain over component assignments by sampling an assignment~$c_i$ for one pattern
$\mathbf{x}_i \in \ds$ at a time:
\begin{equation} \label{eq:cgs_component_sampling}
  c_i \sim p\left(c_i \mid \bm{c}_t \setminus \{i\}, \bm{x}_i, \alpha, \bm{\theta}\right),
\end{equation}
where $\bm{c}_t \setminus \{i\}$ are component assignments in the current Gibbs step for all patterns except
$\mathbf{x}_i$, and~$\bm \theta$ is the set of parameters of the posterior distribution (over component means and
covariances) induced by~$\bm{c}_t \setminus \{i\}$. In the model given by Eq.~\eqref{eq:dpgmm} this sampling is easy,
because density in~Eq.~\eqref{eq:cgs_component_sampling} has a closed-form solution. However, since this scheme updates
one assignment at a time, consecutive Gibbs steps tend to be strongly correlated. This negatively affects the rate of
convergence to the posterior distribution and reduces the effective sample count in subsequent Monte Carlo estimates. In
other words, plain CGS exhibits slow mixing.

To improve mixing in collapsed Gibbs sampler, one can sample blocks of component assignments. This is the idea behind
\emph{block collapsed Gibbs sampler}~\citep{BlockCGSJensen}. More formally, block CGS picks in each update a group
of observations $\bm{X}_b = \left\{\bm{x}_i, \bm{x}_{i+1}, \ldots, \bm{x}_{i+b}\right\}$, where~$b$ is a fixed block
size, and sample their component assignments
$\bm{c}_{i\ldots i+b} = \left\{c_i, c_{i+1}, \ldots, c_{i+b}\right\}$ from the distribution:
\begin{equation} \label{eq:block_cgs_component_sampling}
  \bm{c}_{i\ldots i+b} \sim p \left(\bm{c}_{i\ldots i+b} \mid \bm{c}_t \setminus \left\{i, i+1, \ldots, i+b \right\}, 
                                    \bm{X}_b, \alpha, \bm{\theta}\right).
\end{equation}
Assignments $c_i, c_{i+1}, \ldots, c_{i+b}$ are sampled independently. For the model given by~Eq.~\eqref{eq:dpgmm}, the
density in~\eqref{eq:block_cgs_component_sampling} again has a closed-form solution.

In this work we use block CGS in all experiments involving estimation of posterior predictive distributions and related
quantities. There are many strategies to form the blocks of observations for sampling. For example,
\citet{BlockCGSWilkinson} discuss several strategies designed to improve mixing rate in highly structured conditional
models. Our strategy for splitting observations into blocks is simple: before every pass over the dataset, we randomly
permute observations~$\bm{x}_1, \bm{x}_2, \ldots, \bm{x}_n$ and then form blocks $\bm{X}_b$ by iterating over this
permutation and taking consecutive $b$-sized batches of observations.

%% file: appendix_activations.tex
\begin{figure*}[h]
  \centering
  \includegraphics[width=0.95\linewidth]{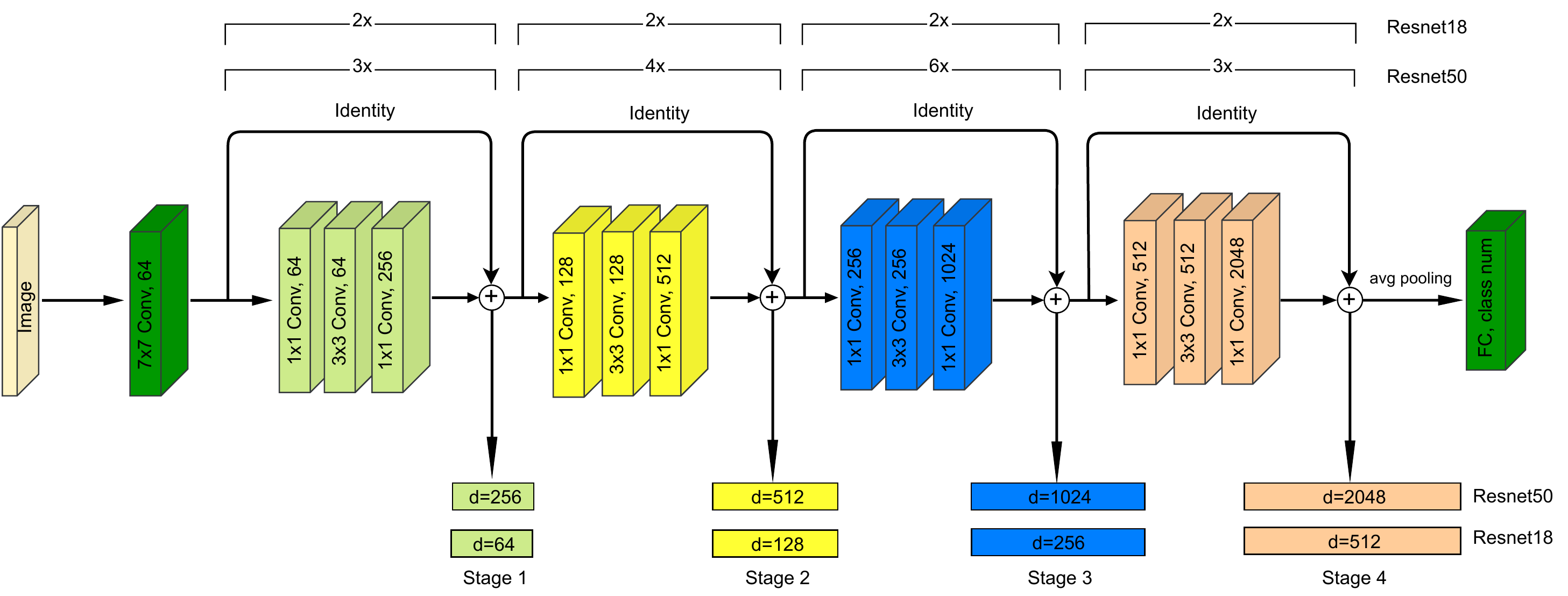}
  \caption{Activation vectors used in this work. ResNet architecture has four \emph{stages} that correspond to
           progressively smaller spatial dimensions. We extract activations at the end of each of these stages.}
  \label{fig:resnet_stages}
\end{figure*}

ResNet architecture reduces spatial dimensions in four stages. We calculate neural representations from the output
of the summation operation at the end of each stage (Fig.~\ref{fig:resnet_stages}). Main results in this work focus on
the representations calculated from the last stage, namely input to the classification head.

\begin{figure*}
  \centering
  \begin{tabular}{cccc}
    \includegraphics[width=0.23\linewidth]{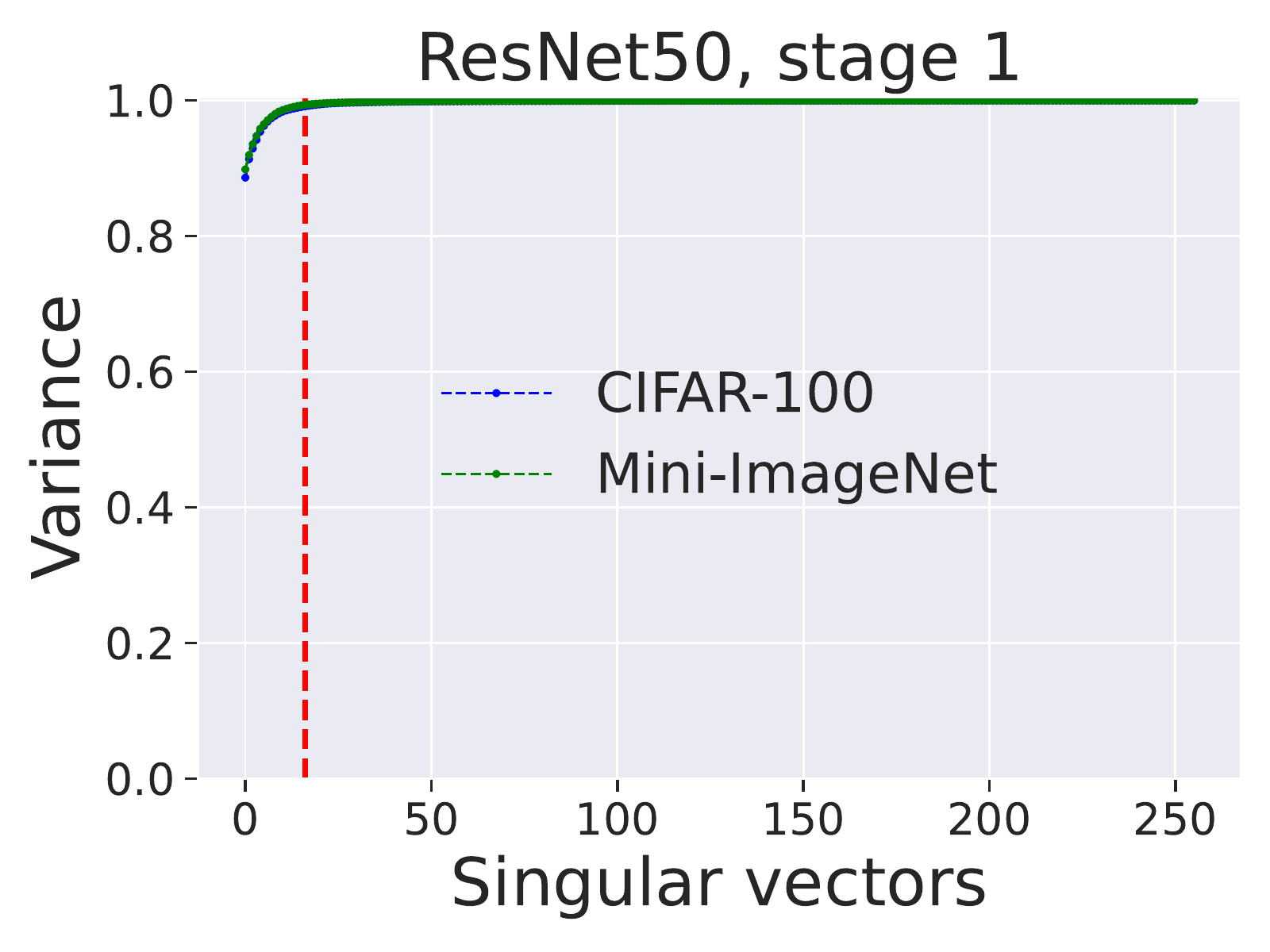} &
    \includegraphics[width=0.23\linewidth]{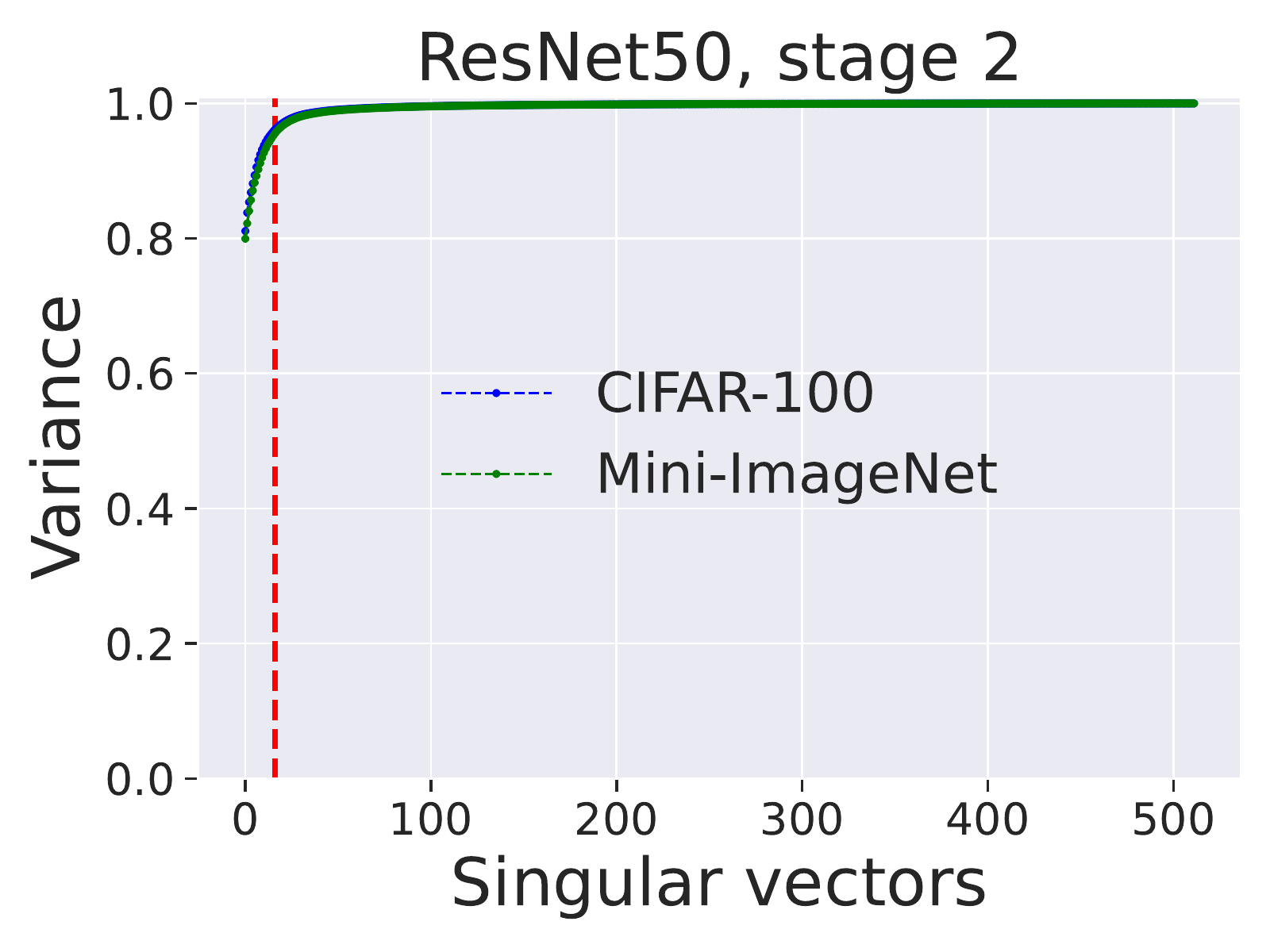} &
    \includegraphics[width=0.23\linewidth]{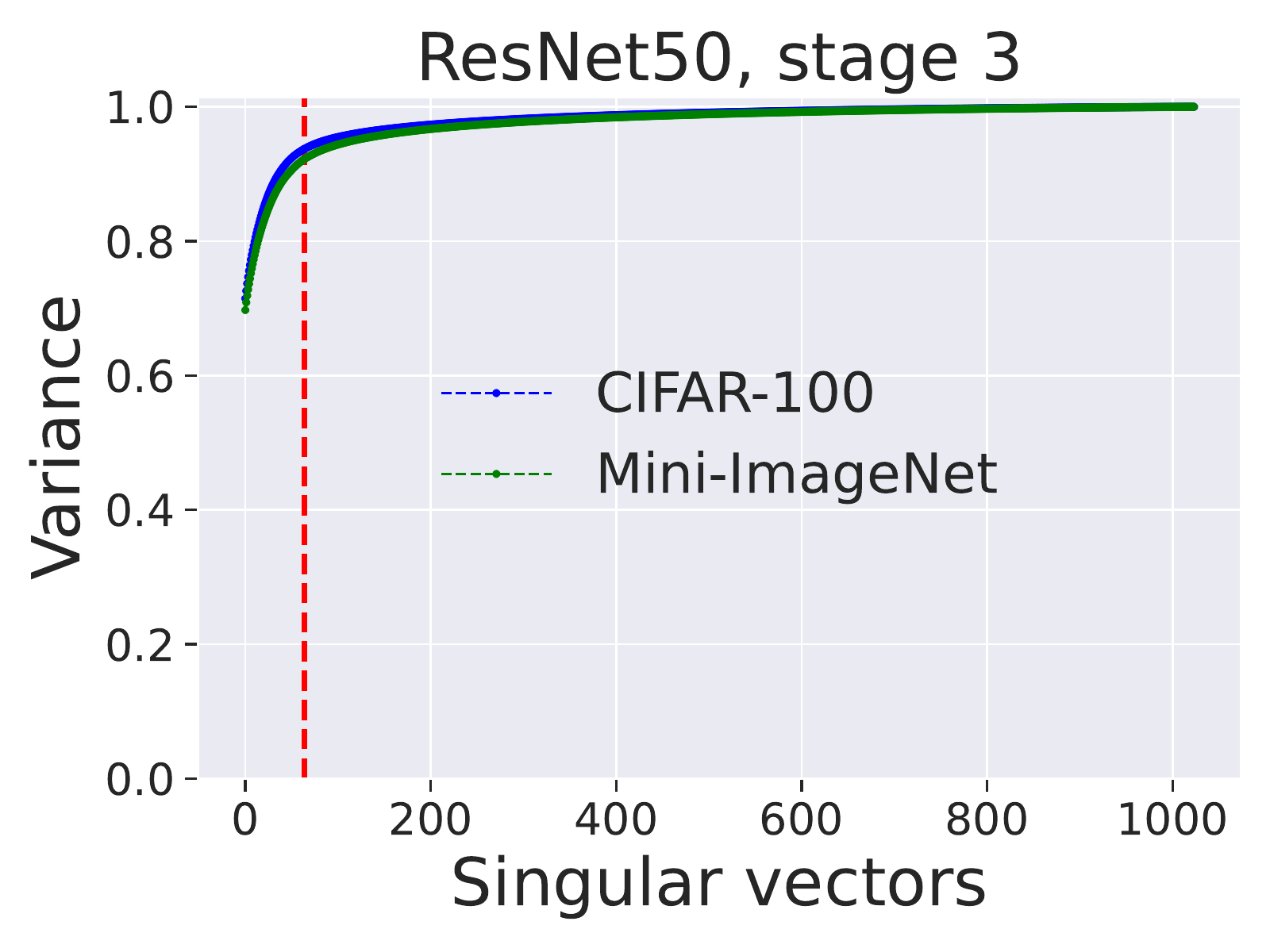} &
    \includegraphics[width=0.23\linewidth]{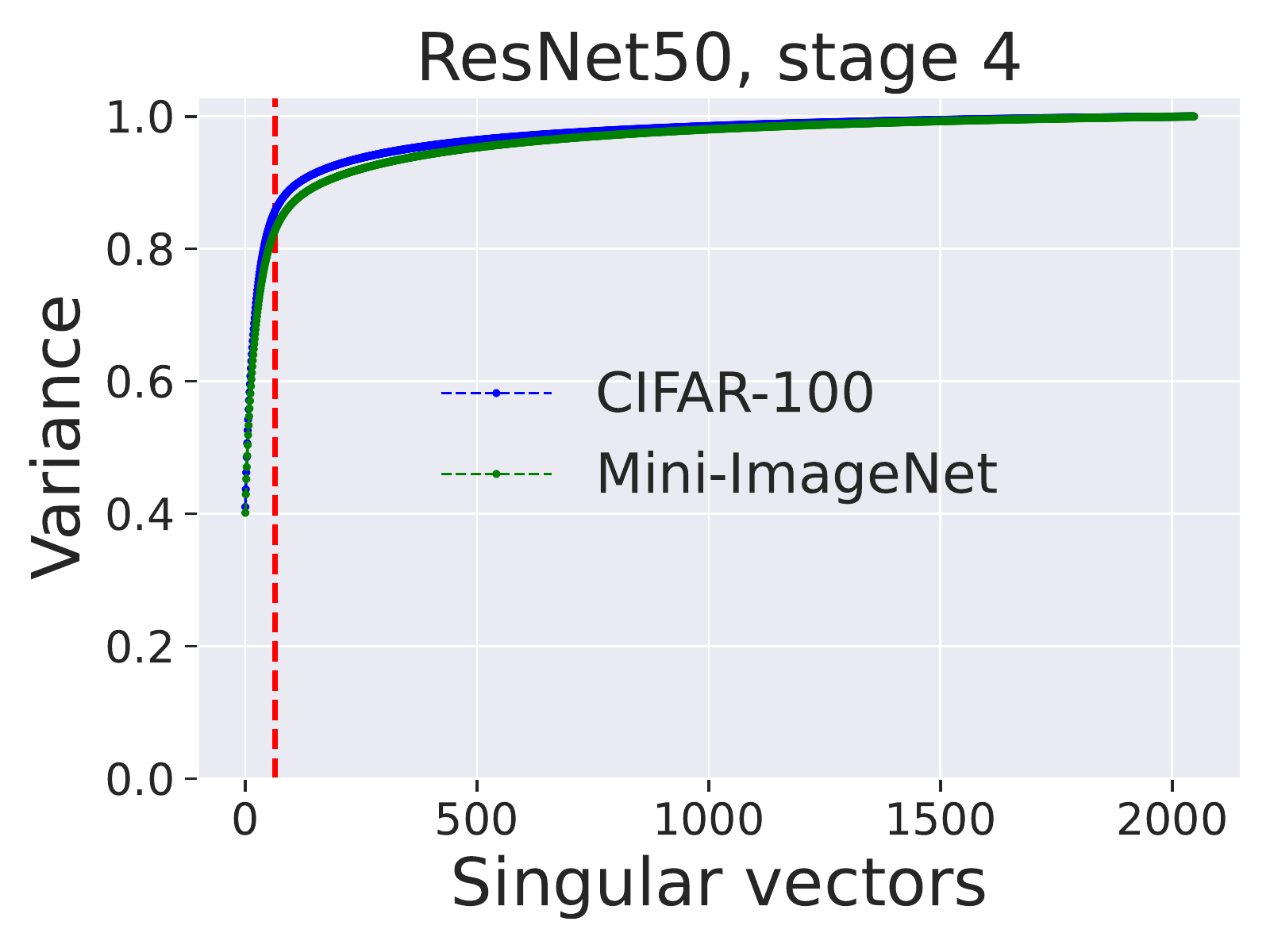} \\
    \includegraphics[width=0.23\linewidth]{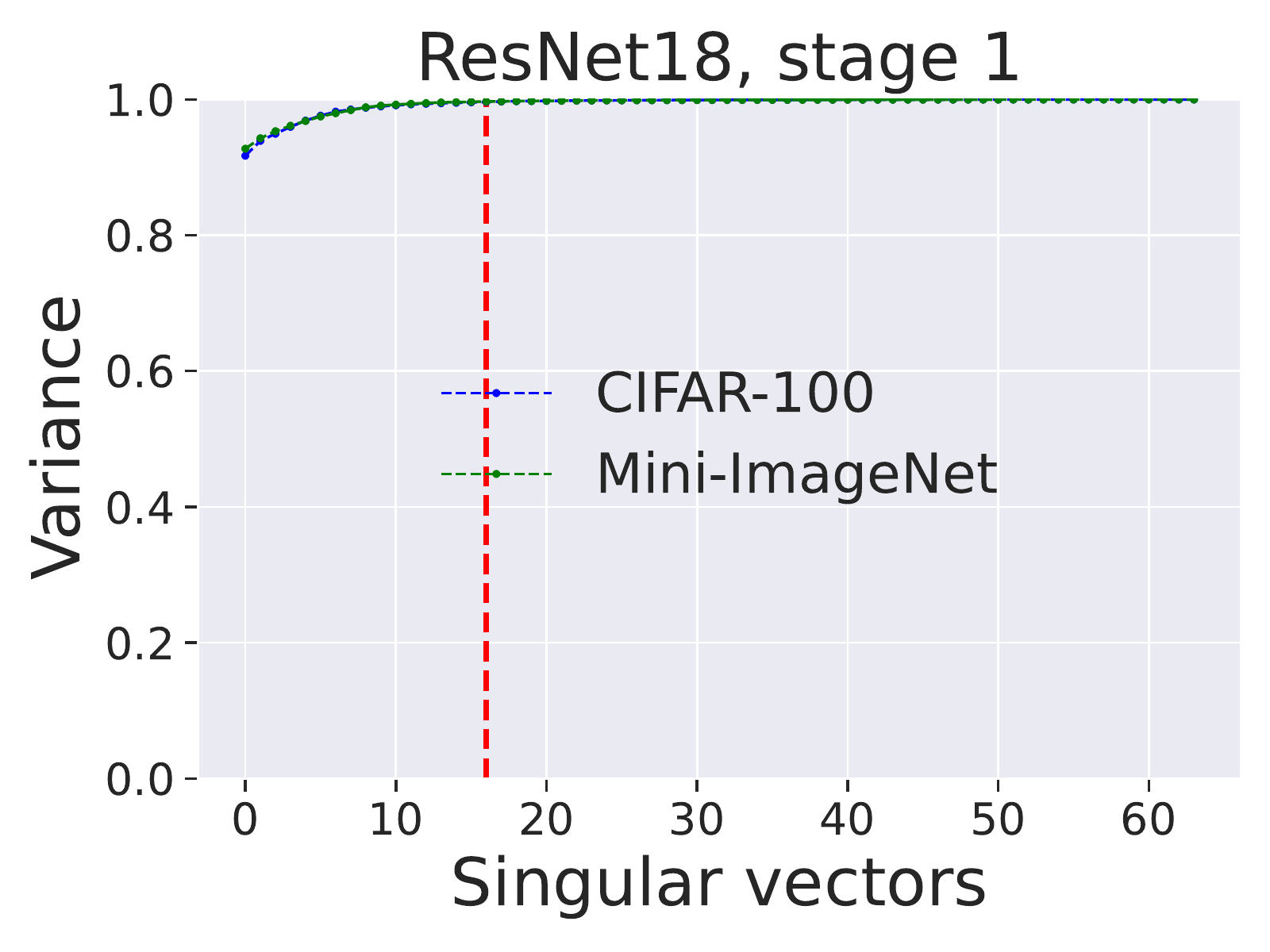} &
    \includegraphics[width=0.23\linewidth]{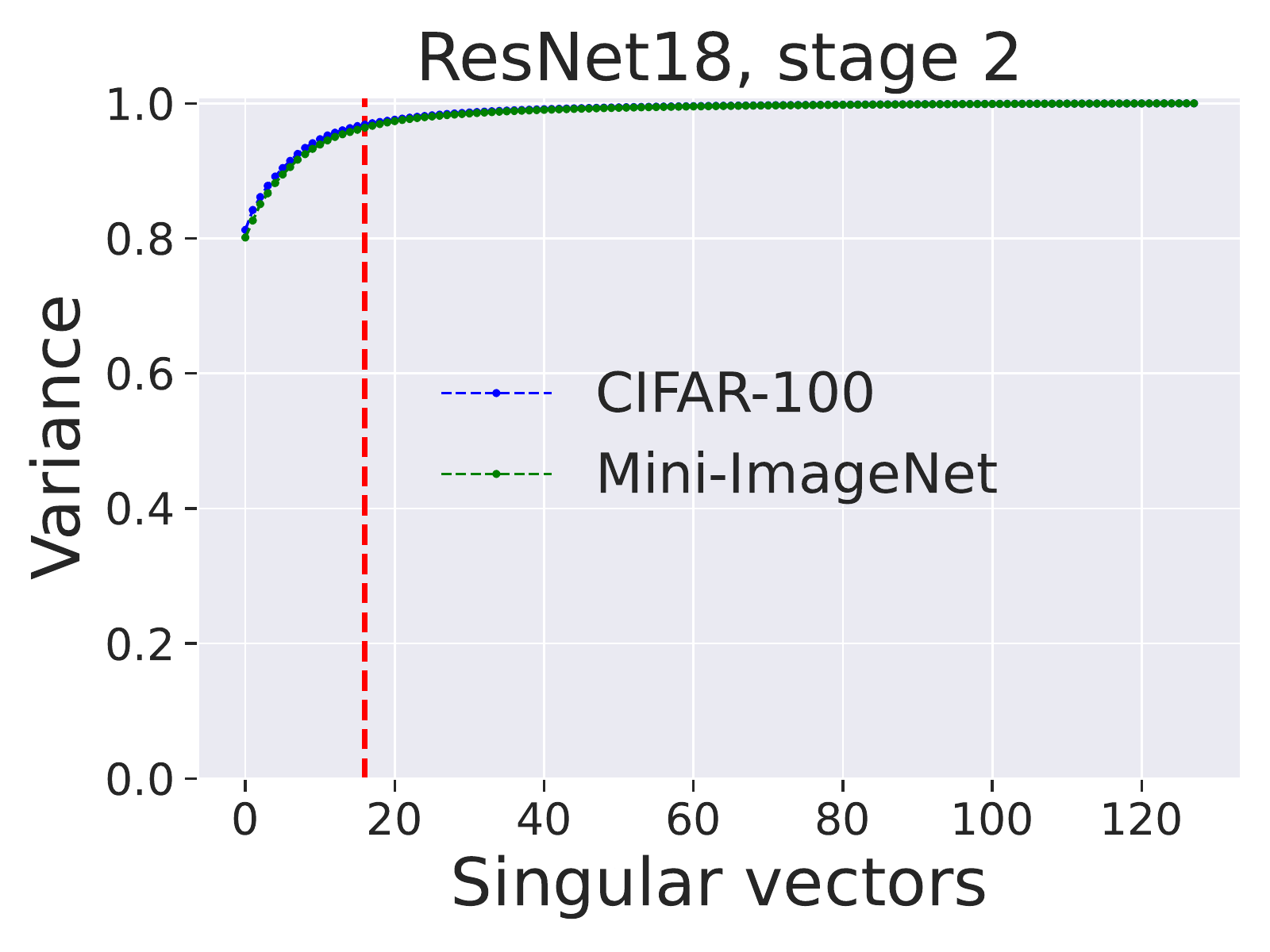} &
    \includegraphics[width=0.23\linewidth]{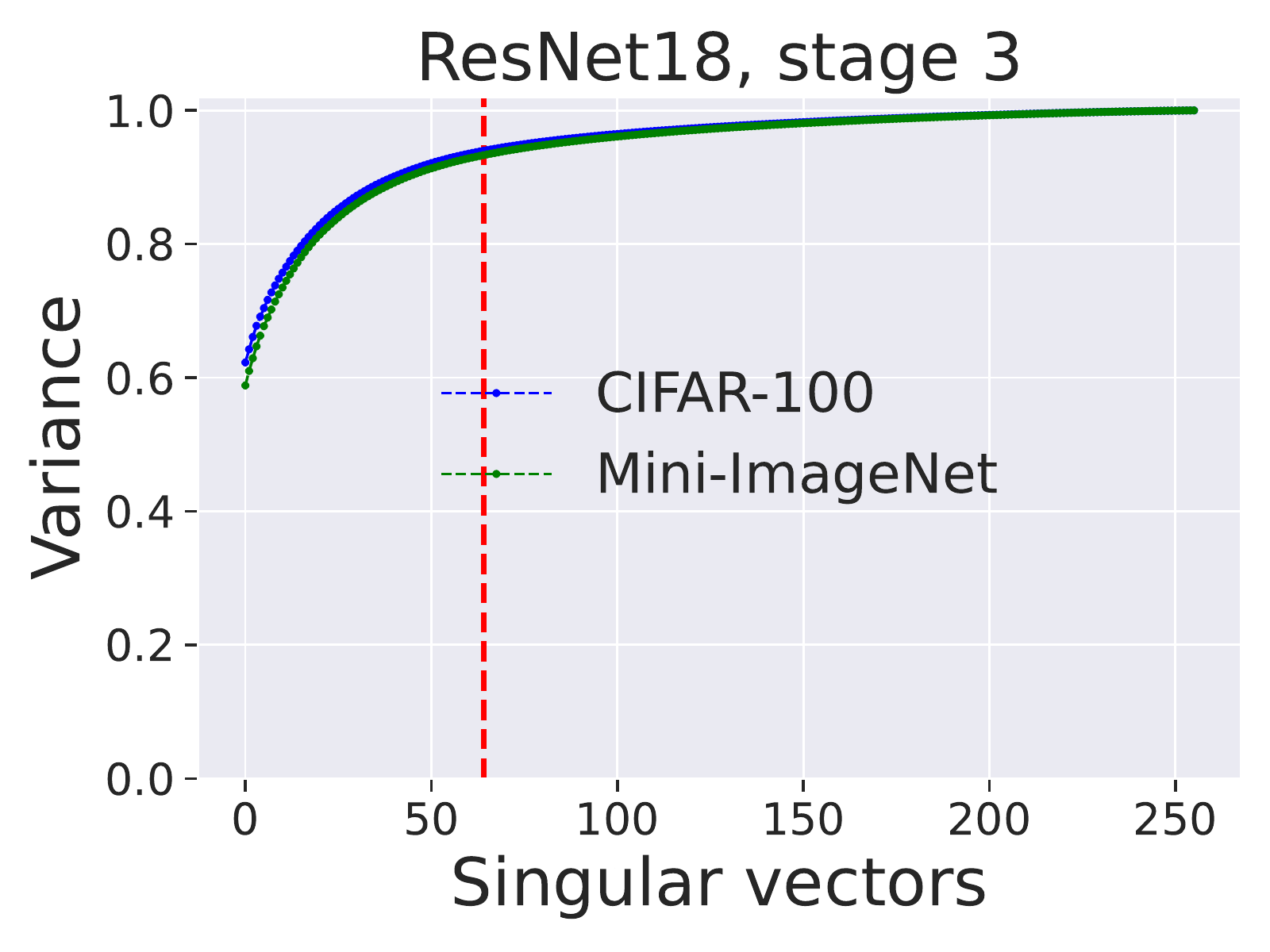} &
    \includegraphics[width=0.23\linewidth]{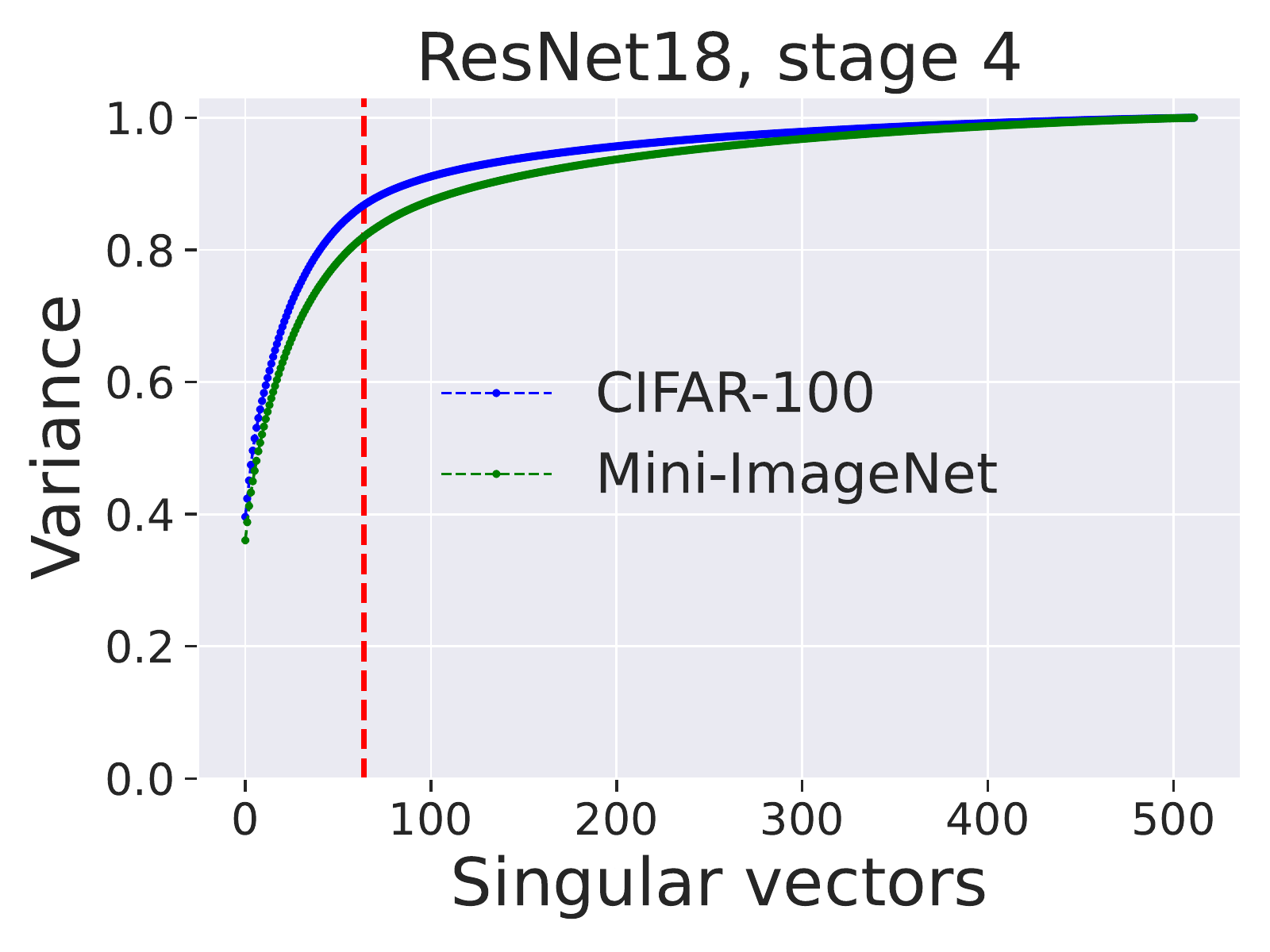}
    \end{tabular}
    \caption{Variance in network activations captured by SVD in function of the number of singular vectors. 
	     Red dashed line marks the number of singular vectors retained in each network stage.}
\label{fig:svd_variances}
\end{figure*}

To reduce the computational cost of estimating full-covariance Gaussian mixture models, we reduce the dimensionality of
the collected neural representations via Singular Value Decomposition~(SVD). That is, we use SVD to decompose the
$n \times d$ matrix that contains neural representations arranged in rows, and then project the data onto a subset of
right-singular vectors corresponding to the largest singular values. We retain~$d'=16$ right-singular vectors for the
first two ResNet stages and $d'=64$ for the third and fourth stage. This allows us to capture most of the variance in
the collected sets of neural representations (Fig. \ref{fig:svd_variances}) and is sufficient to uncover differences
between representations of classes.

%% file: appendix_adv_robustness.tex
For each density bin we train a smoothed classifier proposed by~\citet{Cohen2019} using examples from the bin as
held-out data. We use ResNet18 and ResNet50 networks as baseline models for smoothing. We use the same training setup
as in main experiments, i.e. we follow training hyper-parameters reported by~\citep{Feldman2020b}. However, as suggested
by~\citet{Cohen2019} we add Gaussian noise to the input augmentation pipeline. We use a noise strength $\sigma=0.5$.

We evaluate \textsc{Certify} \citep{Cohen2019} with $n_0 = 100$ and $n = 10^5$ samples. We certify with the Gaussian noise
$\sigma = 0.5$ and the probability of incorrect answer $\alpha = 10^{-3}$.

%% file: appendix_class_fitting.tex
\begin{figure*}[htb]
  \centering
  ResNet18 \\
  \includegraphics[width=0.244\linewidth]{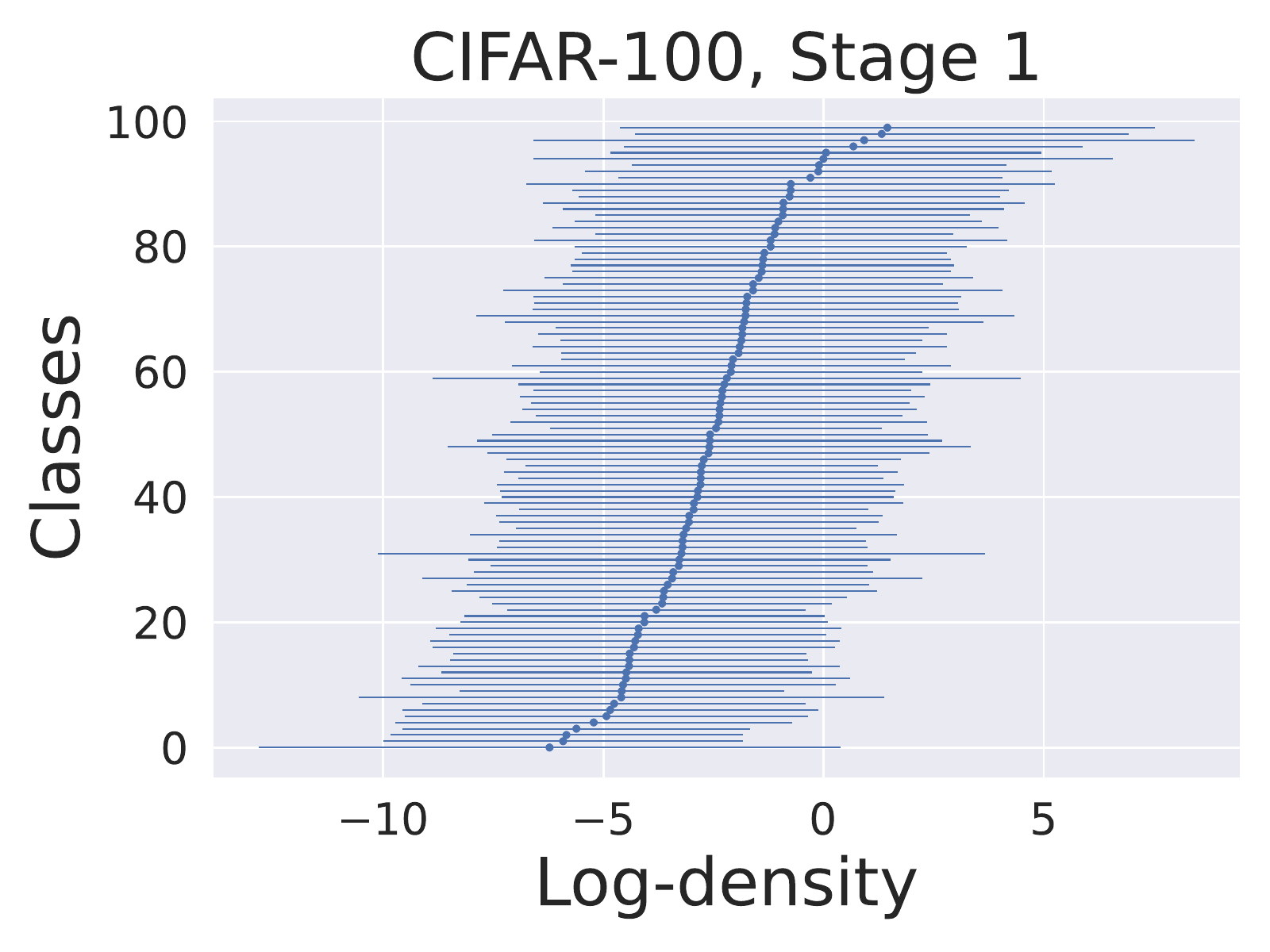}
  \includegraphics[width=0.244\linewidth]{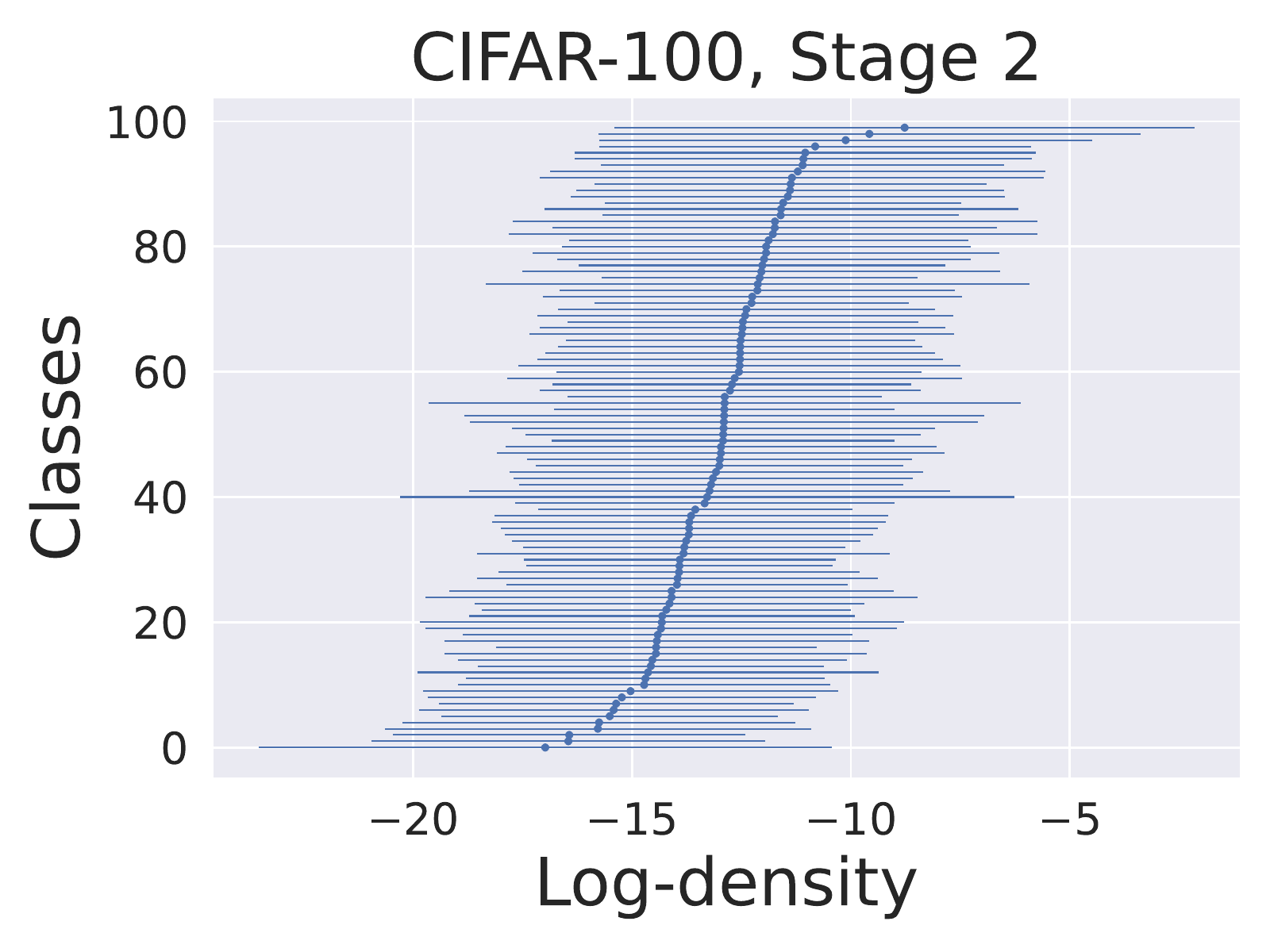}
  \includegraphics[width=0.244\linewidth]{plots/resnet18/cifar100/avg_log_densities_perclass/avg_pooling_2_log_densities.pdf}
  \includegraphics[width=0.244\linewidth]{plots/resnet18/cifar100/avg_log_densities_perclass/avg_pooling_3_log_densities.pdf}
  \\
  \includegraphics[width=0.244\linewidth]{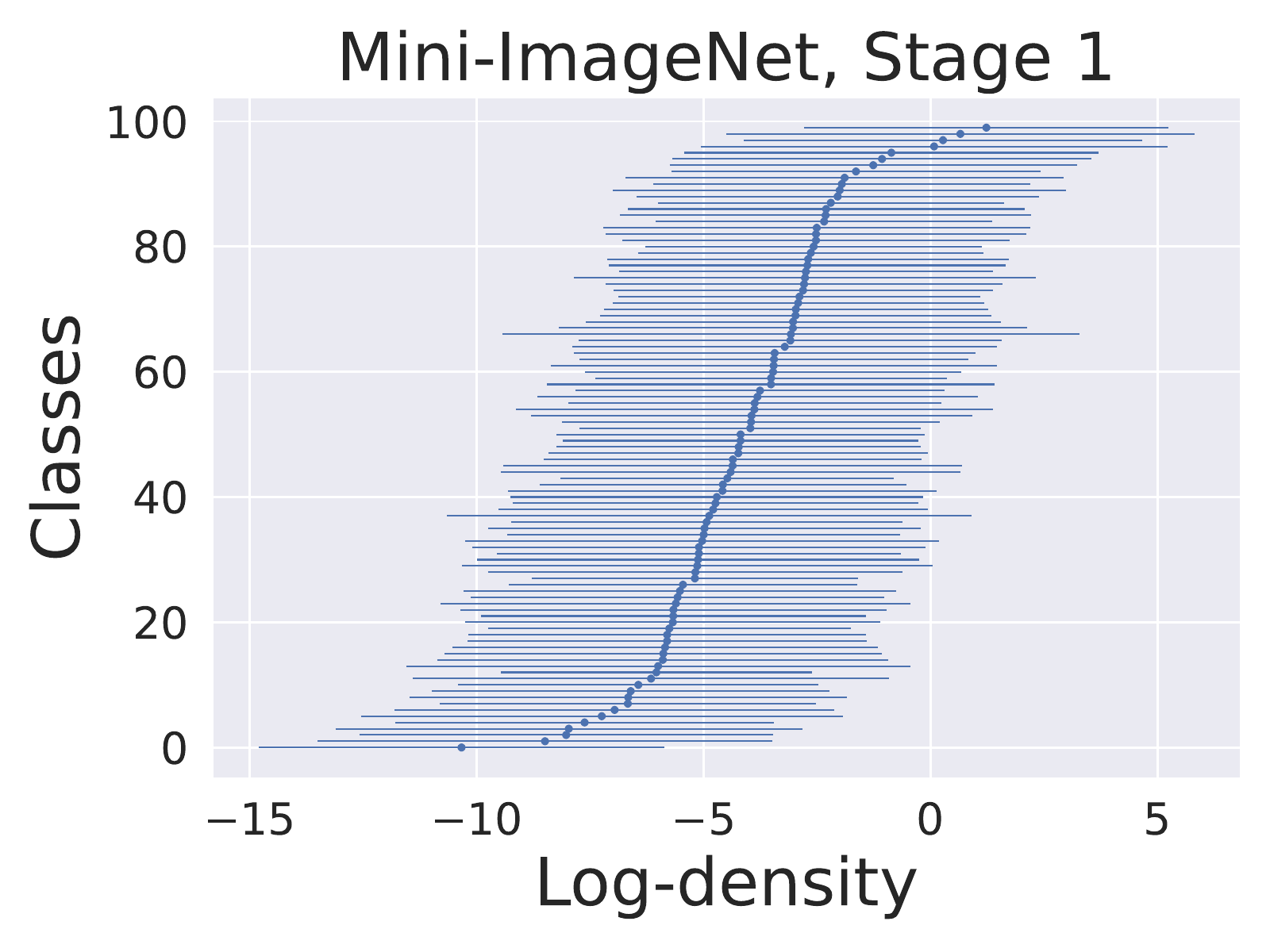}
  \includegraphics[width=0.244\linewidth]{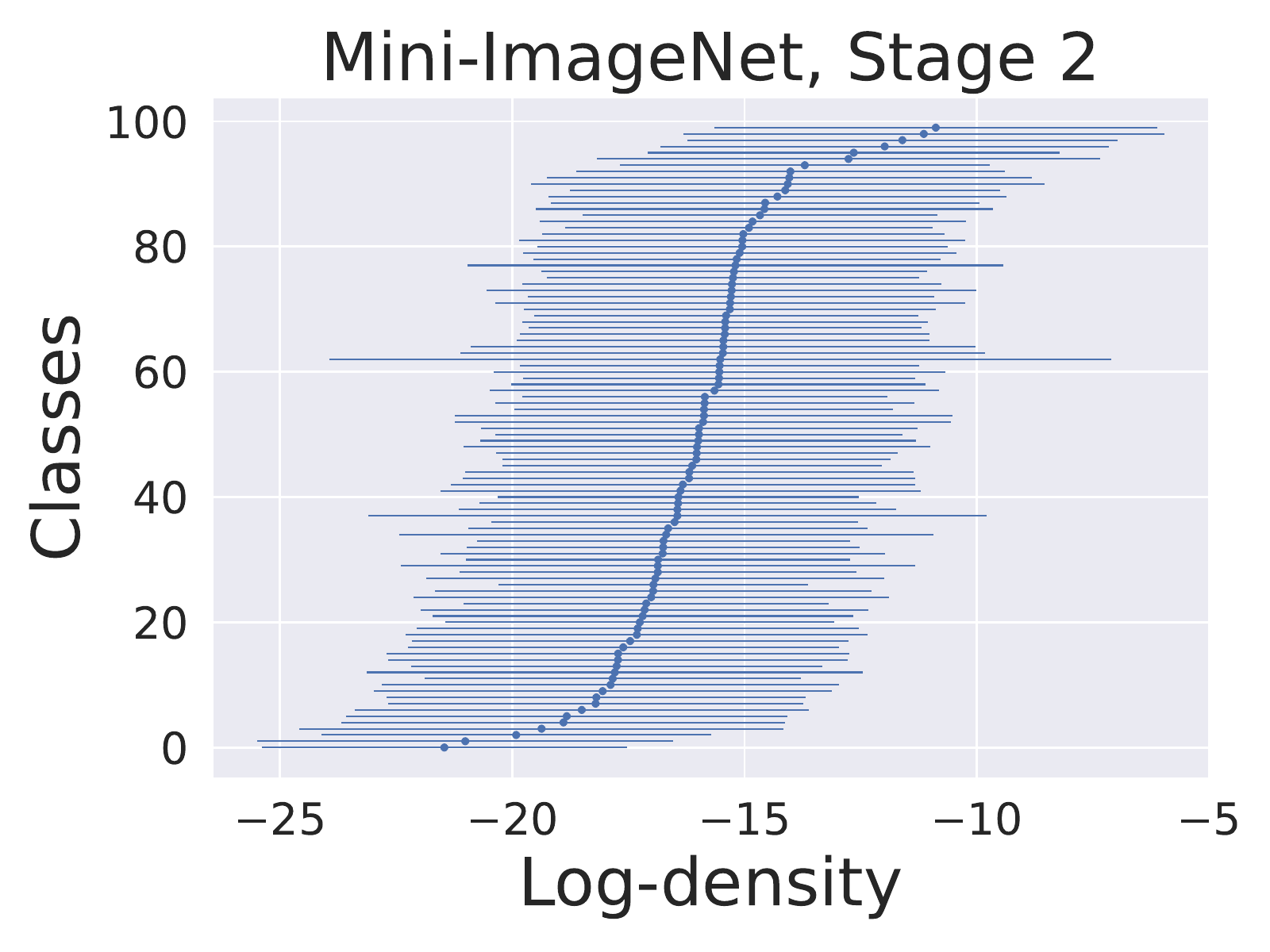}
  \includegraphics[width=0.244\linewidth]{plots/resnet18/mini_imagenet/avg_log_densities_perclass/avg_pooling_2_log_densities.pdf}
  \includegraphics[width=0.244\linewidth]{plots/resnet18/mini_imagenet/avg_log_densities_perclass/avg_pooling_3_log_densities.pdf}
  ResNet50 \\
  \includegraphics[width=0.244\linewidth]{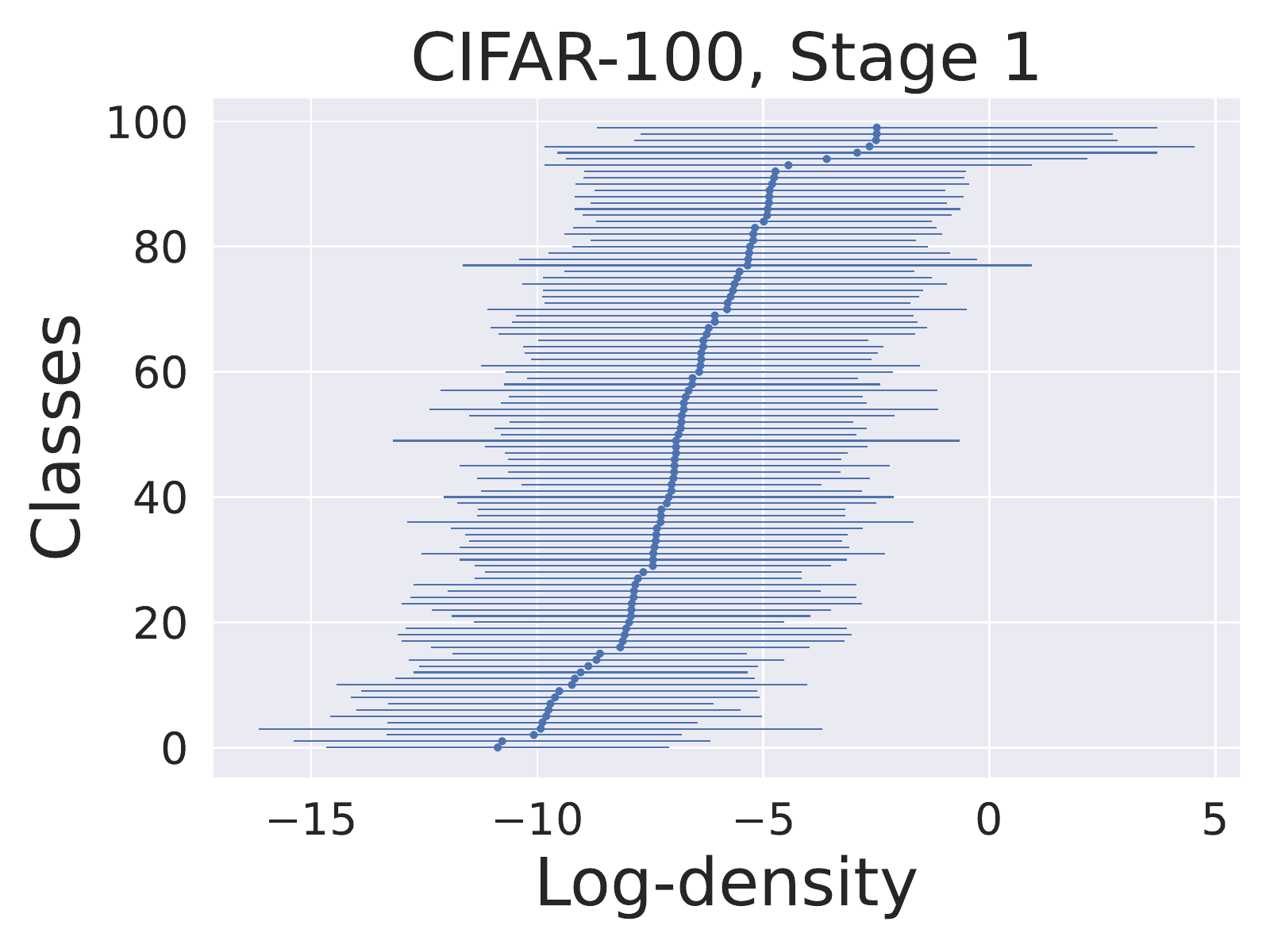}
  \includegraphics[width=0.244\linewidth]{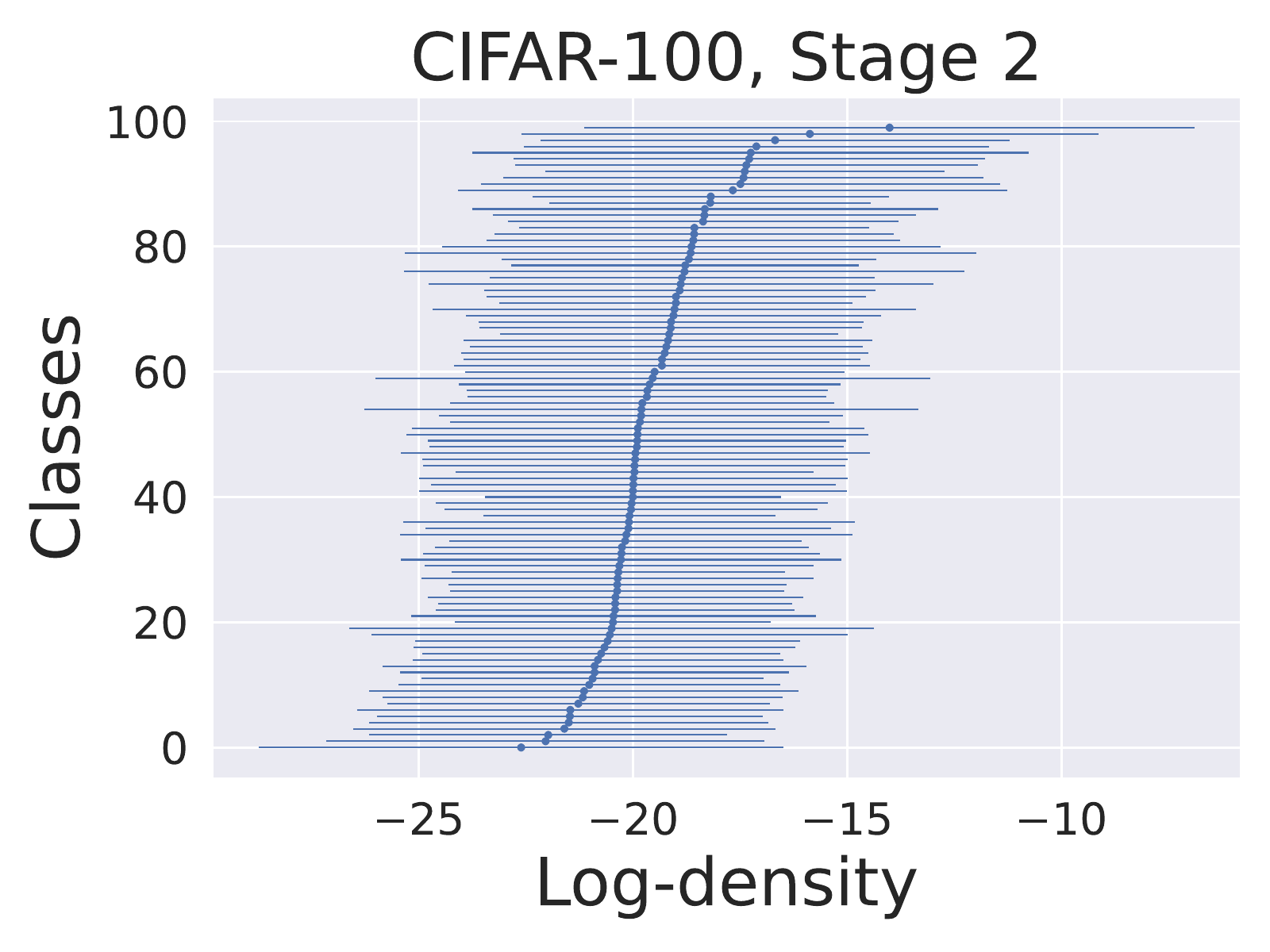}
  \includegraphics[width=0.244\linewidth]{plots/resnet50/cifar100/avg_log_densities_perclass/avg_pooling_2_log_densities.pdf}
  \includegraphics[width=0.244\linewidth]{plots/resnet50/cifar100/avg_log_densities_perclass/avg_pooling_3_log_densities.pdf}
  \\
  \includegraphics[width=0.244\linewidth]{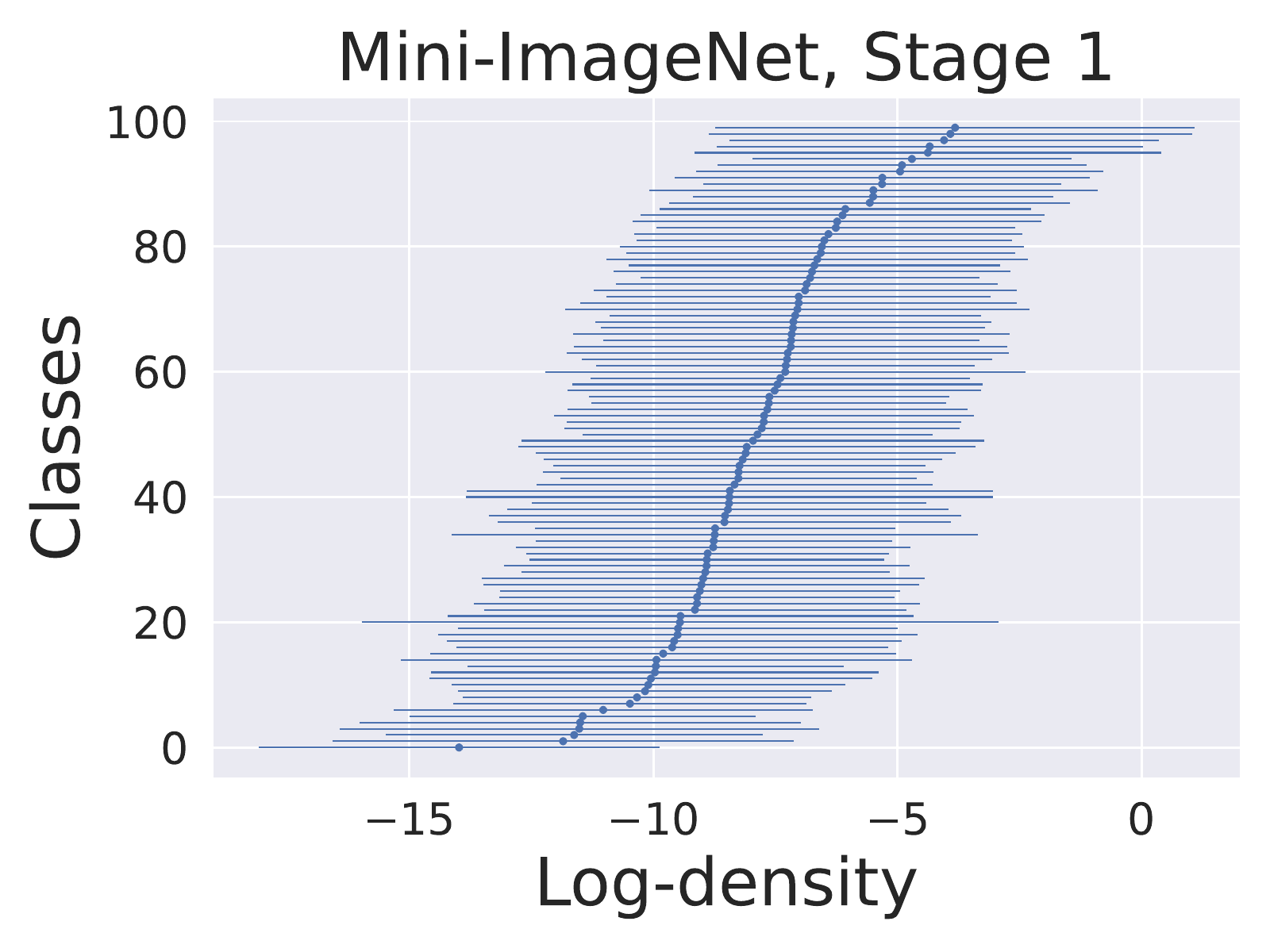}
  \includegraphics[width=0.244\linewidth]{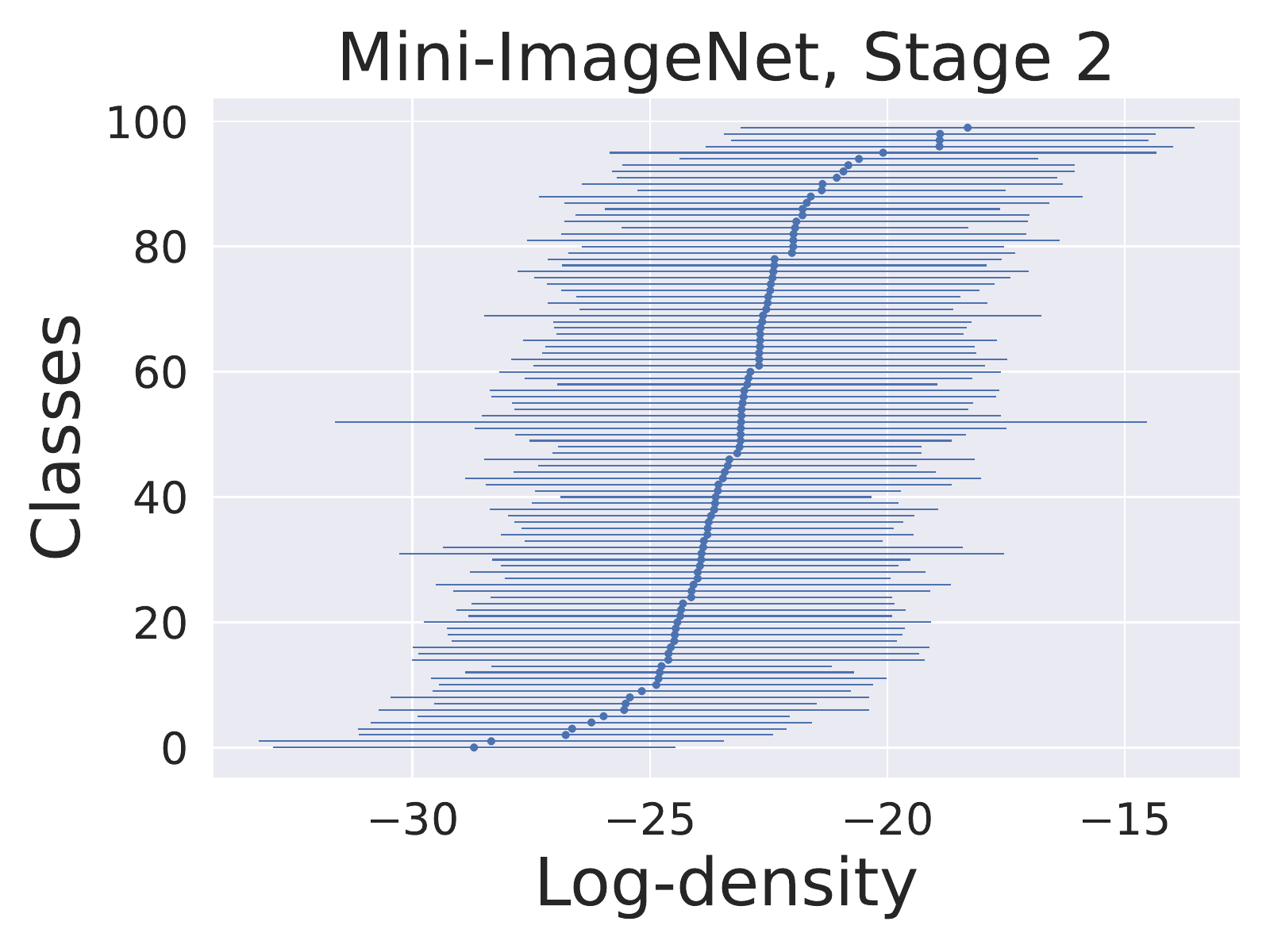}
  \includegraphics[width=0.244\linewidth]{plots/resnet50/mini_imagenet/avg_log_densities_perclass/avg_pooling_2_log_densities.pdf}
  \includegraphics[width=0.244\linewidth]{plots/resnet50/mini_imagenet/avg_log_densities_perclass/avg_pooling_3_log_densities.pdf}
  \caption{Mean and standard deviation of class-conditional log-densities estimated for inputs from each class. Classes
           on the vertical axis are sorted according to the mean class-conditional density.}
  \label{fig:log_cc_densities_all}
\end{figure*}

\begin{figure*}[htb]
  \centering
  Plain ConvNet, CIFAR100 \\[0.25em]
  \includegraphics[width=0.244\linewidth]{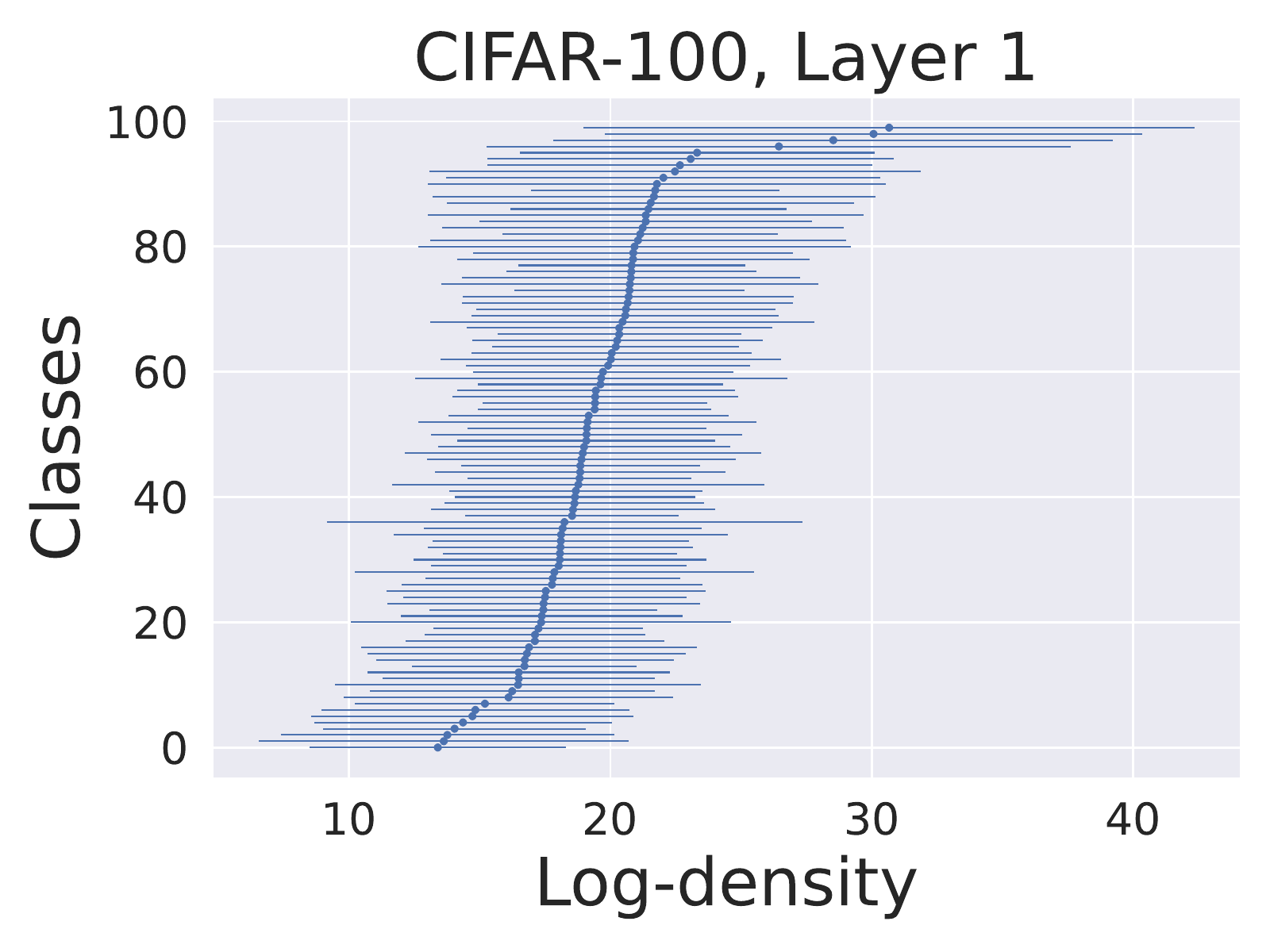}
  \includegraphics[width=0.244\linewidth]{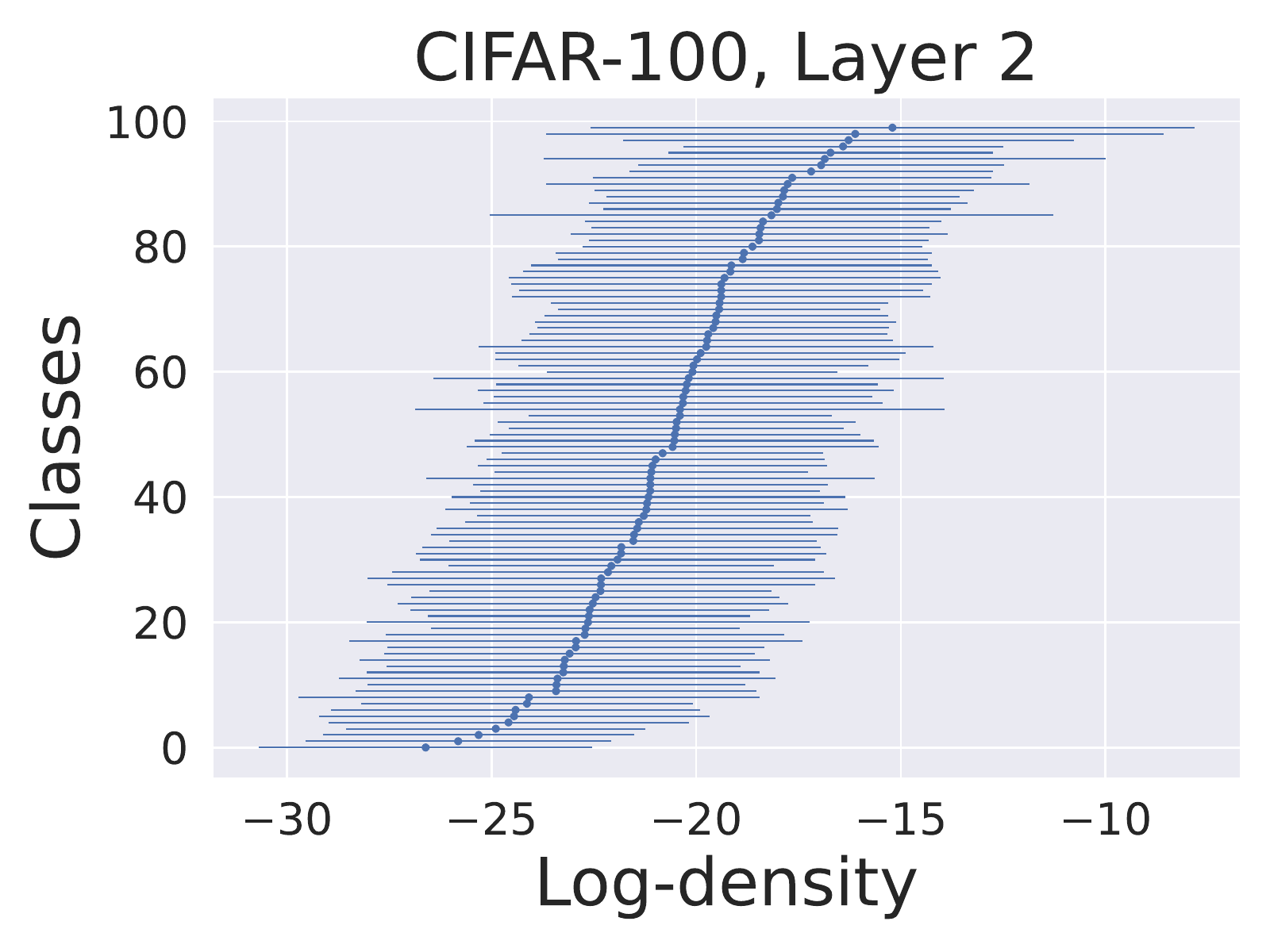}
  \includegraphics[width=0.244\linewidth]{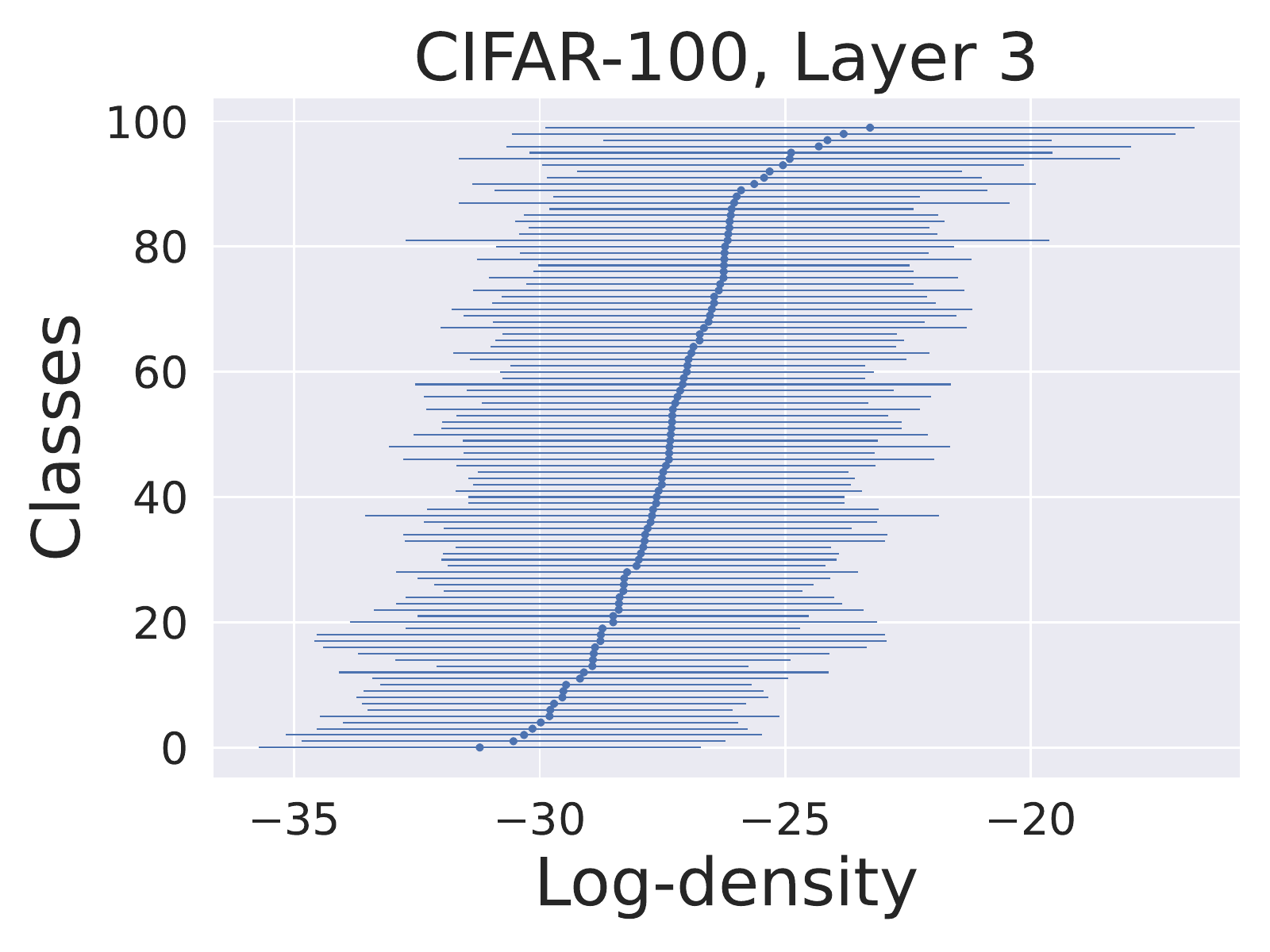}
  \includegraphics[width=0.244\linewidth]{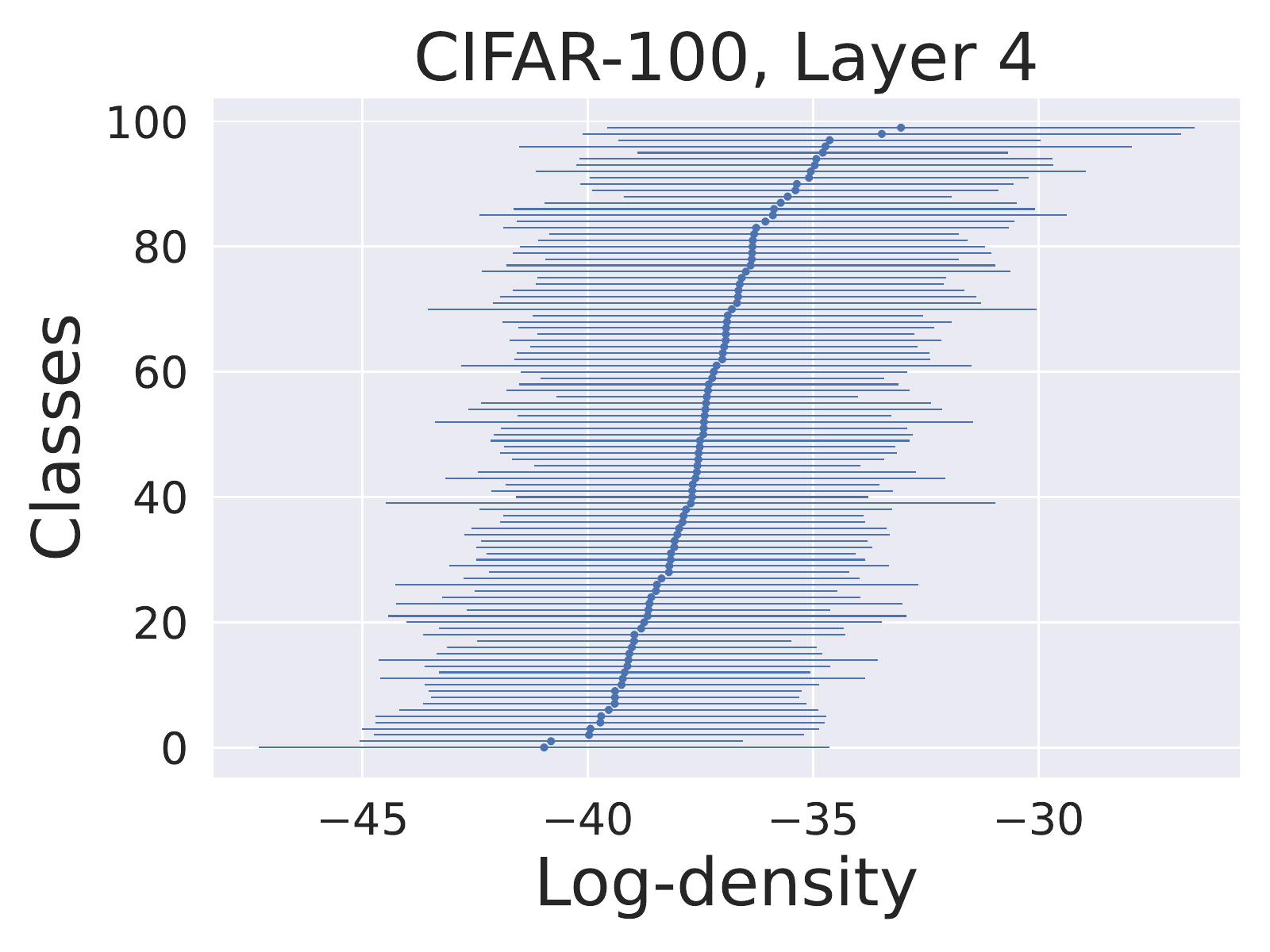}
  \\
  \includegraphics[width=0.244\linewidth]{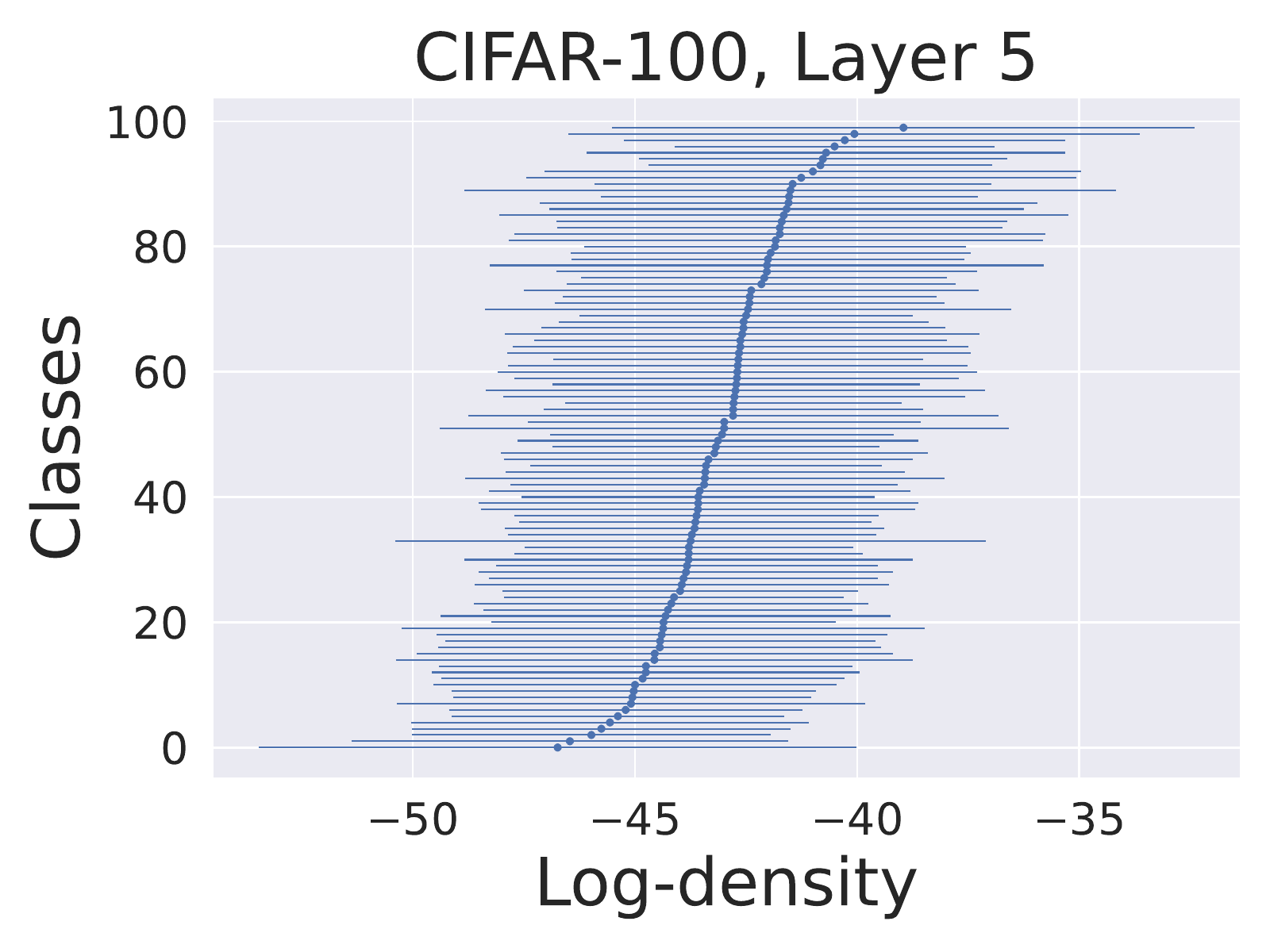}
  \includegraphics[width=0.244\linewidth]{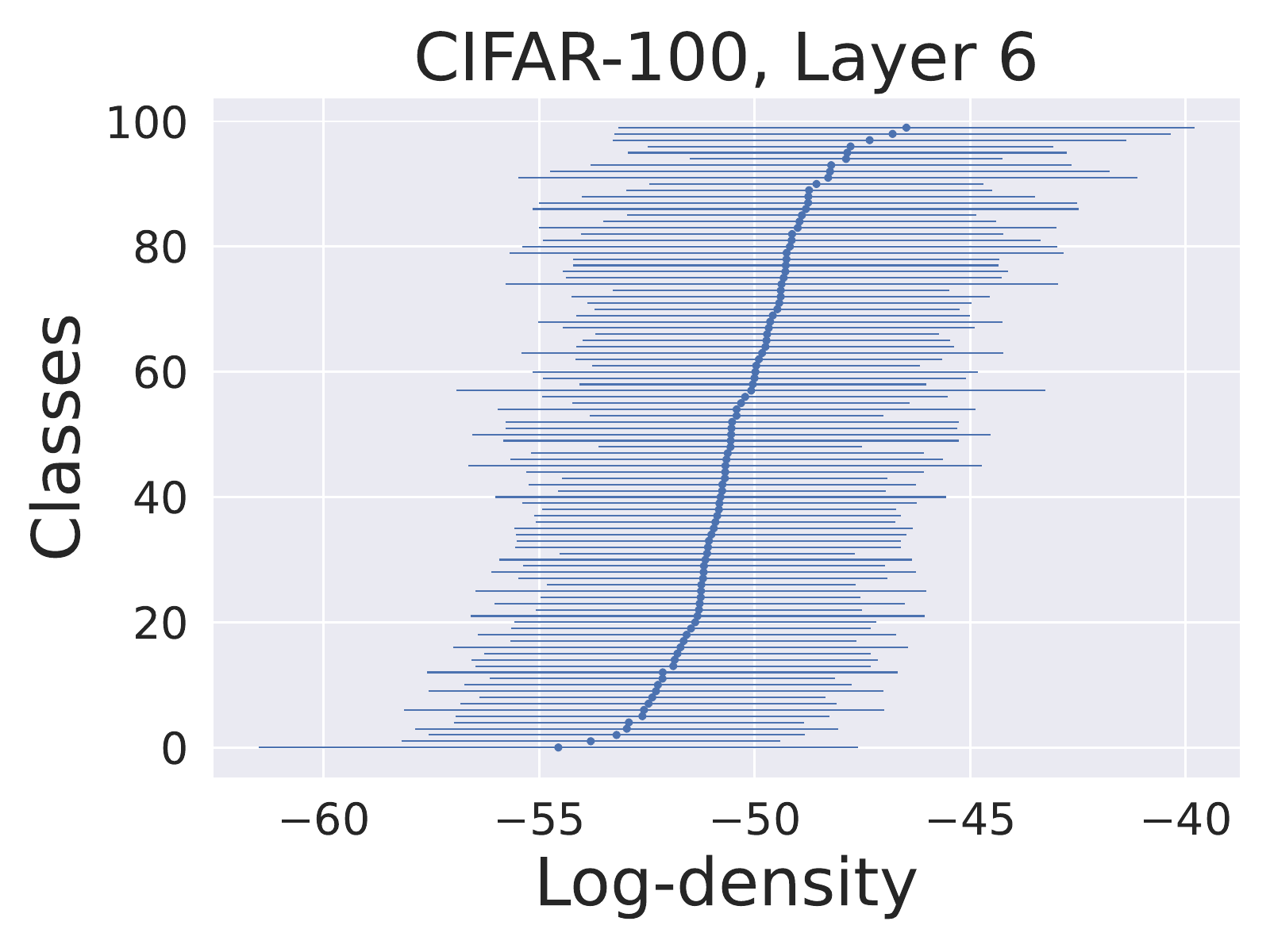}
  \includegraphics[width=0.244\linewidth]{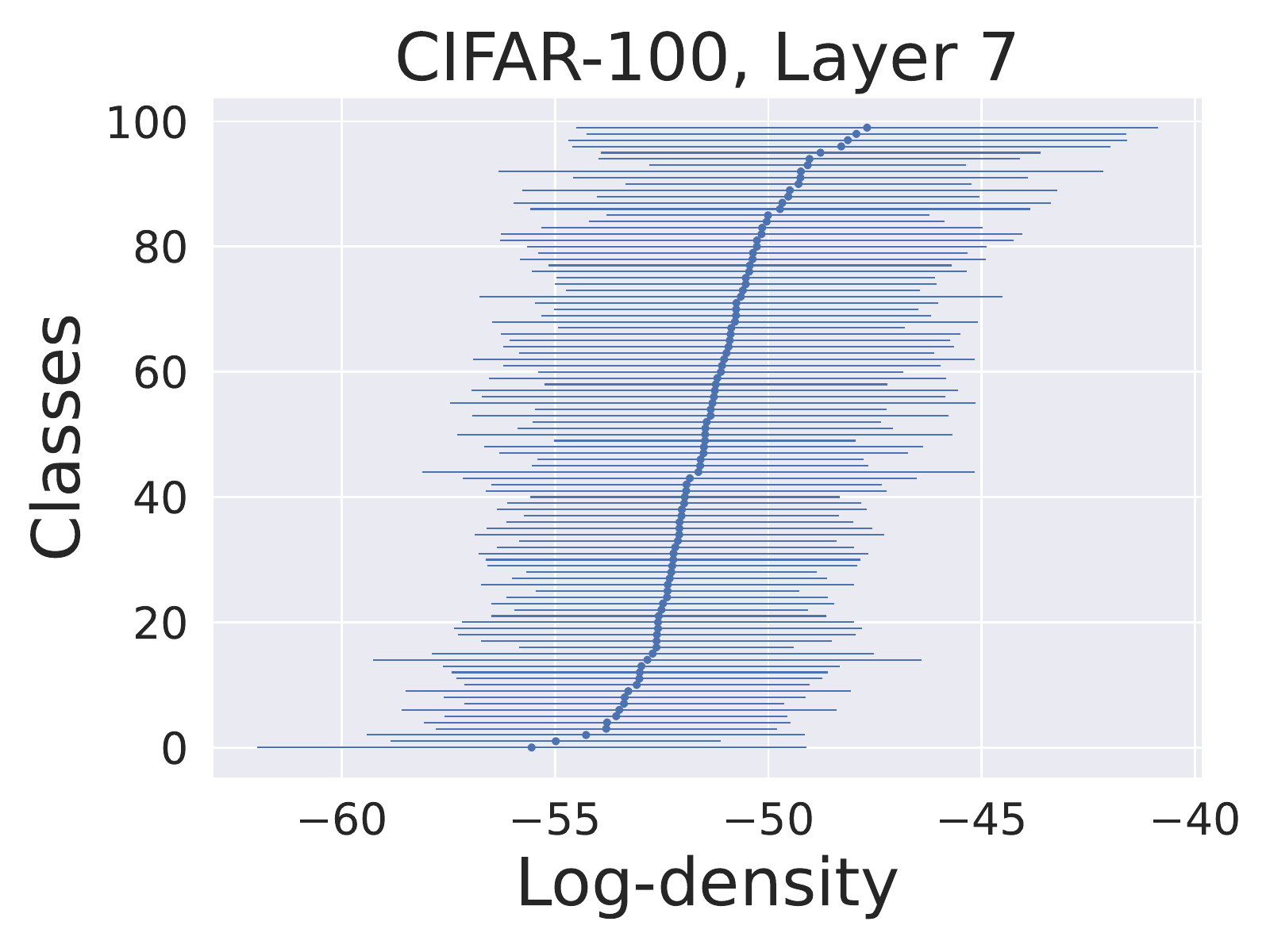}
  \includegraphics[width=0.244\linewidth]{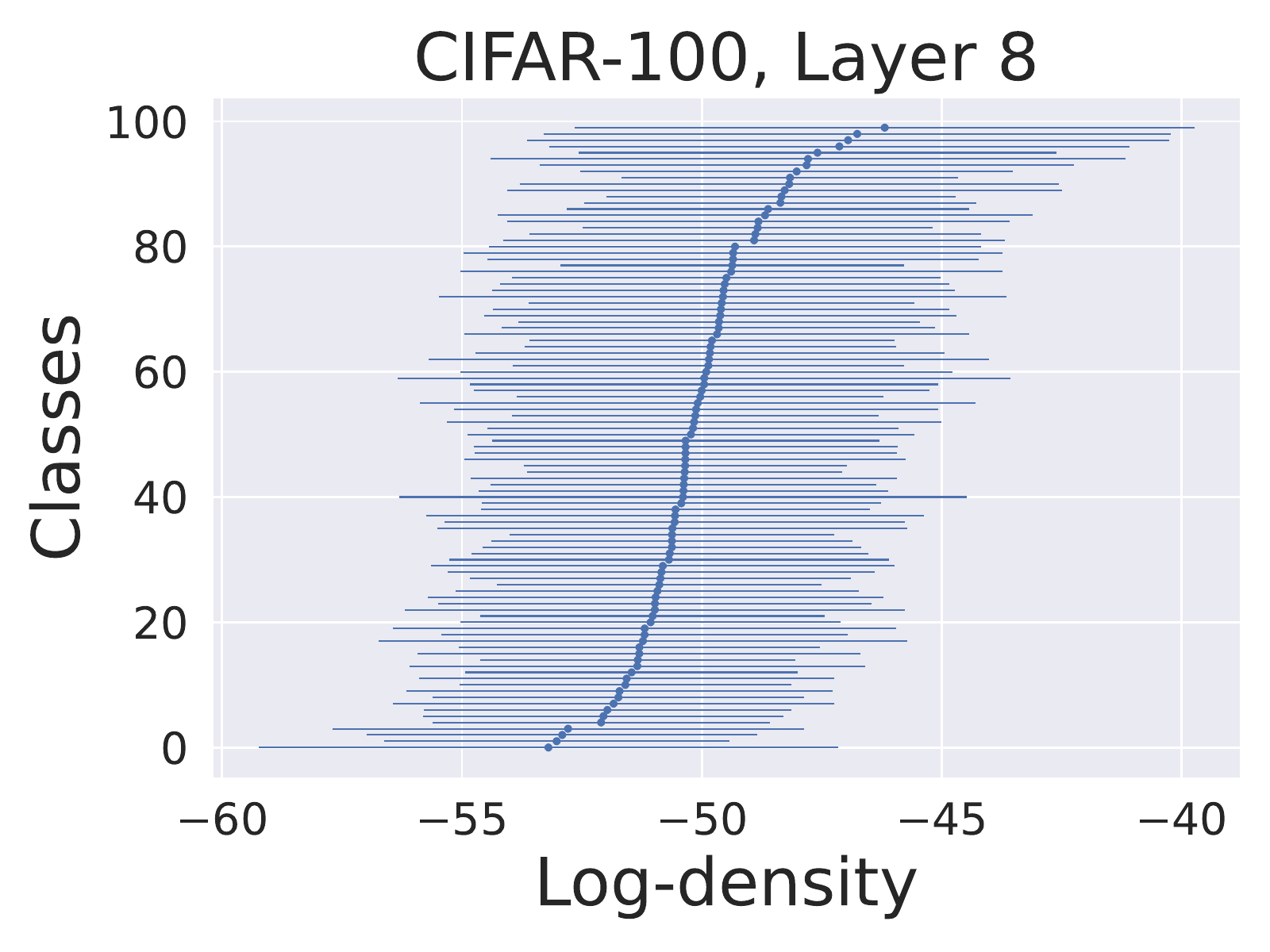}
  \\[1.0em]
  Plain ConvNet, Mini-ImageNet \\[0.25em]
  \includegraphics[width=0.244\linewidth]{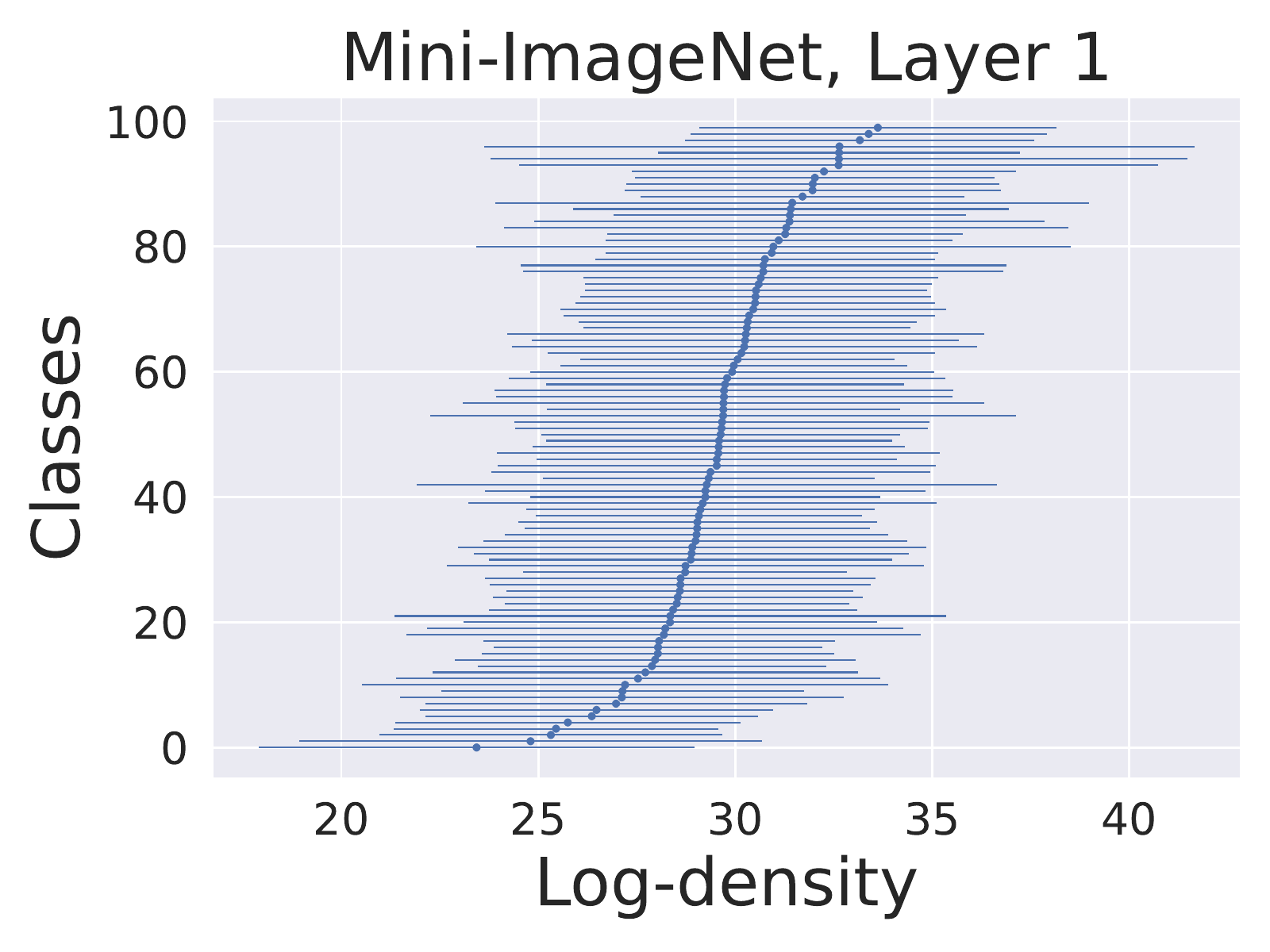}
  \includegraphics[width=0.244\linewidth]{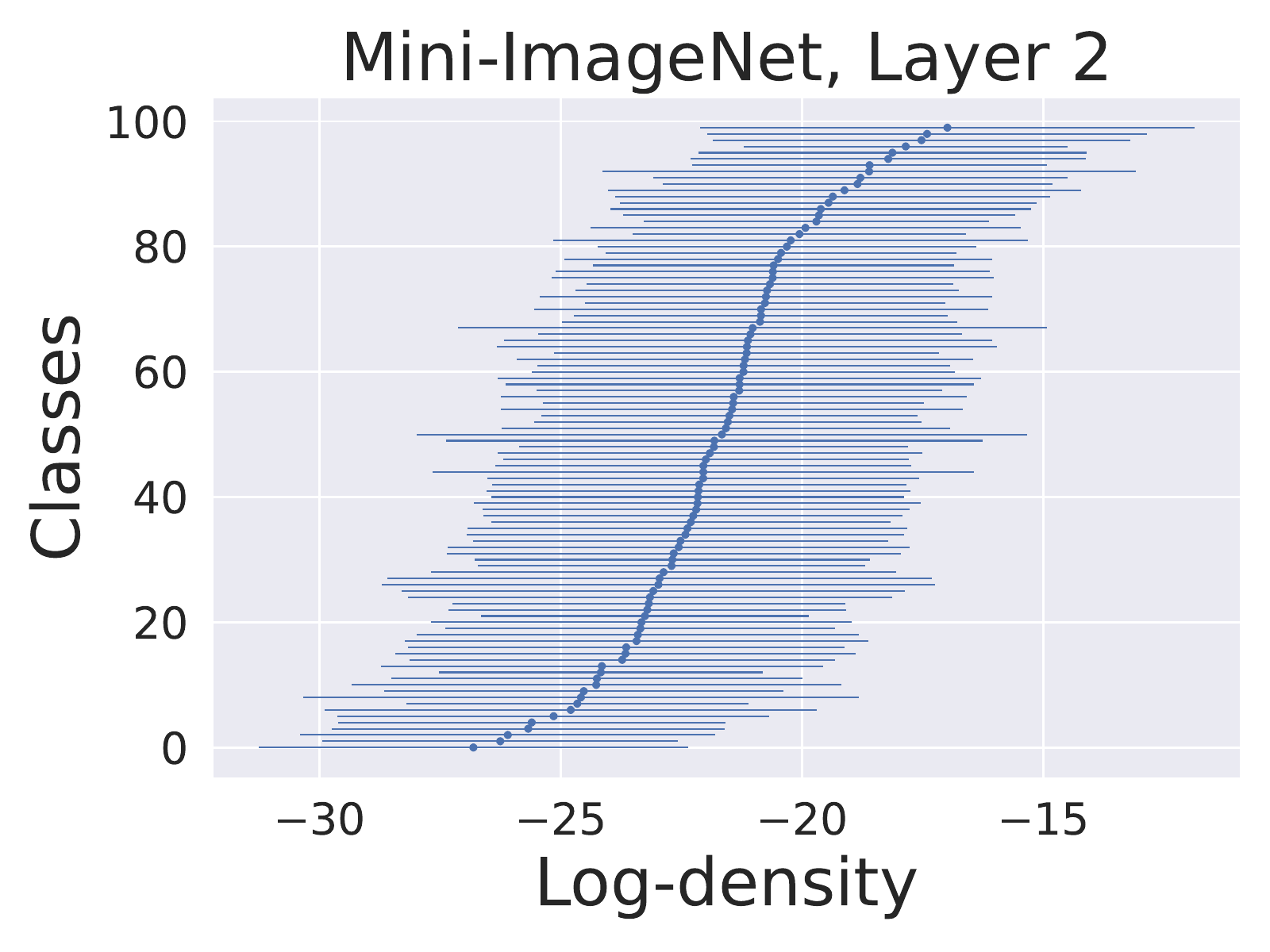}
  \includegraphics[width=0.244\linewidth]{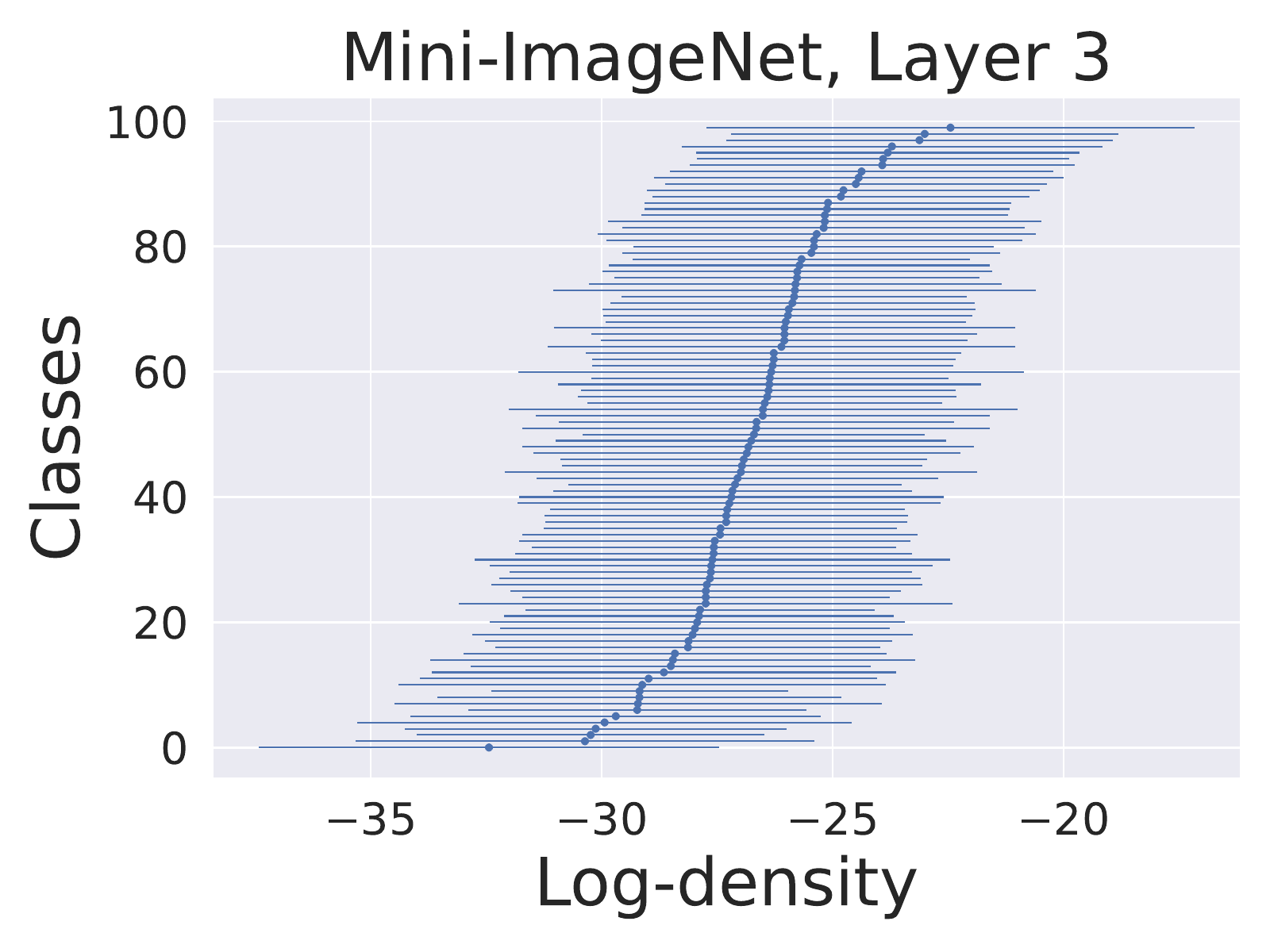}
  \includegraphics[width=0.244\linewidth]{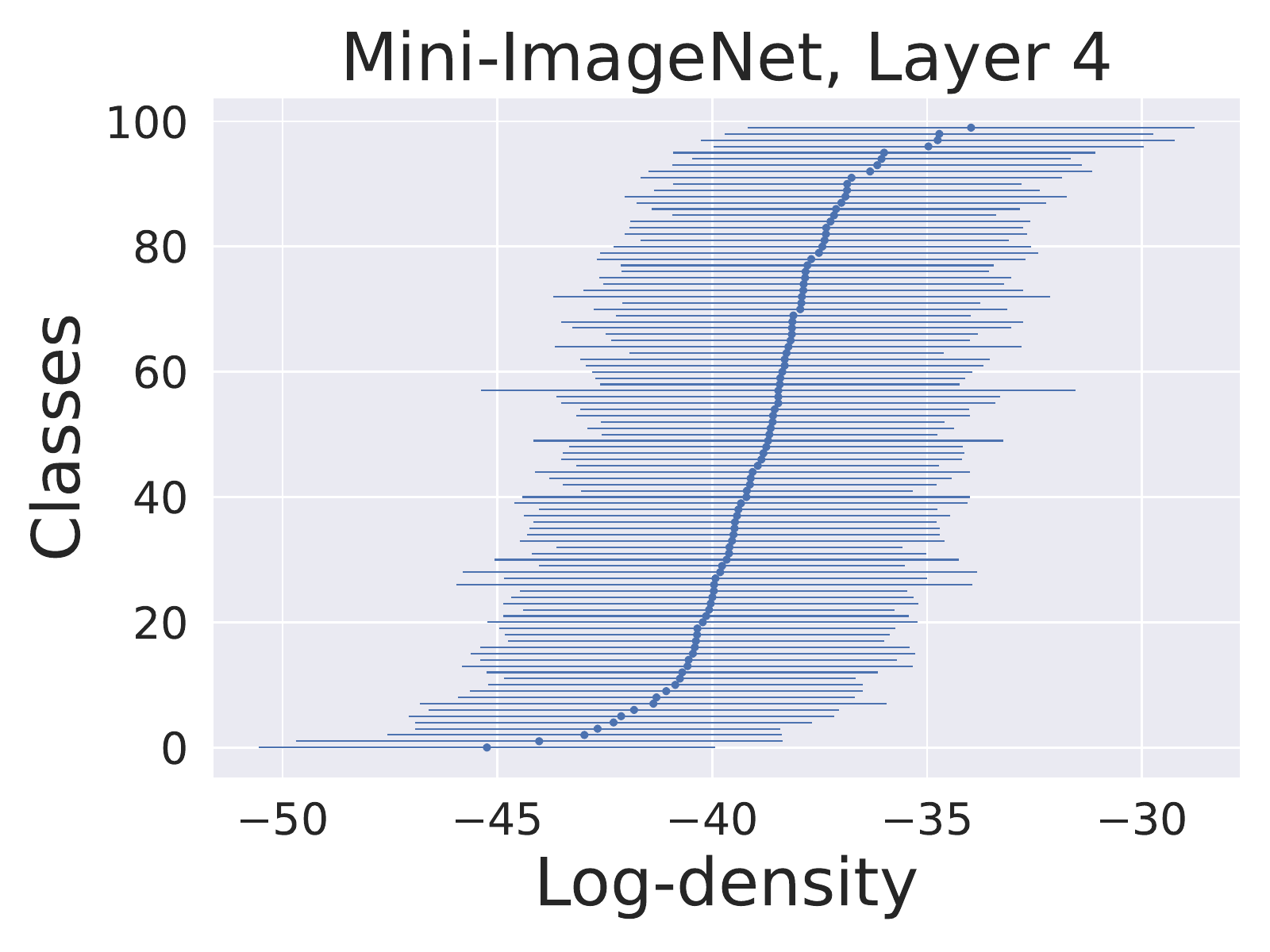}
  \\
  \includegraphics[width=0.244\linewidth]{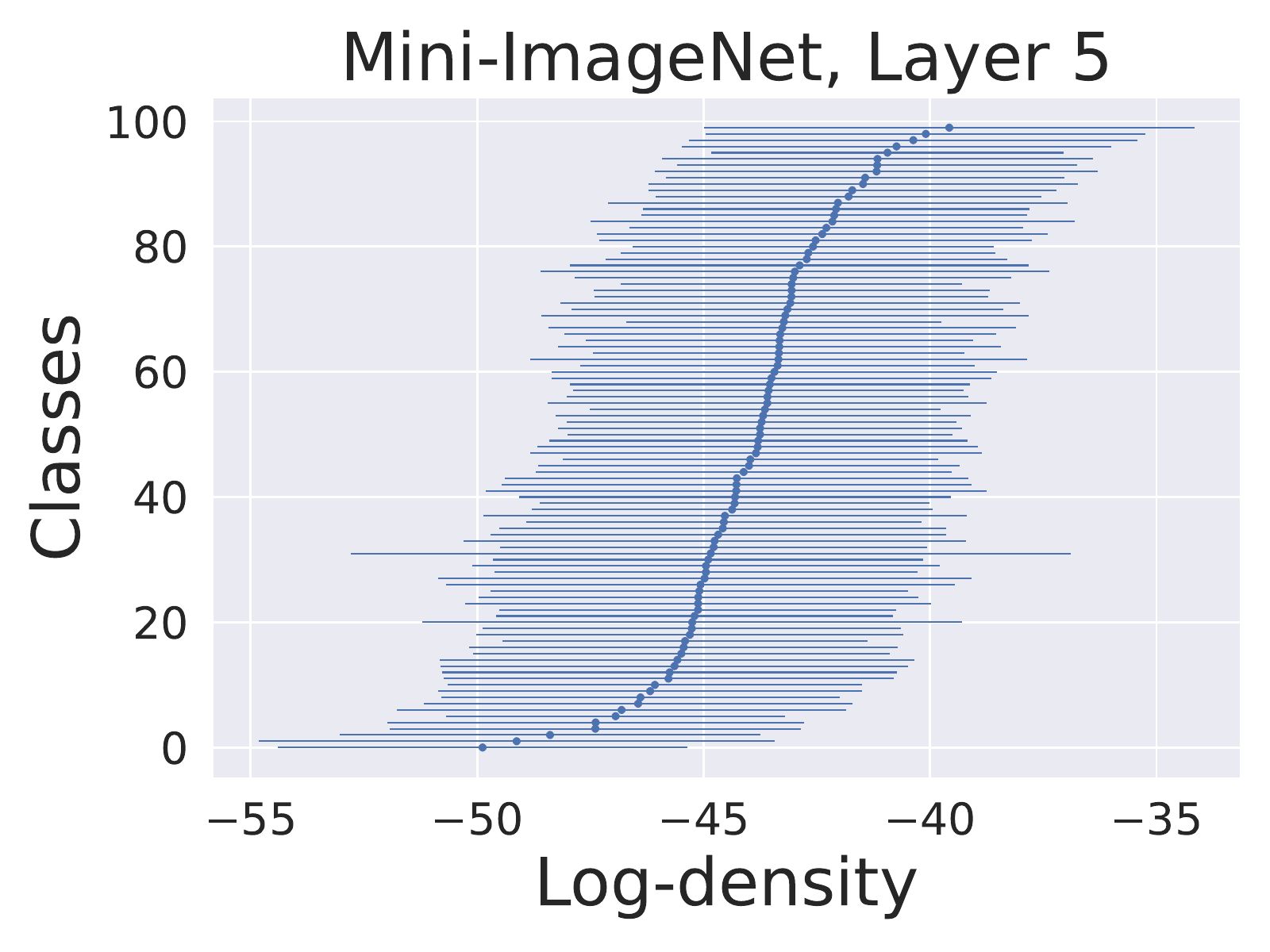}
  \includegraphics[width=0.244\linewidth]{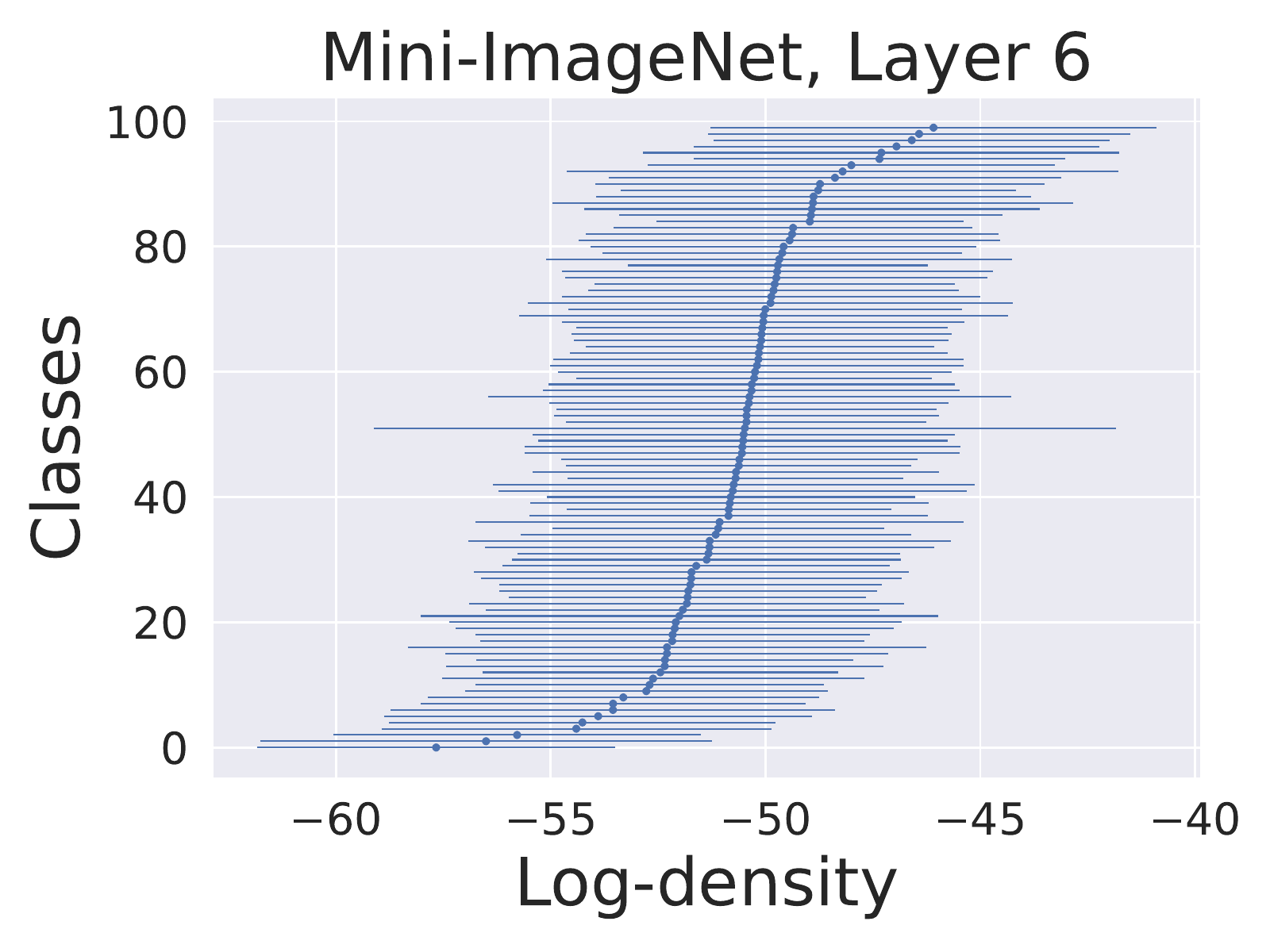}
  \includegraphics[width=0.244\linewidth]{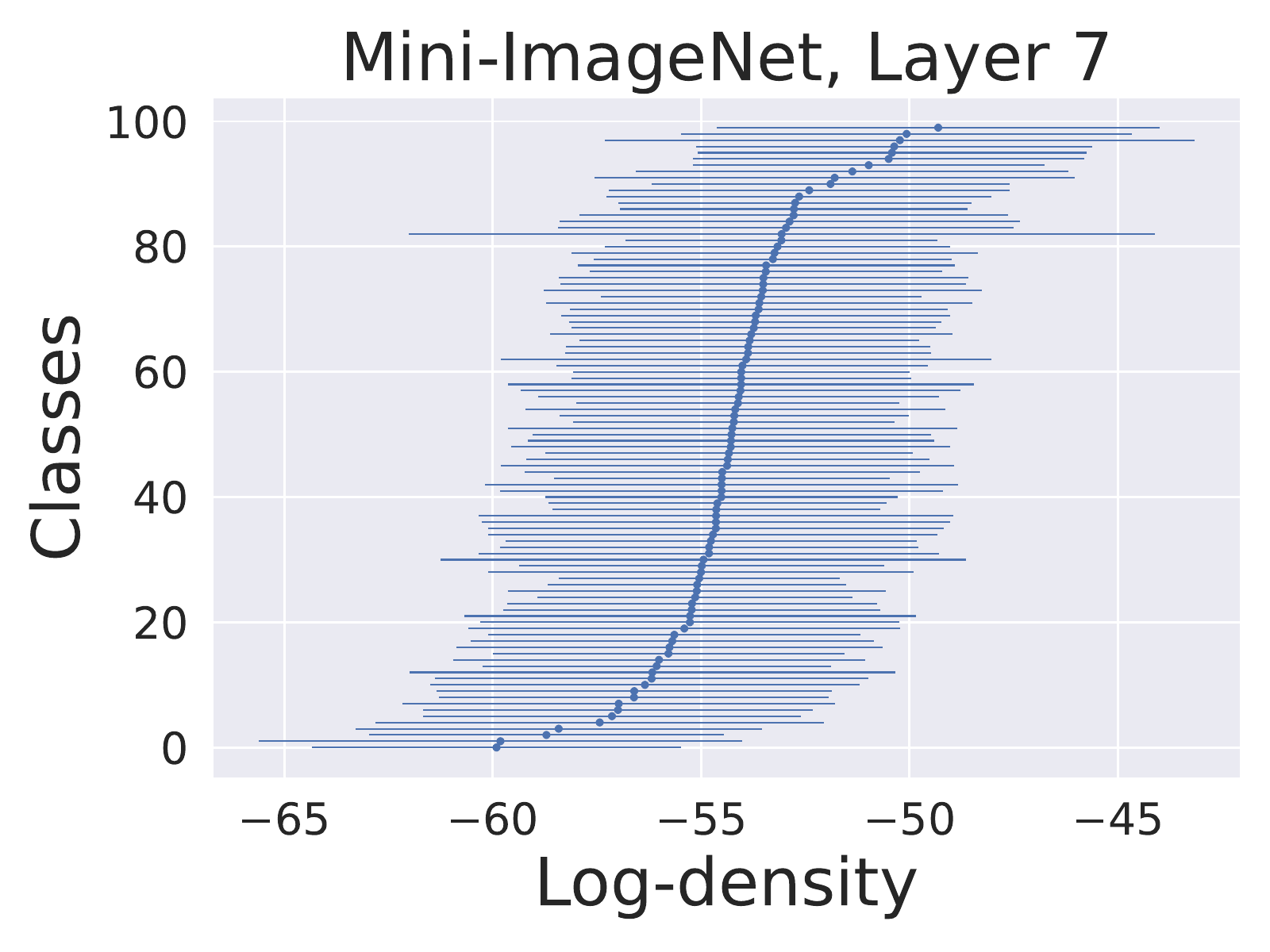}
  \includegraphics[width=0.244\linewidth]{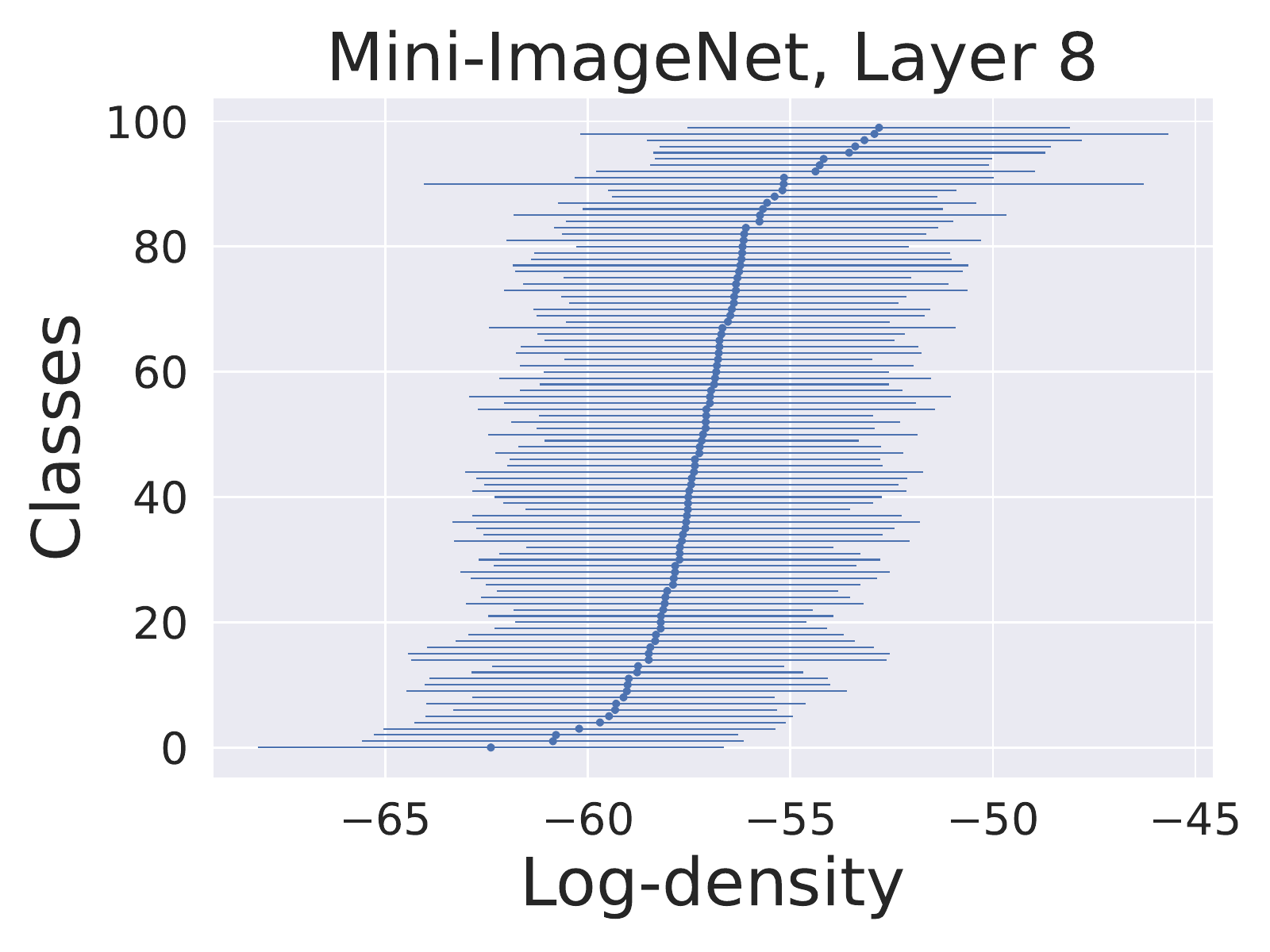}
  \caption{Mean and standard deviation of class-conditional log-densities estimated for representations from a plan
           8-layer convolutional network.}
  \label{fig:log_cc_densities_ccacnn}
\end{figure*}

In the initial two stages of a ResNet model classes are mostly indistinguishable with respect to the class-conditional
log-densities of neural representations (Fig.~\ref{fig:log_cc_densities_all}). However, in the third and fourth stage we
observe two distinct groups of classes that cluster around different mean log-density values. The two groups also differ
in variance of the estimated log-densities. In principle, this structure could be simply a product of the datasets used
in experiments. However, our results show that this is not the case. Specifically, we replicated our experiments using a
plain (i.e. non-residual) 8-layer convolutional network. This model fits classes in an uniform way across all hidden
layers (Fig.~\ref{fig:log_cc_densities_ccacnn}).

Our results suggests that classes with high mean log-density values are formed by collections of compact but spatially
separated components. This raises a following question: is this component structure observed in the posterior over
parameters of the model in Eq.~\eqref{eq:dpgmm}? We do observe that posterior samples in the CGS chains for the
high-density classes involve many more components than samples for the low-density classes. However, we do not report
these numbers, as estimation of component counts is fragile. Concretely, the model in Eq.~\eqref{eq:dpgmm} is an
infinite mixture.  As such, it is consistent for the density, but is not consistent for the number of
components~\cite{Miller14}.  While there are mixture models consistent for the number of components~\citep{Miller18},
they require knowledge of the components' distribution. Any misspecification of this base distribution would lead to
incorrect estimates for component counts.

%% file: appendix_vit_mixer.tex
The striking difference between class representations in ResNets and plain convolutional networks suggests that other
architectures with residual connections may also display non-trivial structure in neural representations. Here we report
our initial findings for two such architectures, namely Vision Transformer (ViT)~\citep{Dosovitskiy2021} and
MLP-Mixer~\citep{Tolstikhin2021}. Specifically, we took the ViT-B/16 and Mixer-B/16 models\footnote{Available at:
\url{https://github.com/google-research/vision_transformer}} pre-trained on the ImageNet dataset with Sharpness-aware
Minimization~\citep{Foret2021} and fine-tuned them to CIFAR100 and Mini-ImageNet classes. Vision Transformer models
typically contain a separate class token. To form class representations for the ViT-B/16 network we therefore fitted
density models to class token activations extracted at the end of each encoder block. MLP-Mixer models construct input
to the classification head by global average pooling of activation maps. Consequently, in this case we fitted density
models to activations formed by global average pooling of each mixer block output. When fitting density models we
followed the procedure used for ResNets, including dimensionality reduction and a weakly-informative prior distribution.

\begin{figure*}[htb]
  \centering
  Vision Transformer, CIFAR100 \\[0.25em]
  \includegraphics[width=0.244\linewidth]{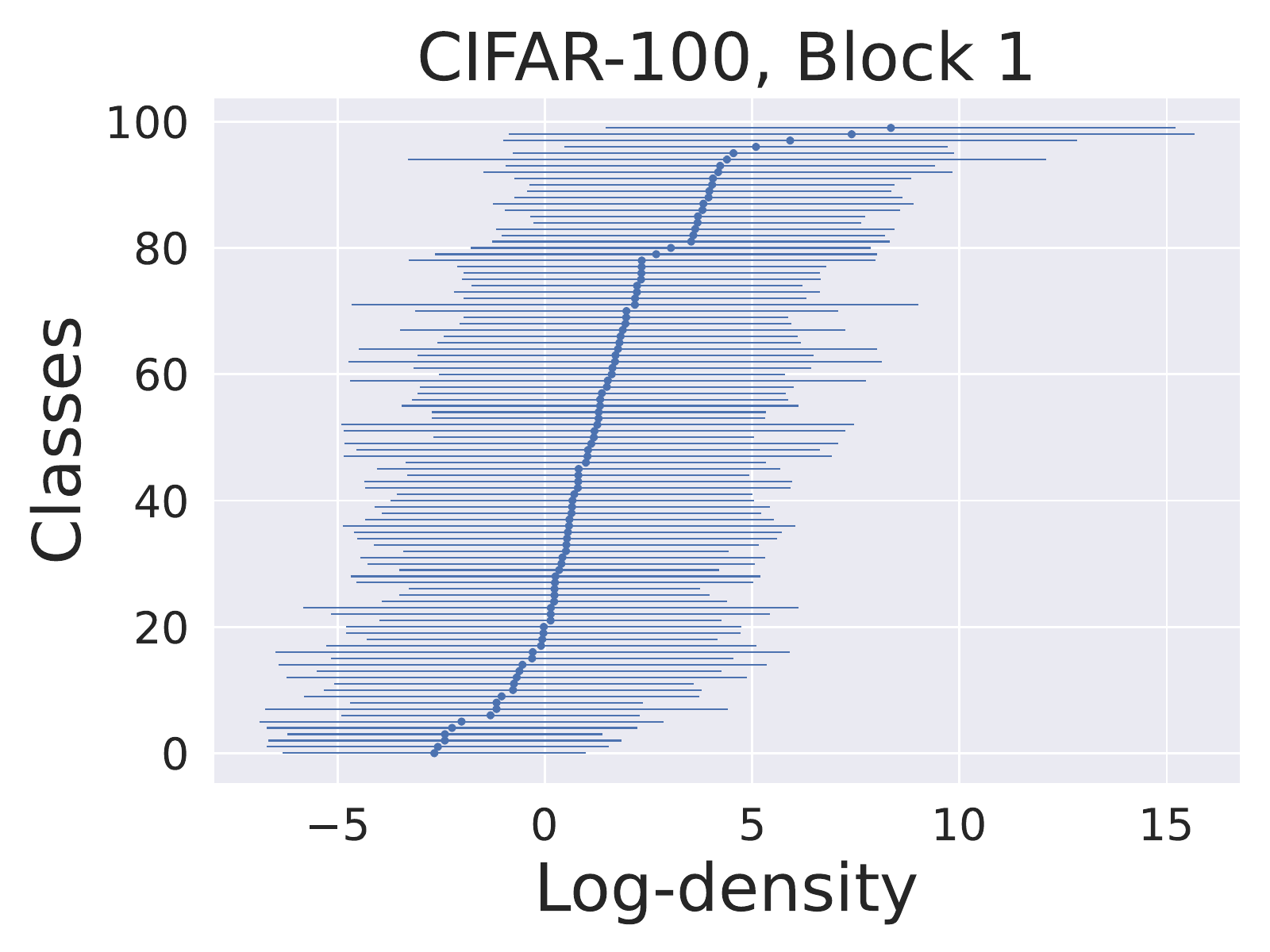}
  \includegraphics[width=0.244\linewidth]{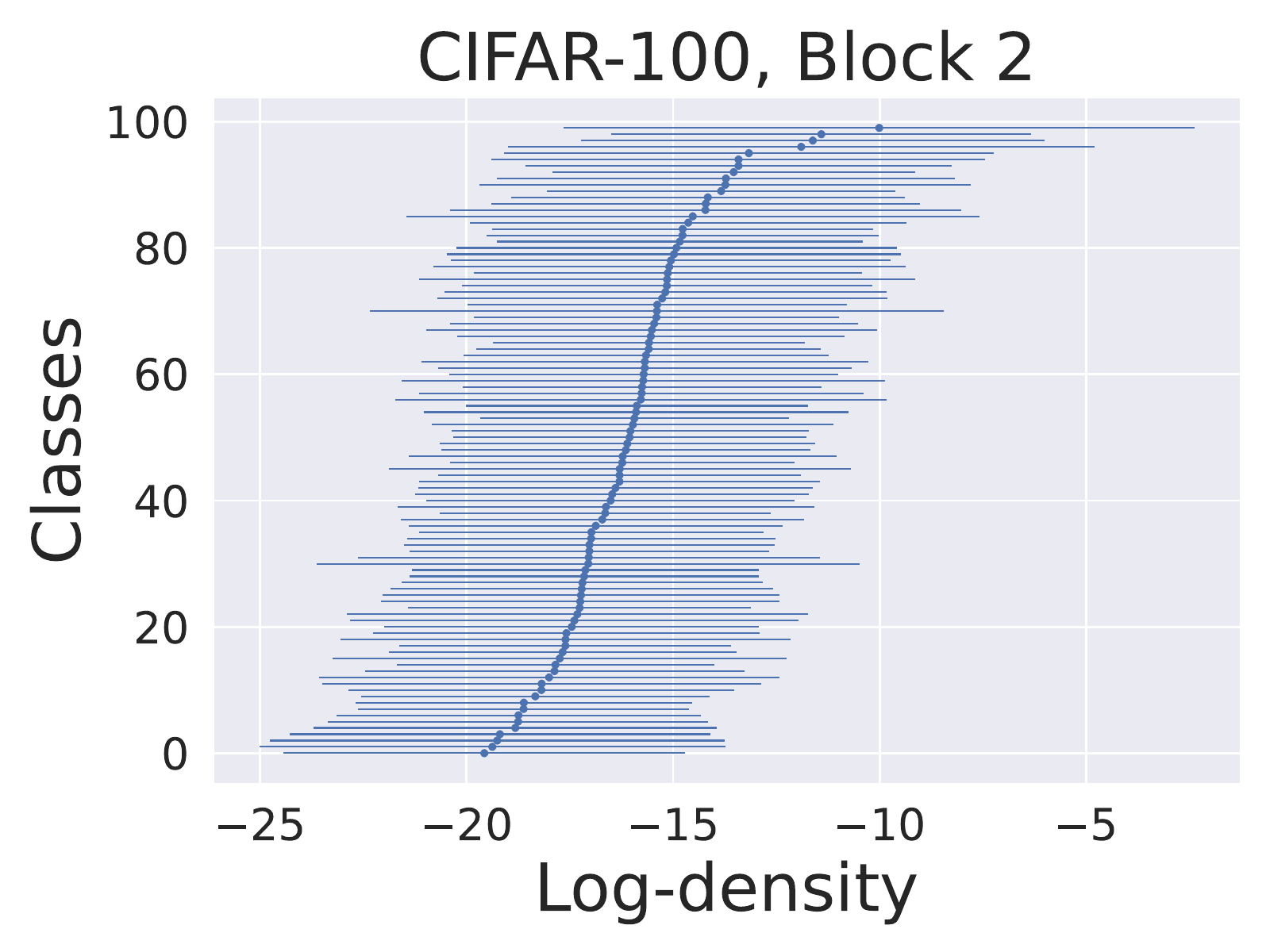}
  \includegraphics[width=0.244\linewidth]{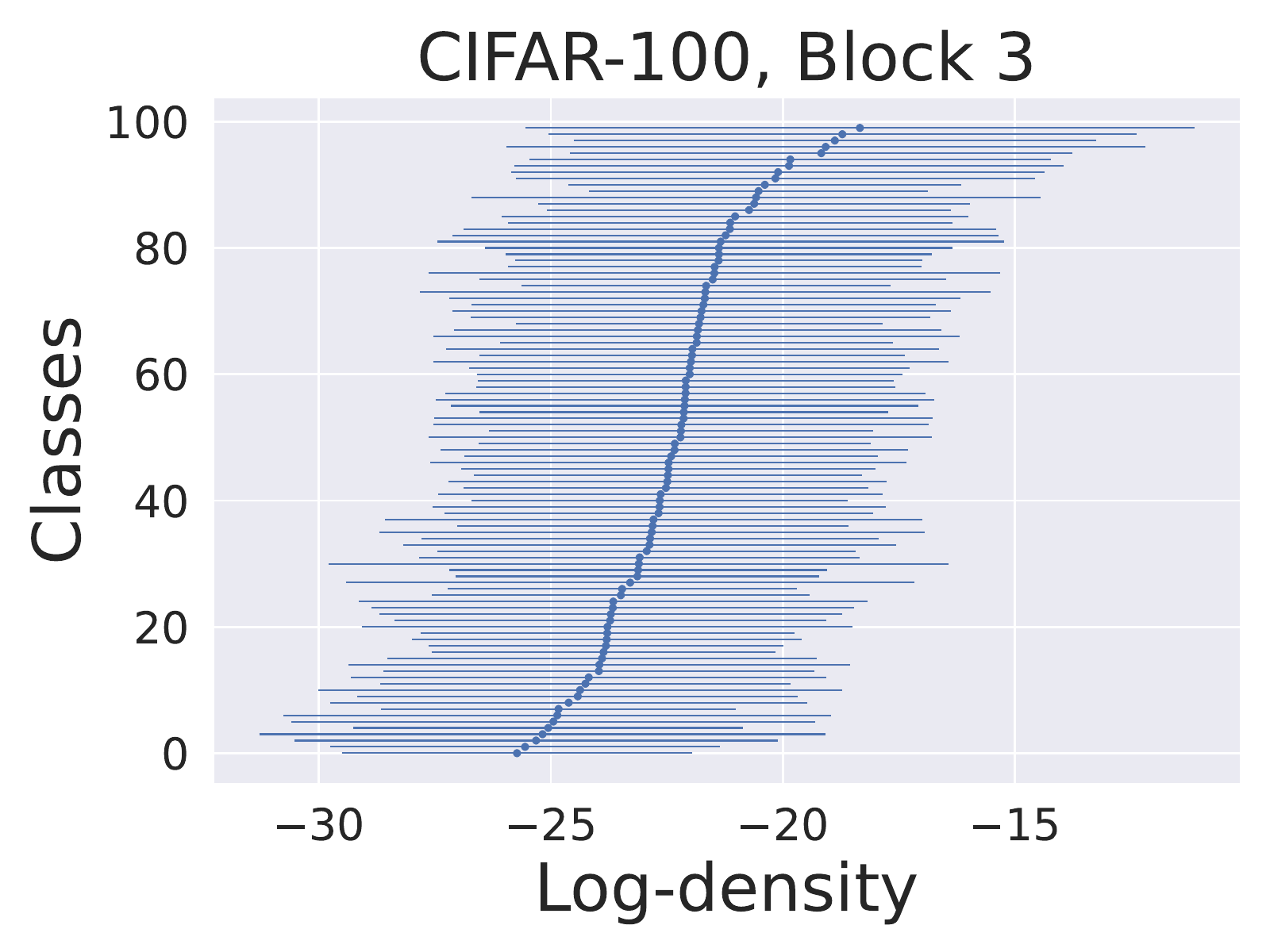}
  \includegraphics[width=0.244\linewidth]{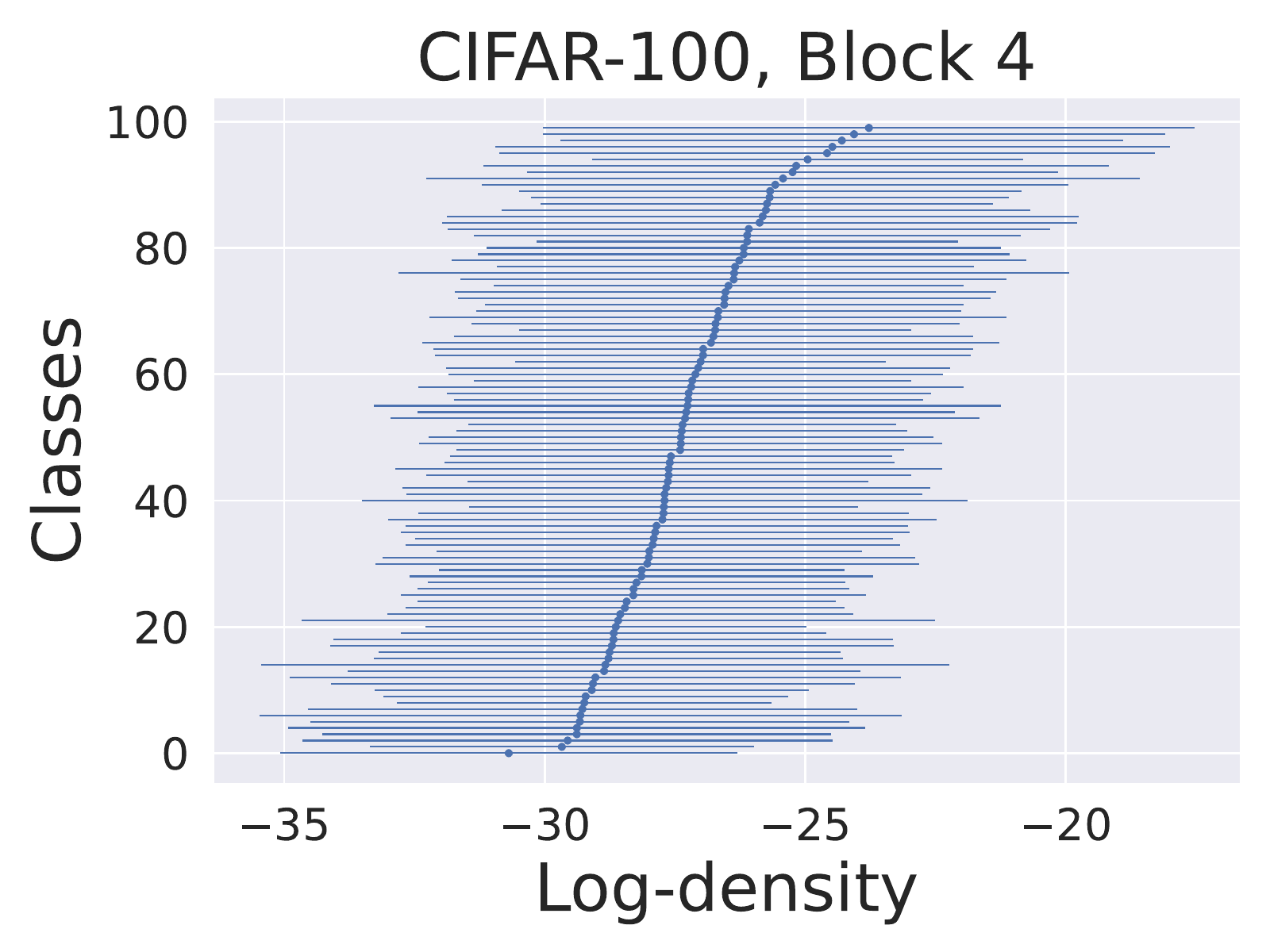}
  \\
  \includegraphics[width=0.244\linewidth]{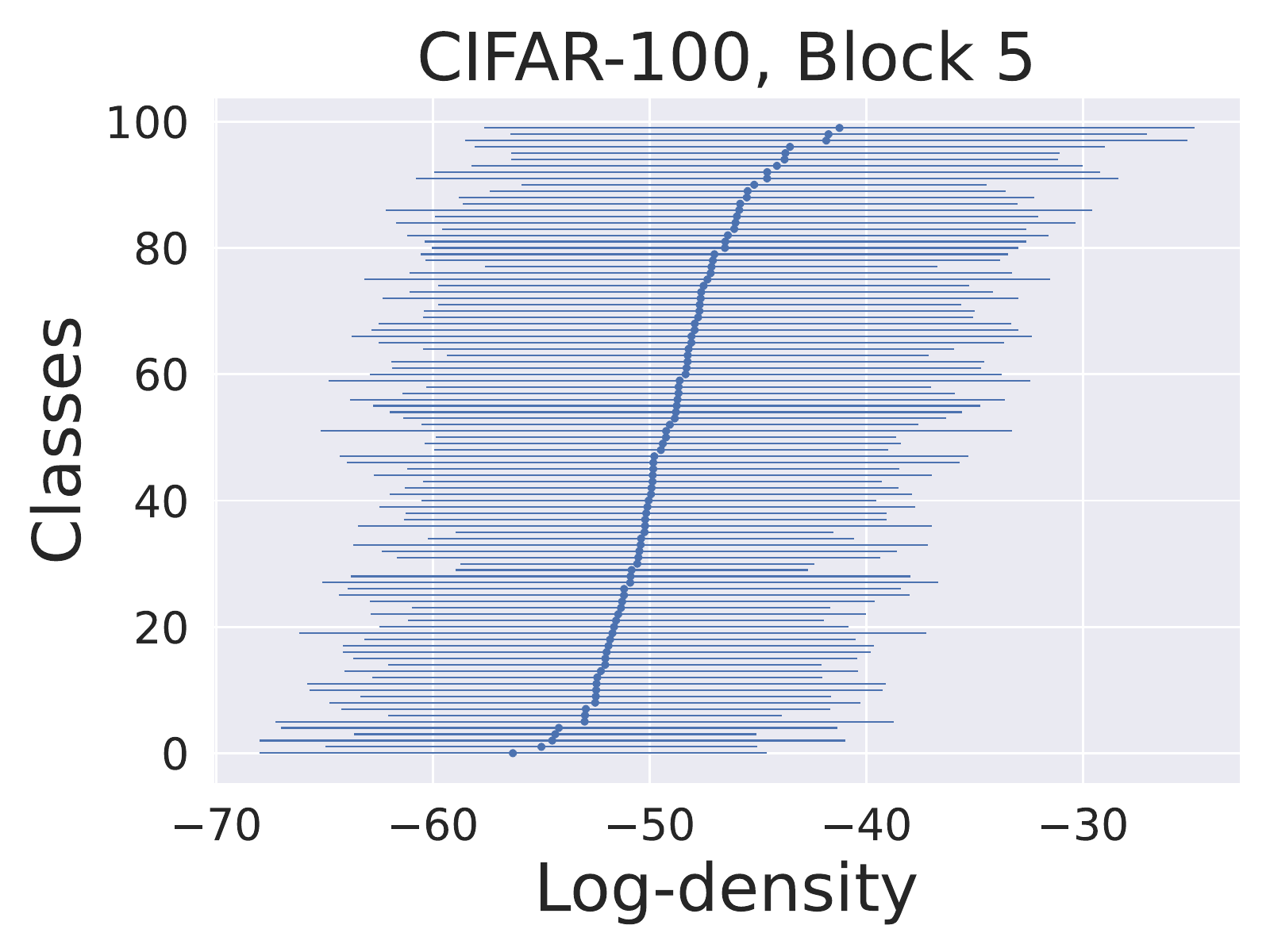}
  \includegraphics[width=0.244\linewidth]{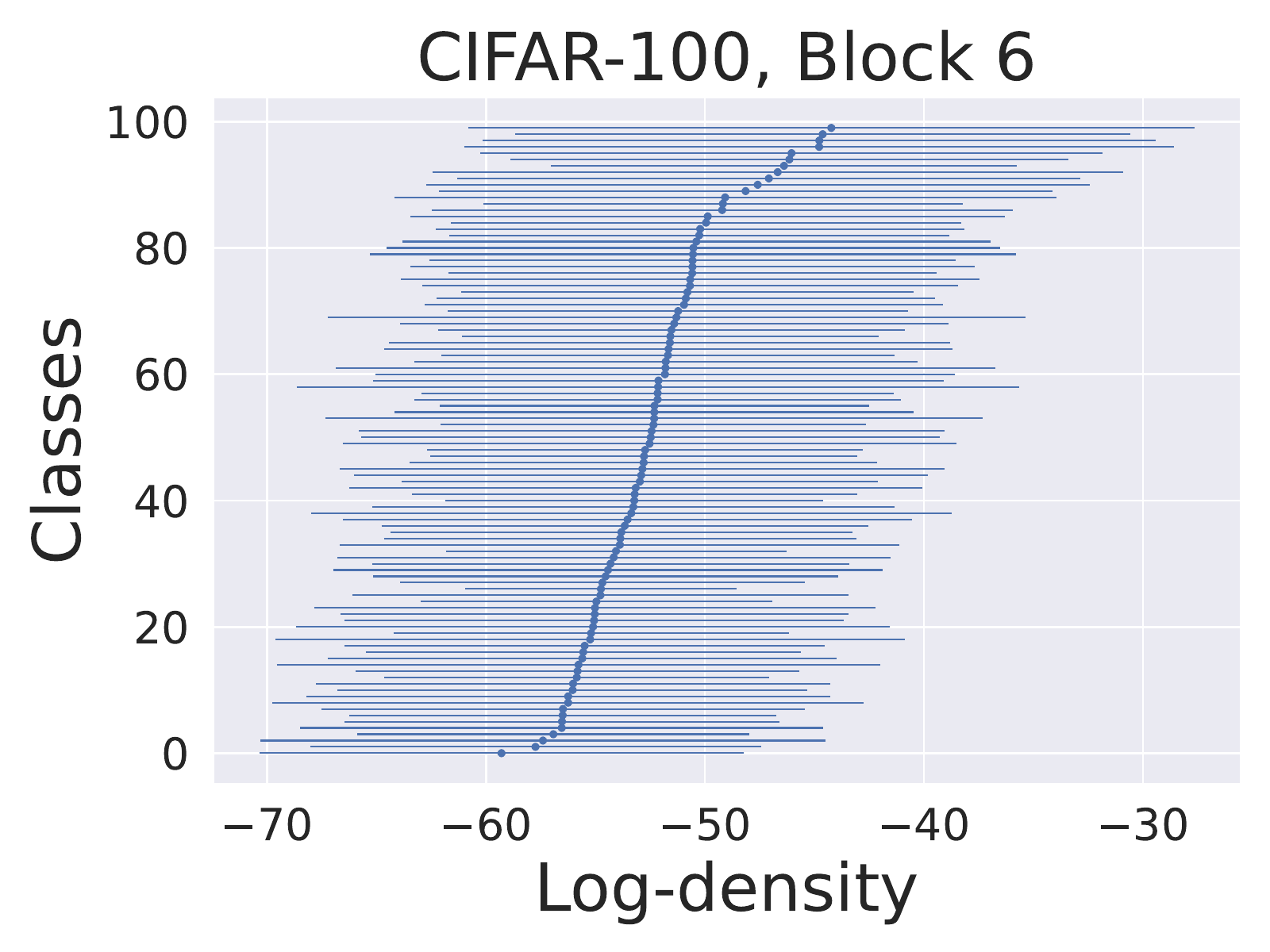}
  \includegraphics[width=0.244\linewidth]{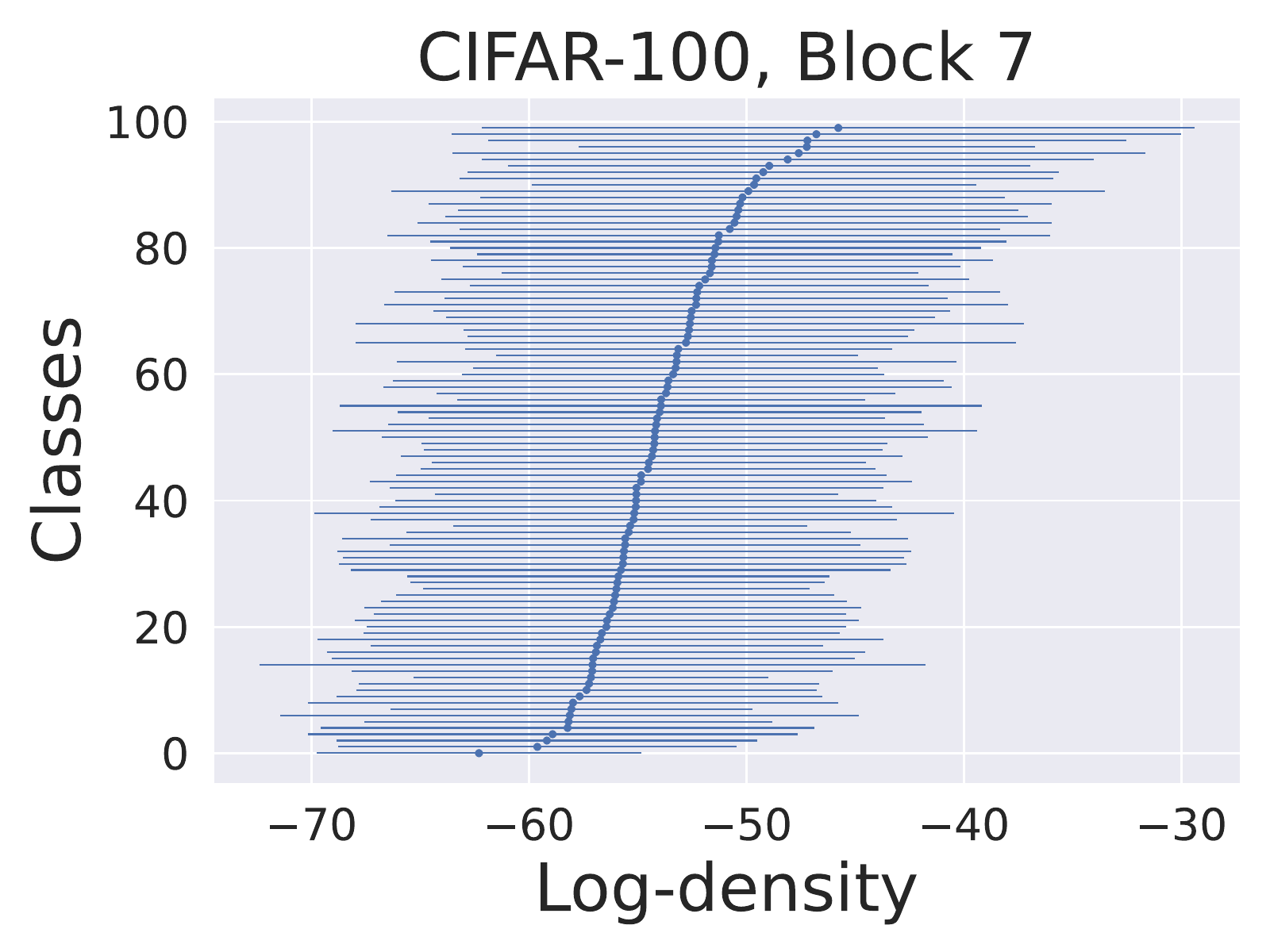}
  \includegraphics[width=0.244\linewidth]{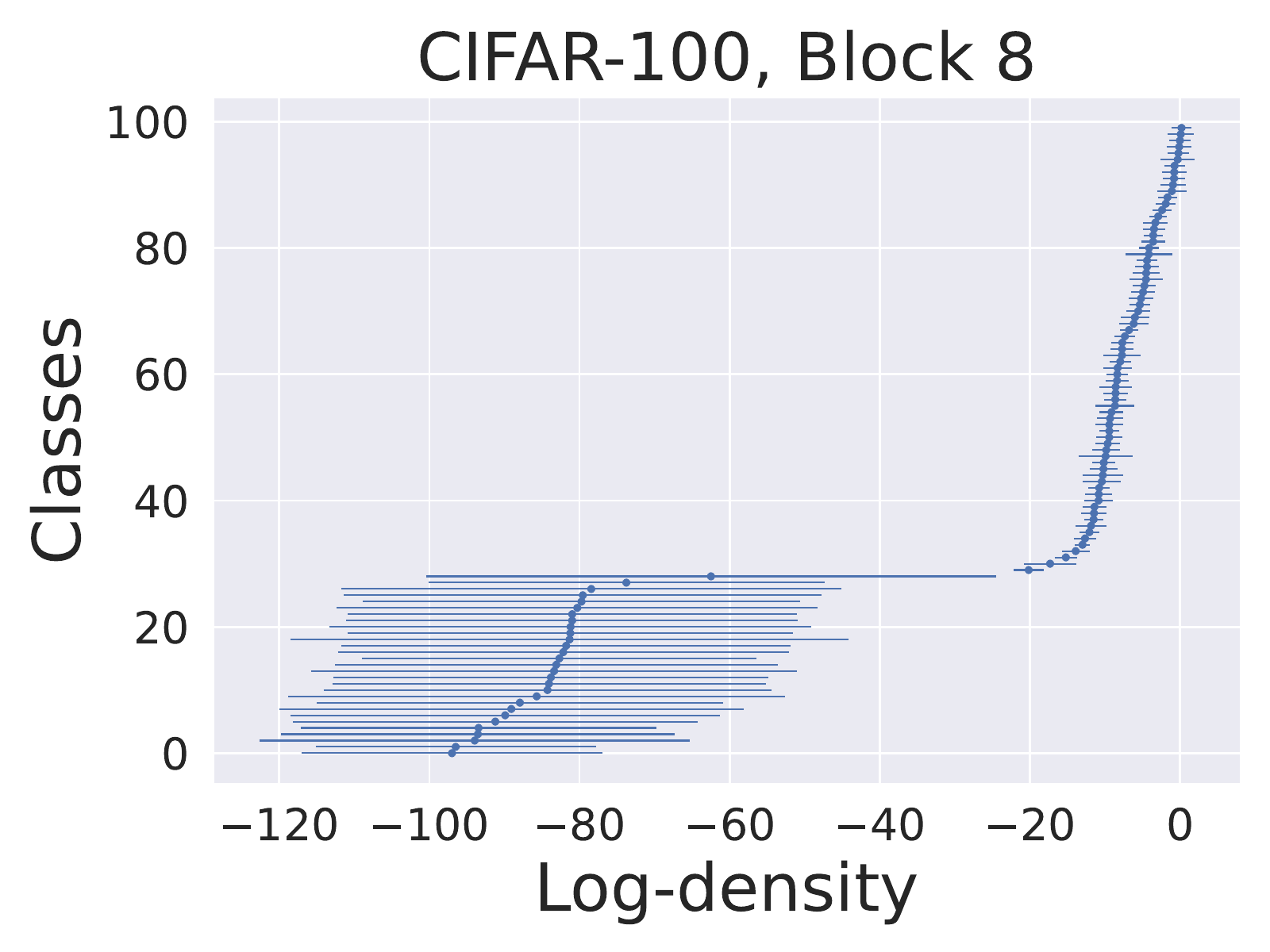}
  \\
  \includegraphics[width=0.244\linewidth]{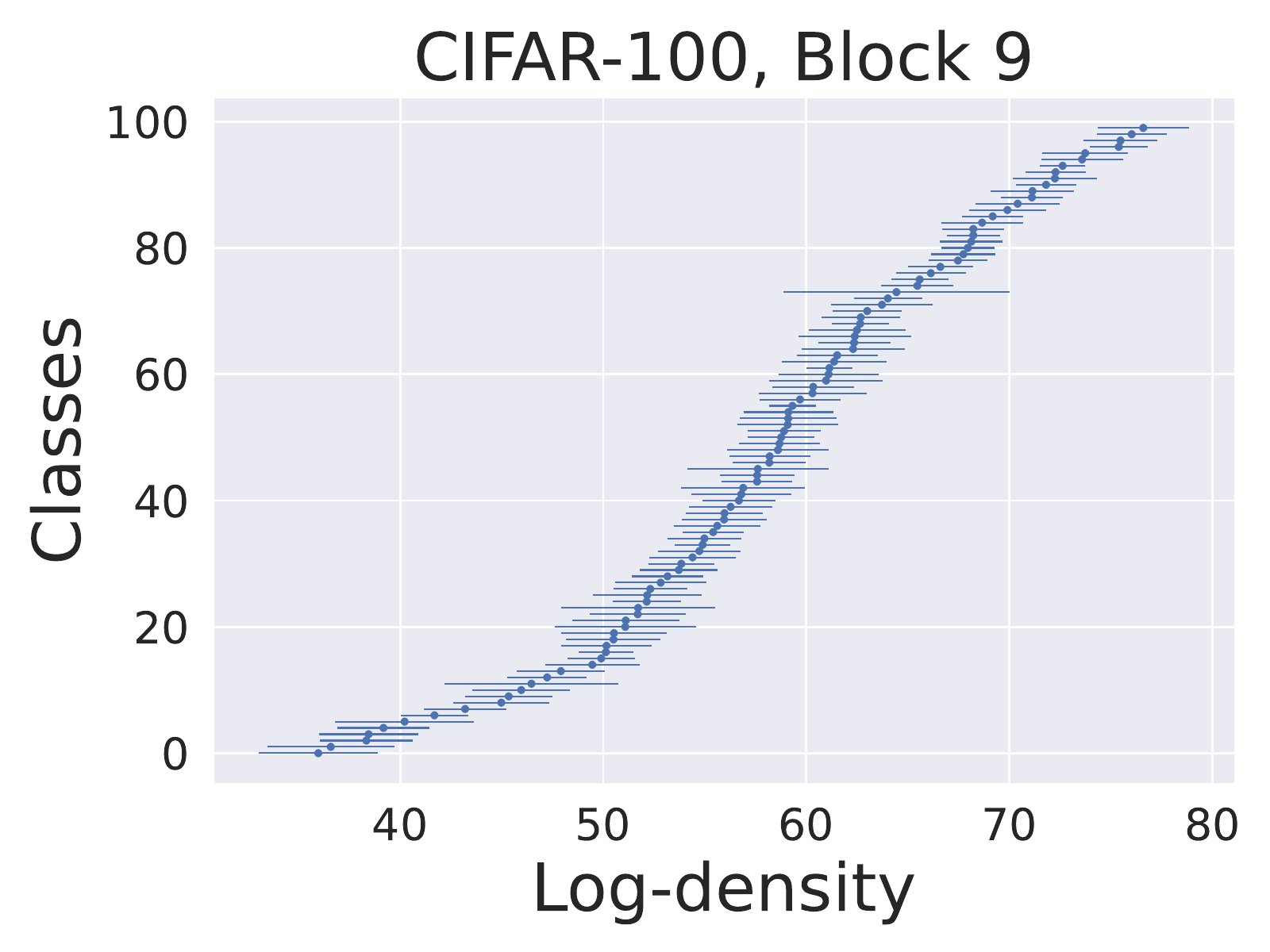}
  \includegraphics[width=0.244\linewidth]{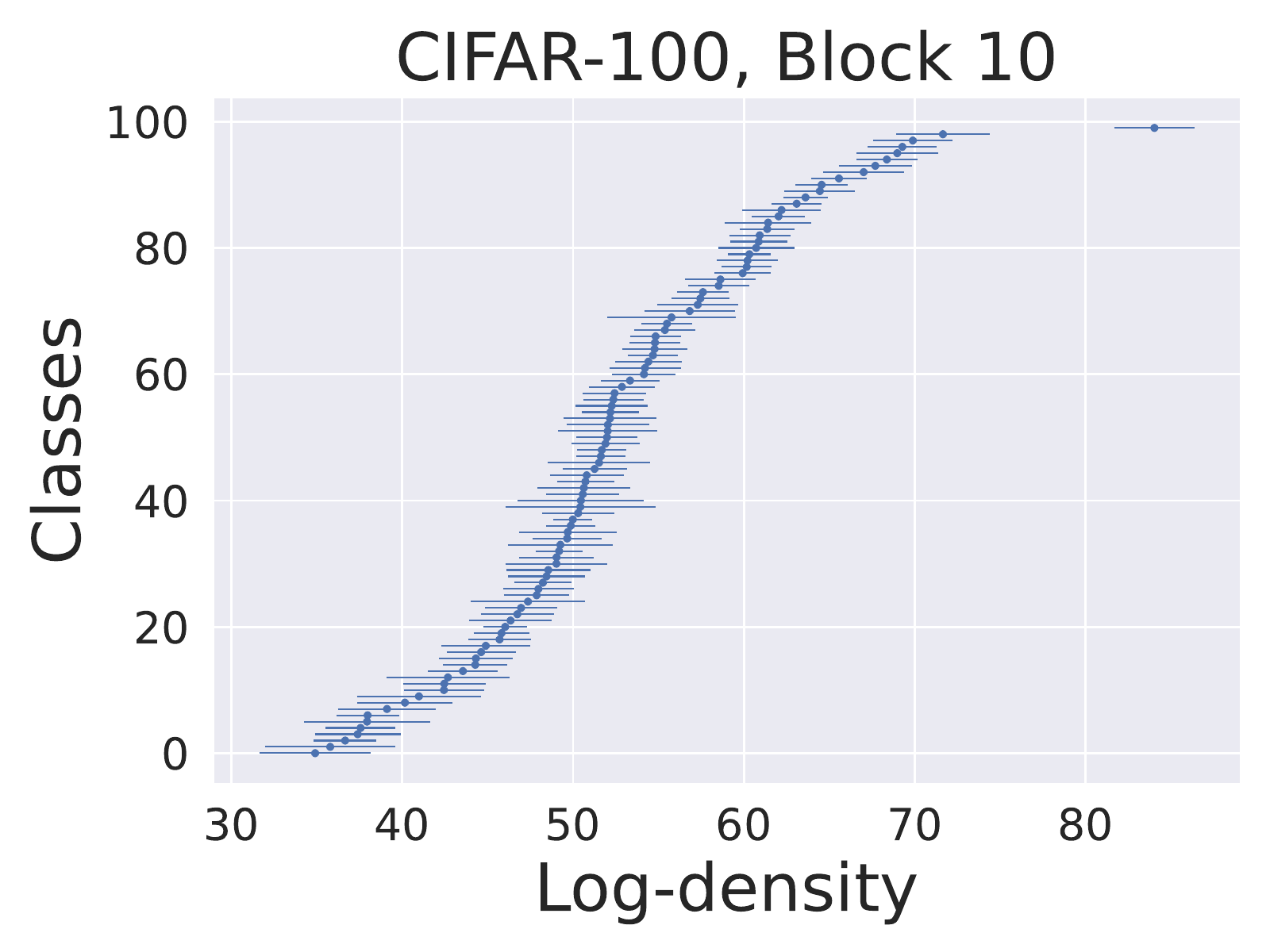}
  \includegraphics[width=0.244\linewidth]{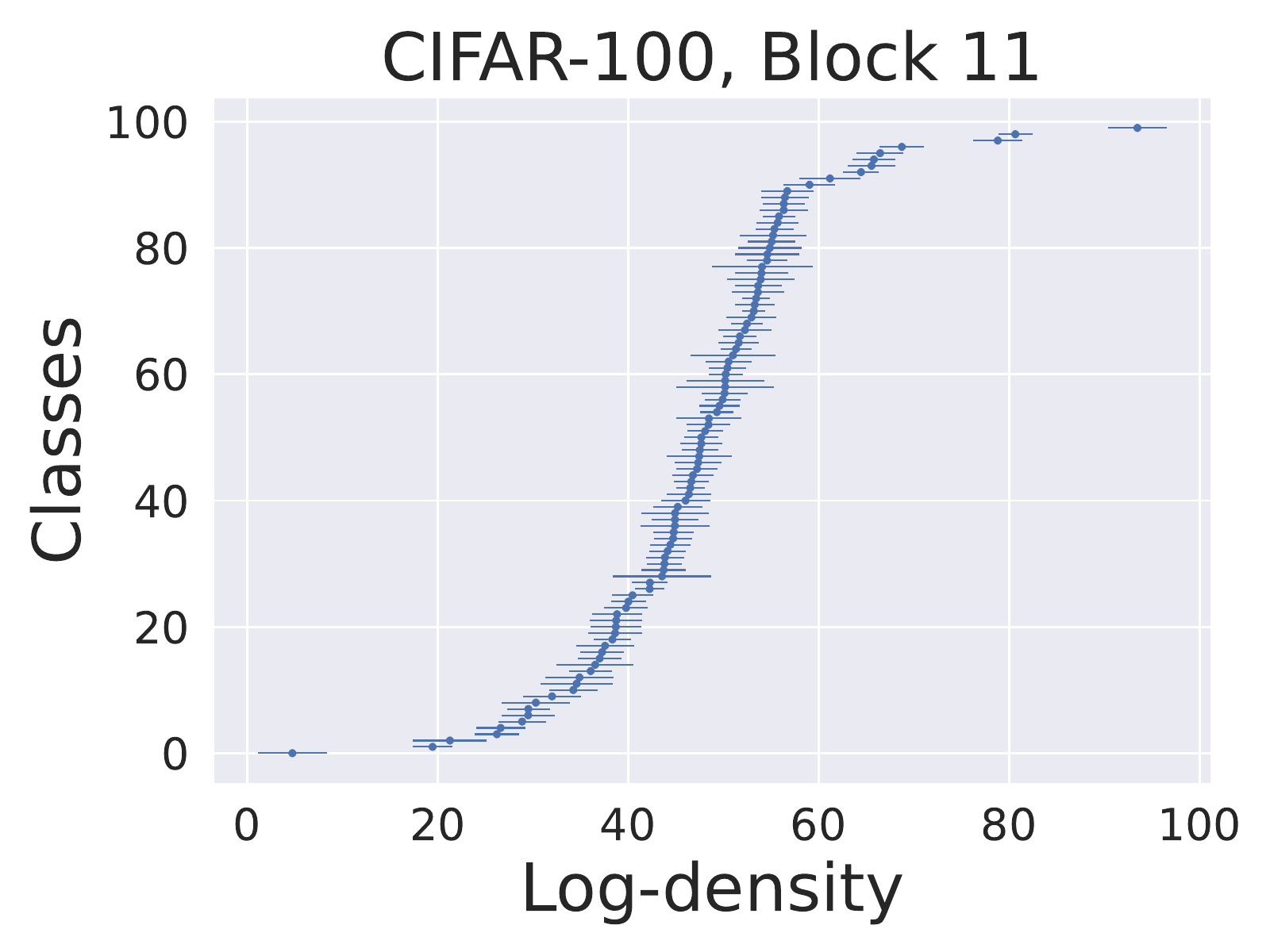}
  \includegraphics[width=0.244\linewidth]{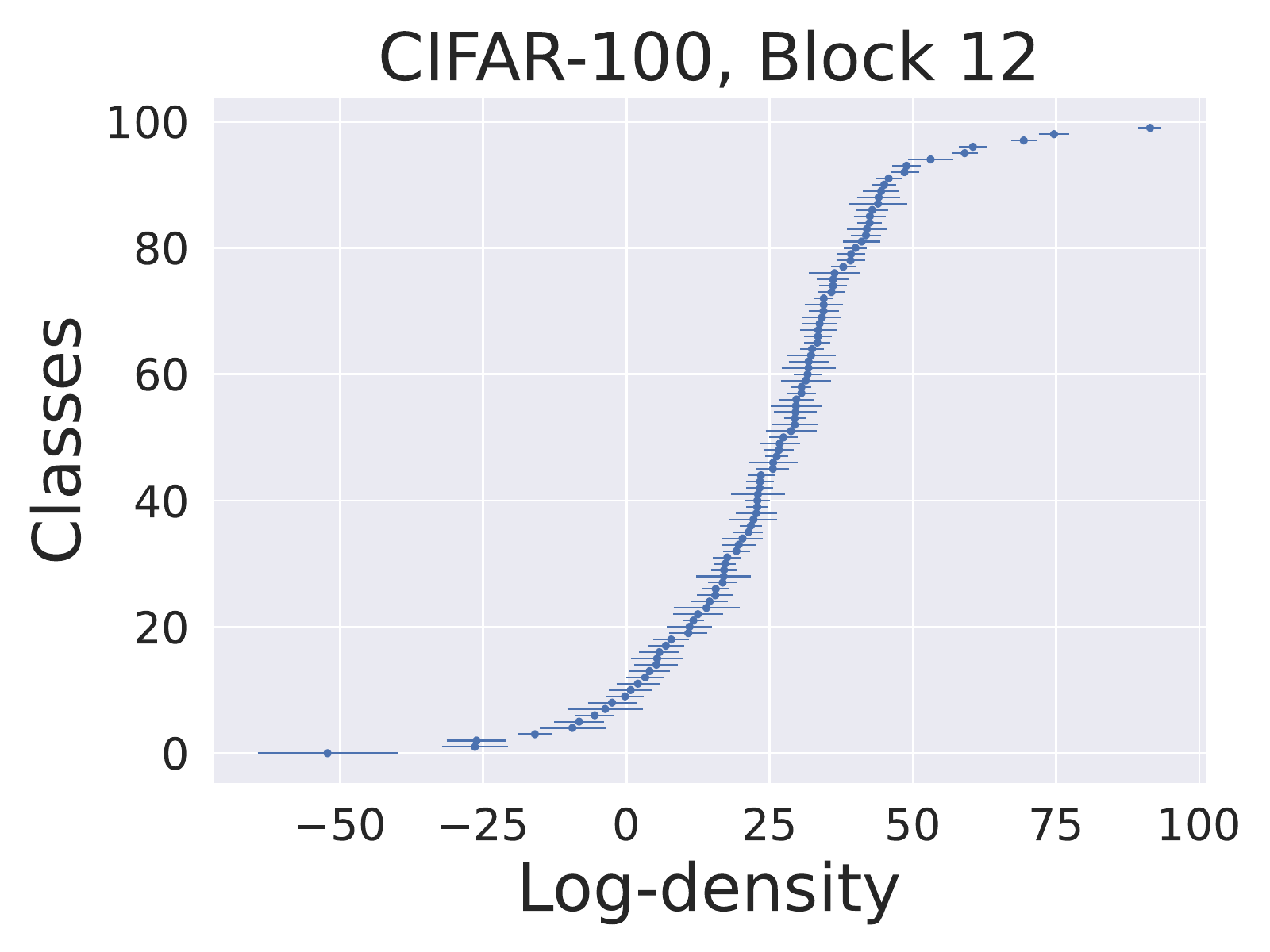}
  \\[1.0em]
  Vision Transformer, Mini-ImageNet \\[0.25em]
  \includegraphics[width=0.244\linewidth]{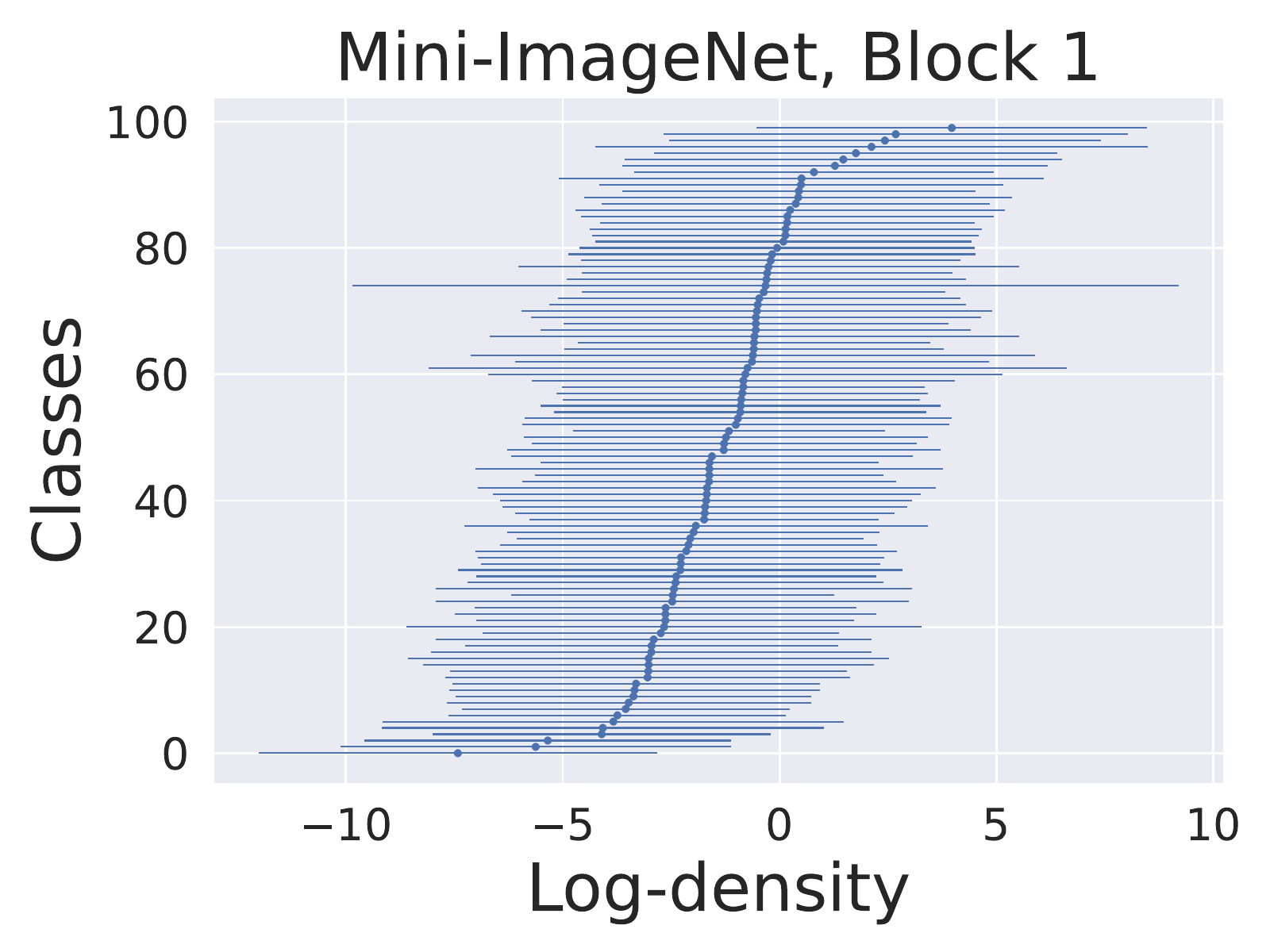}
  \includegraphics[width=0.244\linewidth]{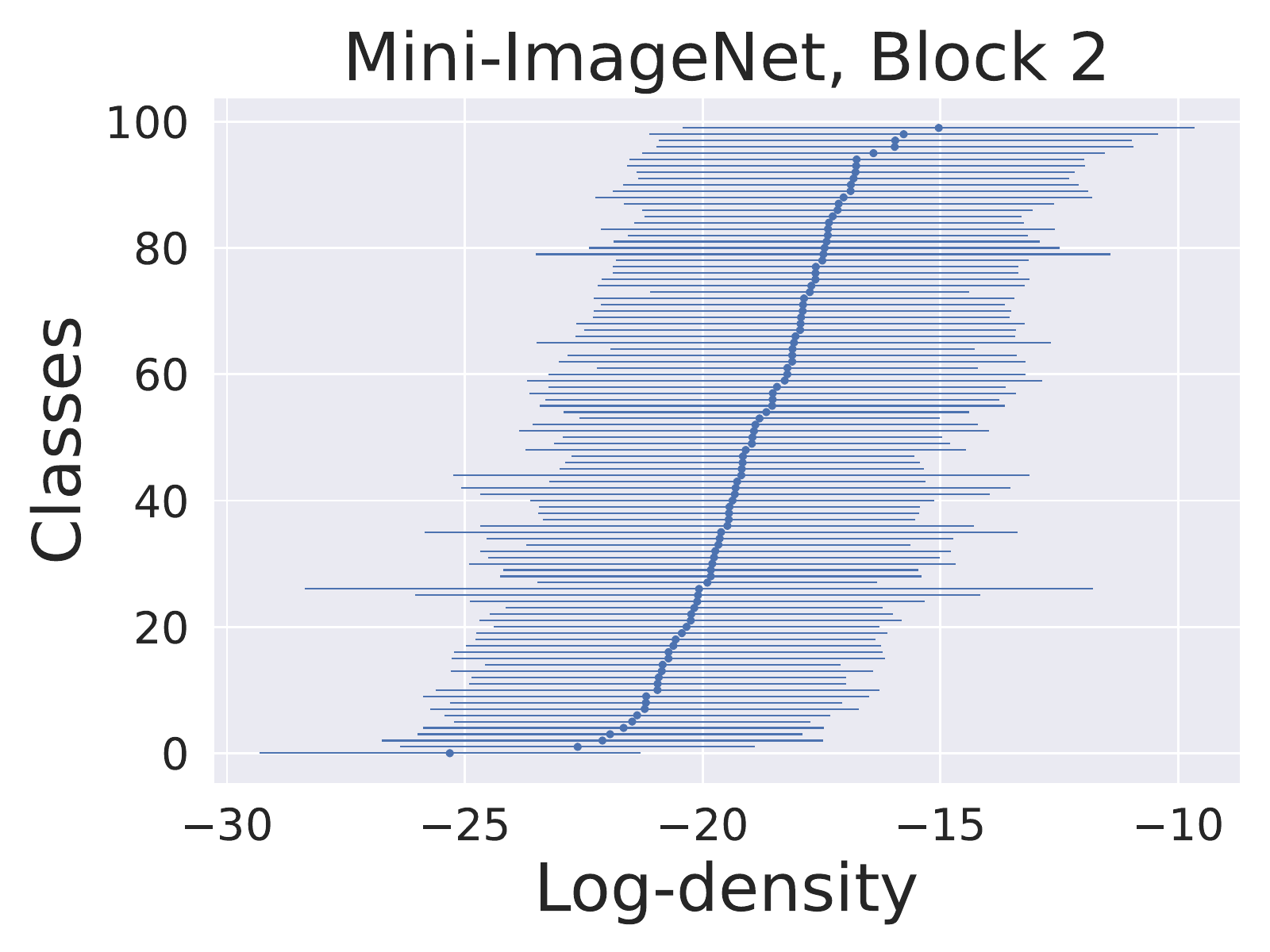}
  \includegraphics[width=0.244\linewidth]{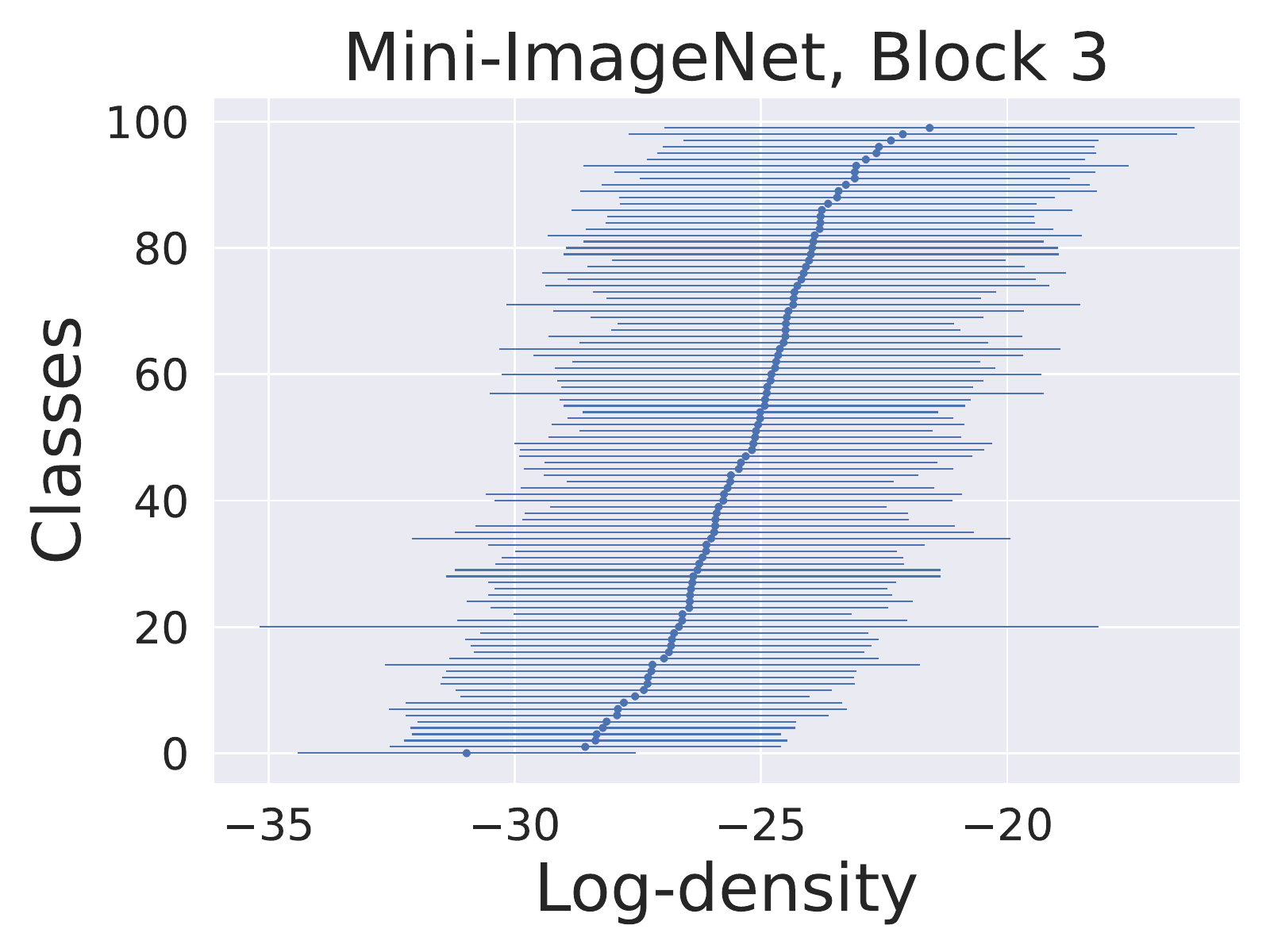}
  \includegraphics[width=0.244\linewidth]{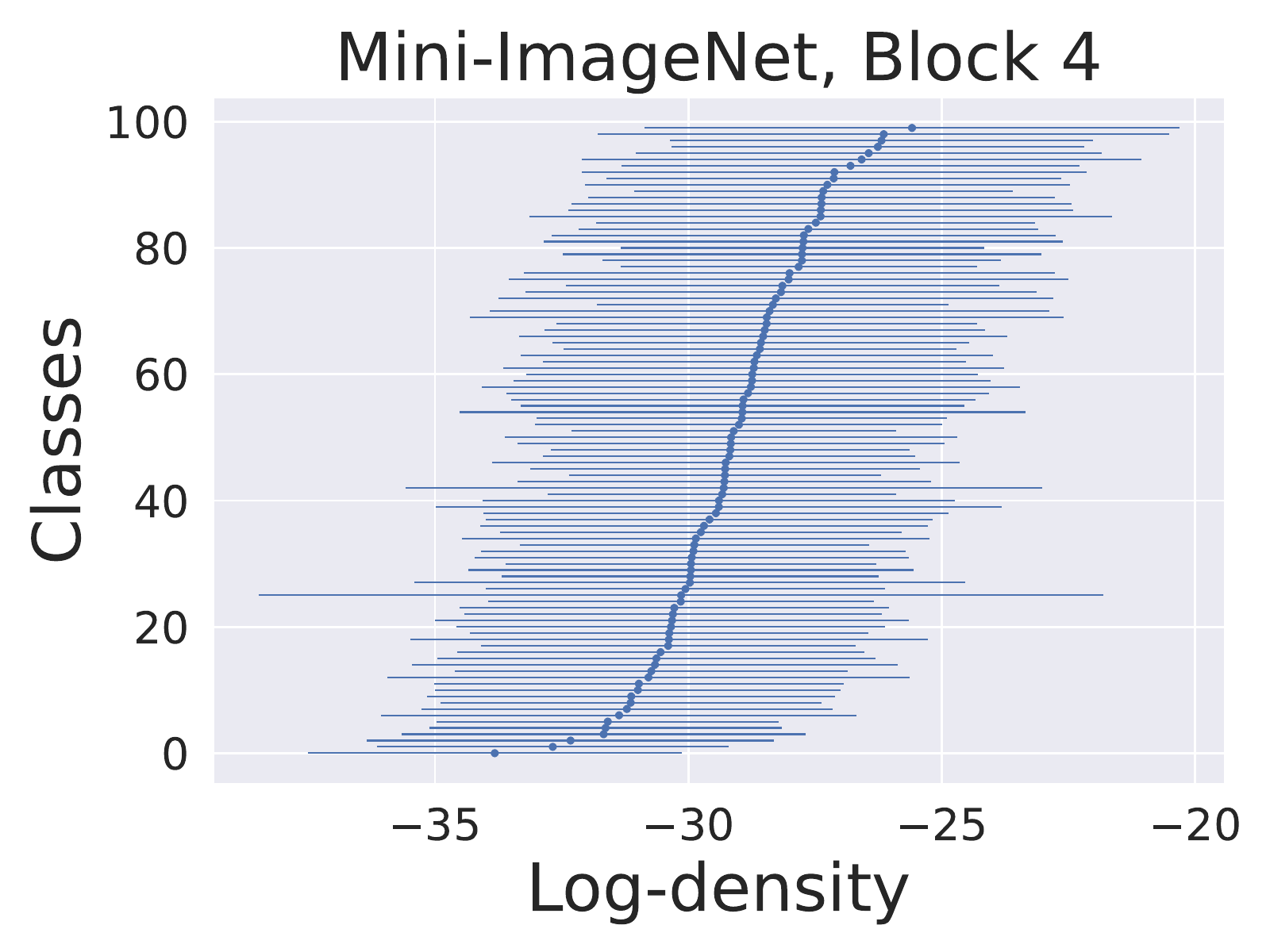}
  \\
  \includegraphics[width=0.244\linewidth]{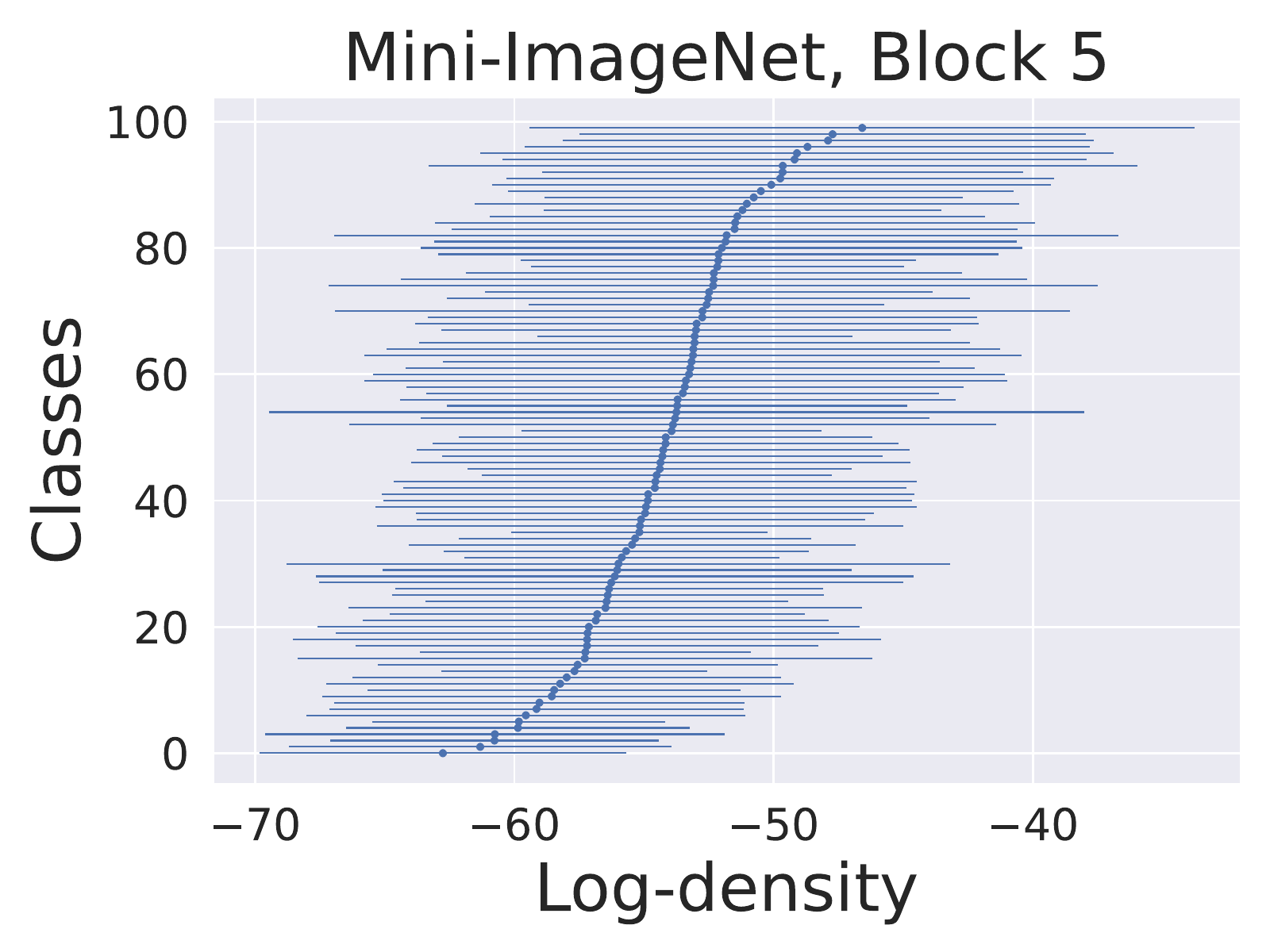}
  \includegraphics[width=0.244\linewidth]{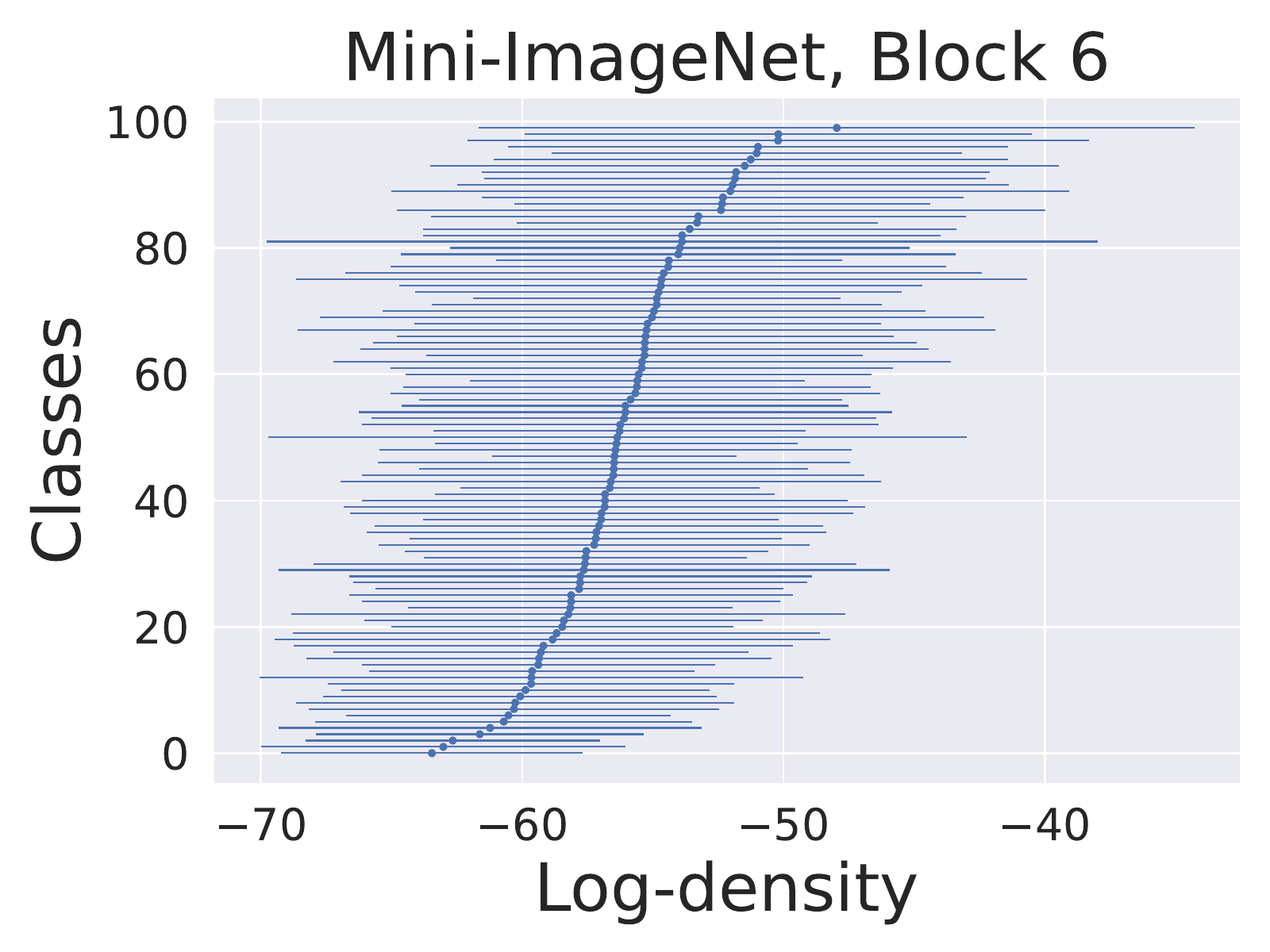}
  \includegraphics[width=0.244\linewidth]{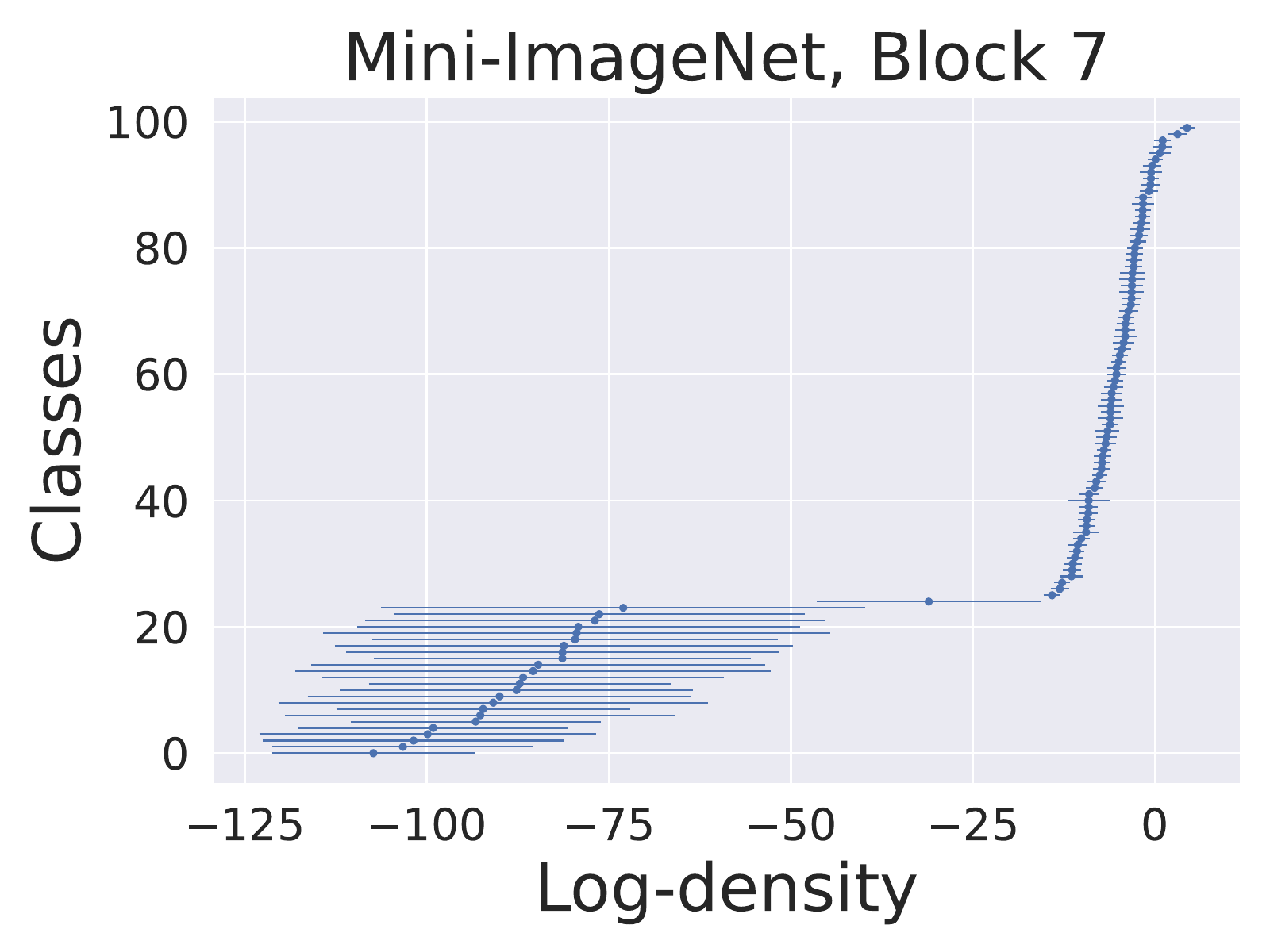}
  \includegraphics[width=0.244\linewidth]{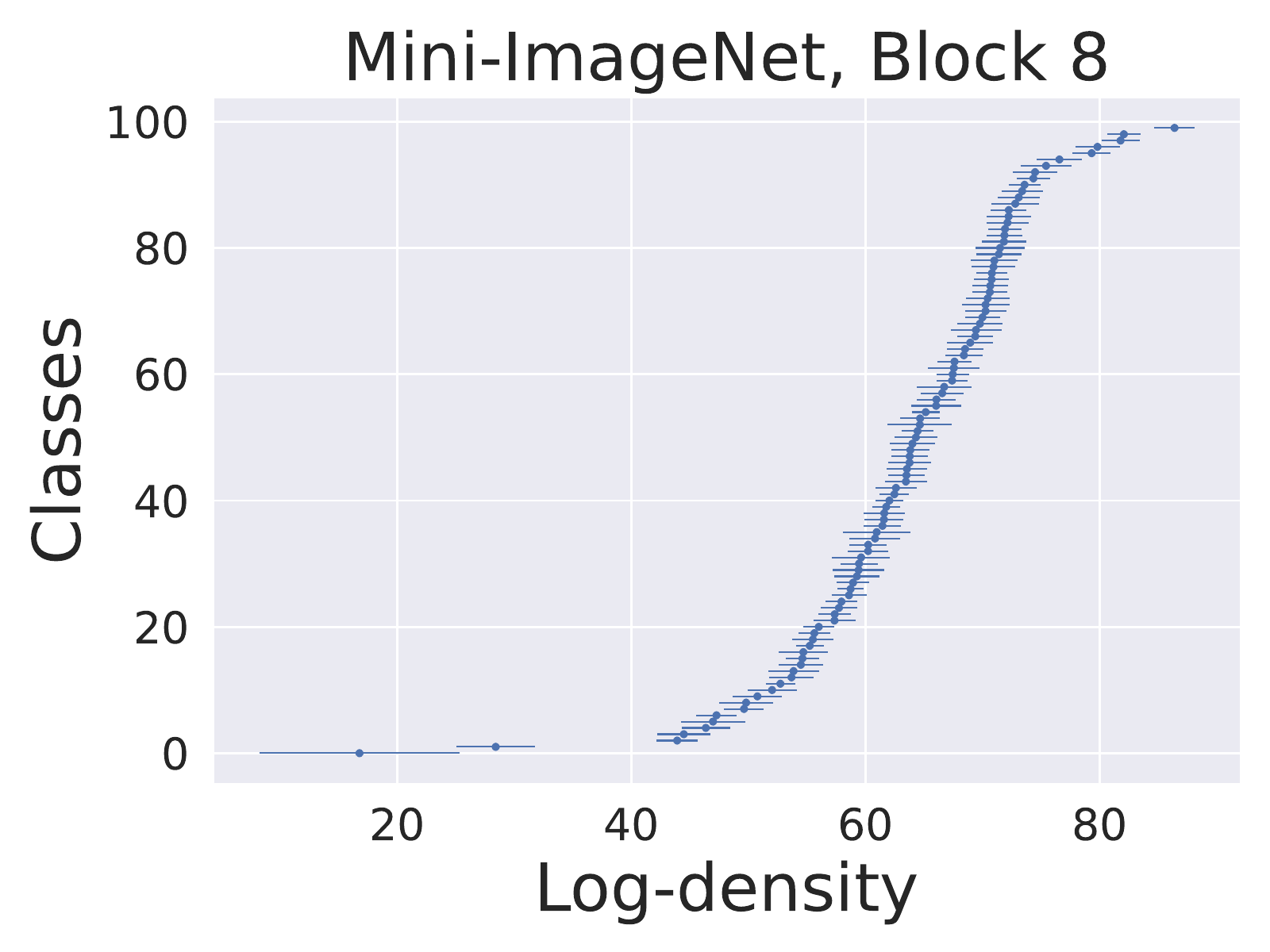}
  \\
  \includegraphics[width=0.244\linewidth]{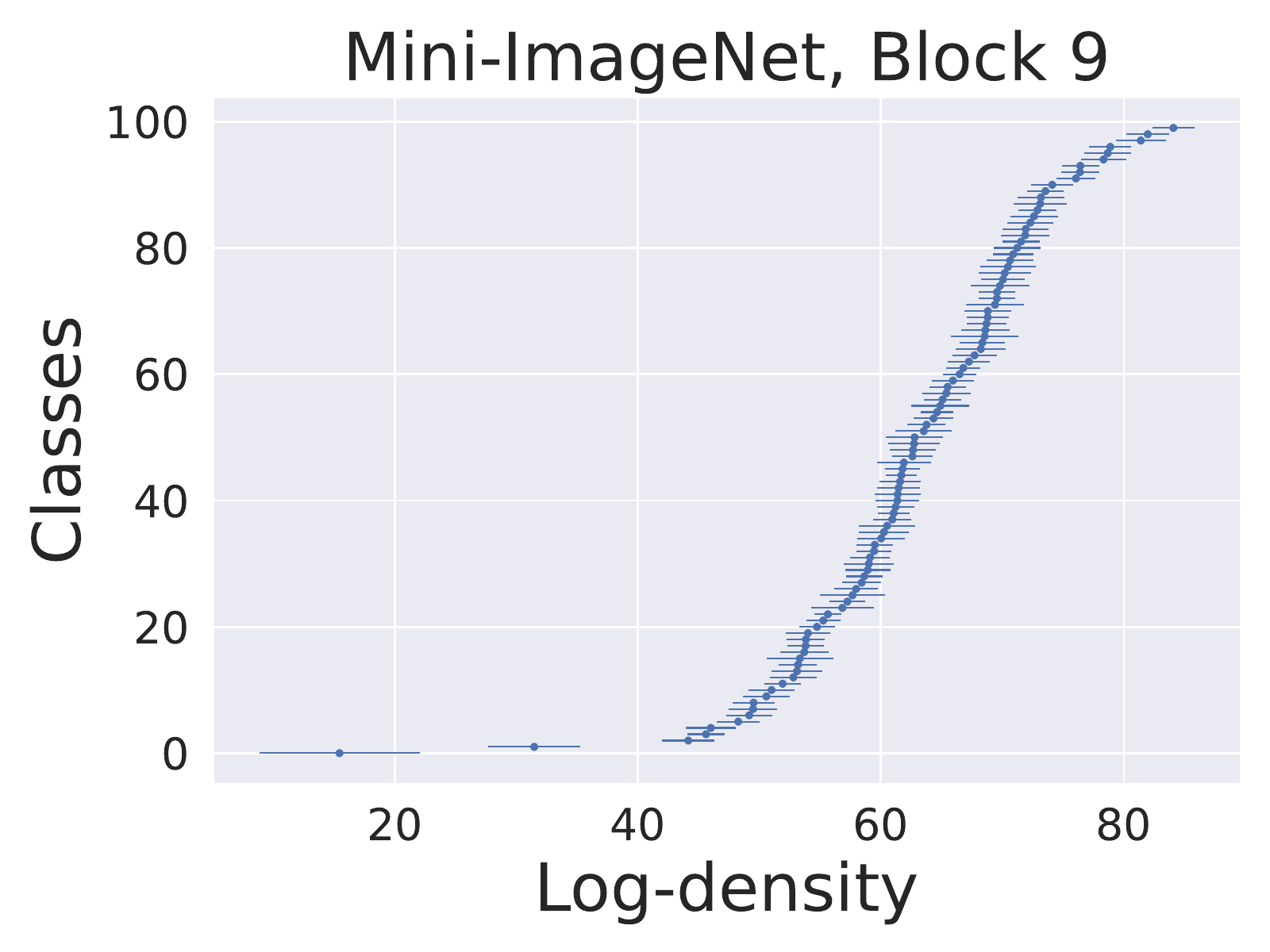}
  \includegraphics[width=0.244\linewidth]{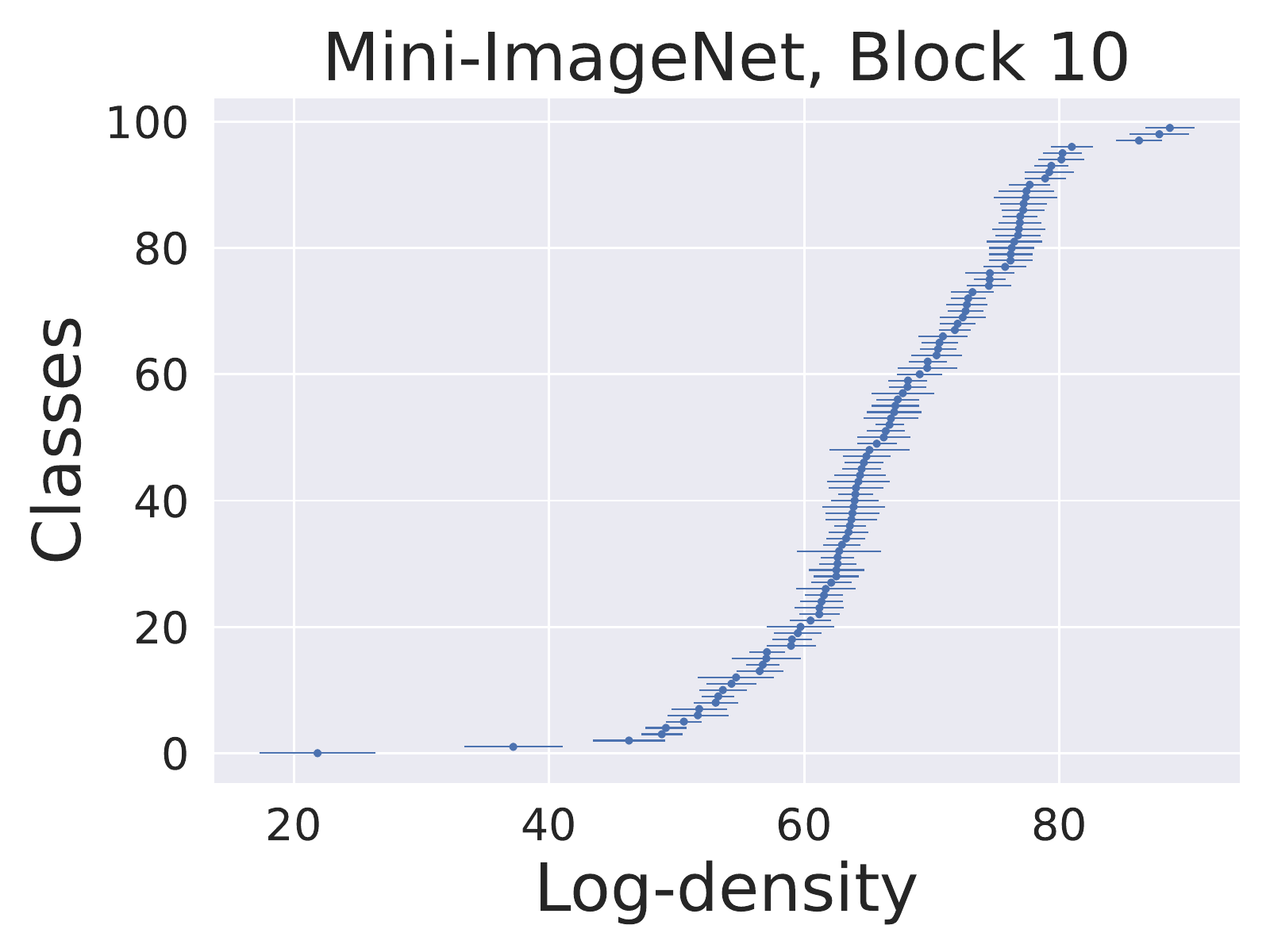}
  \includegraphics[width=0.244\linewidth]{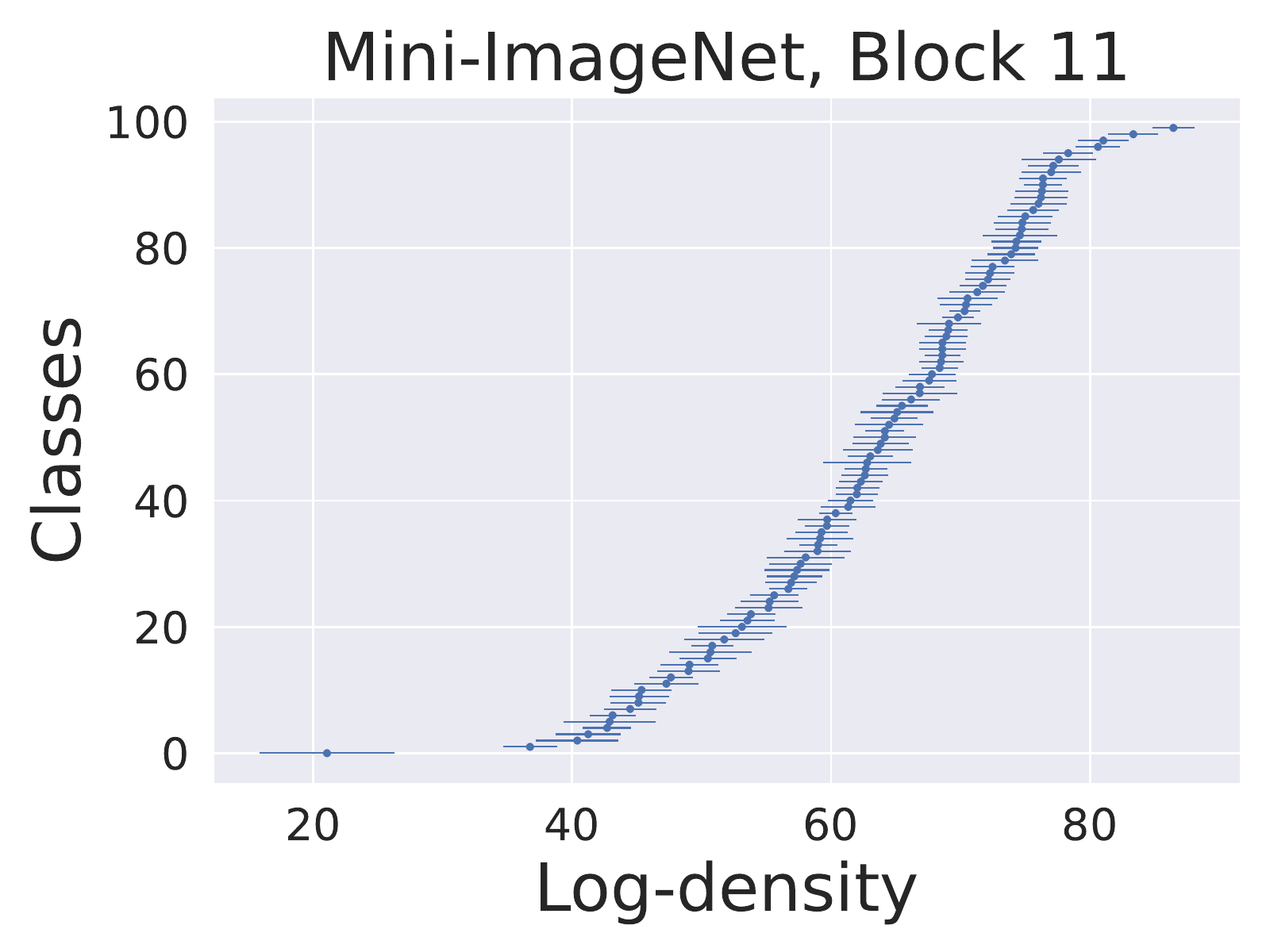}
  \includegraphics[width=0.244\linewidth]{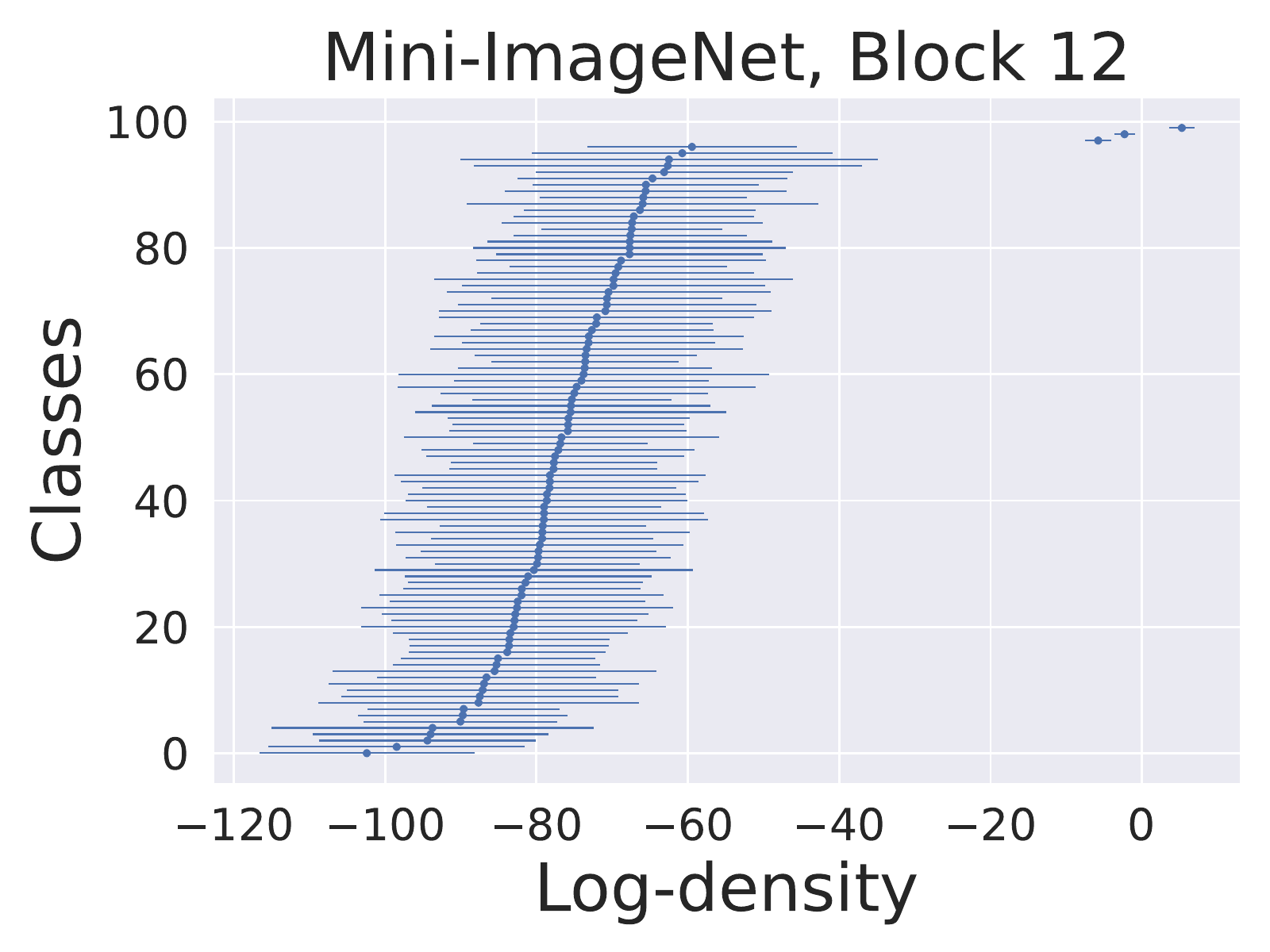}
  \caption{Mean and standard deviation of class-conditional log-densities estimated for representations from a 12-block
           Vision Transformer model.}
  \label{fig:log_cc_densities_vit}
\end{figure*}

\begin{figure*}[htb]
  \centering
  MLP-Mixer, CIFAR100 \\[0.25em]
  \includegraphics[width=0.244\linewidth]{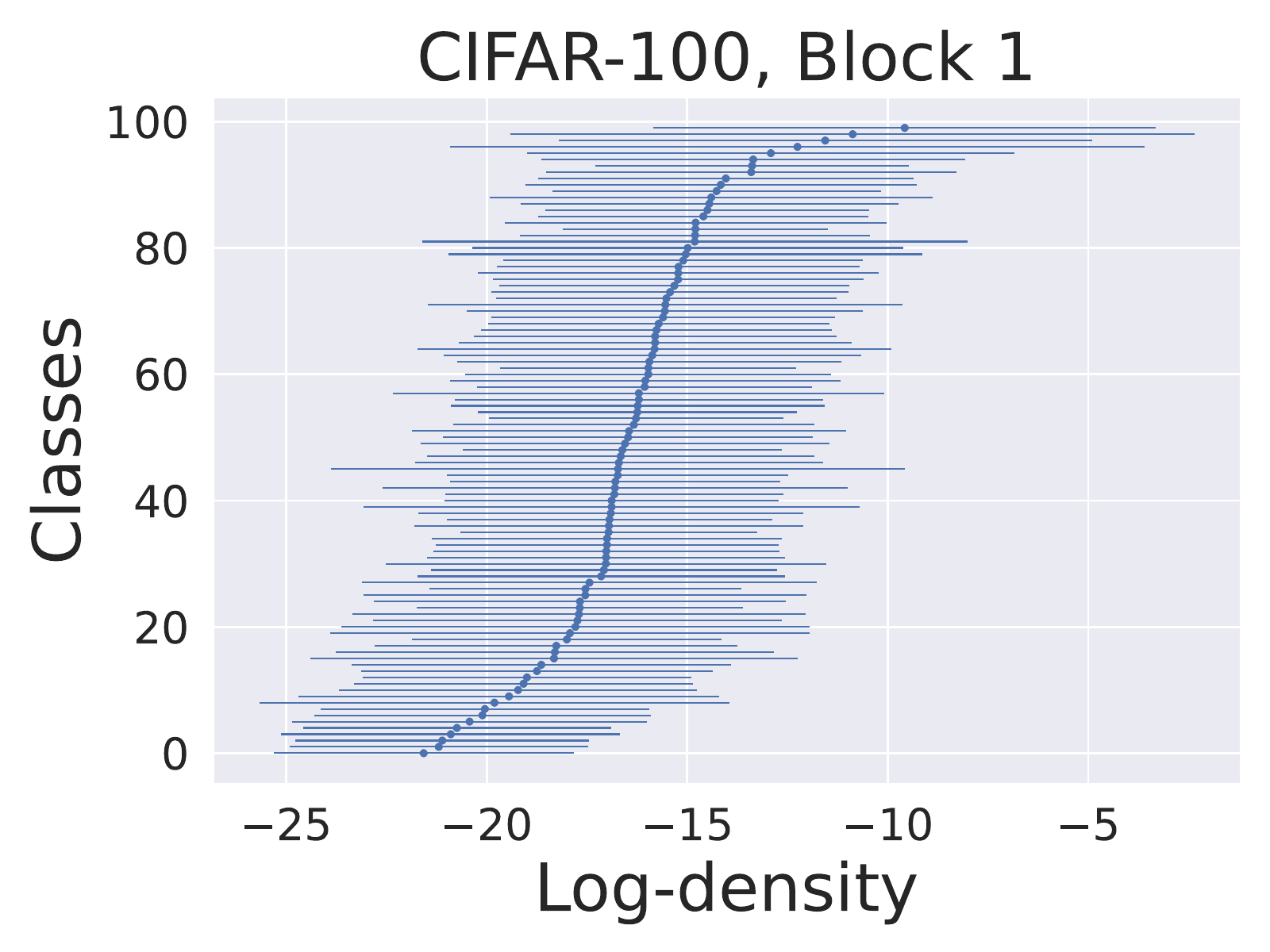}
  \includegraphics[width=0.244\linewidth]{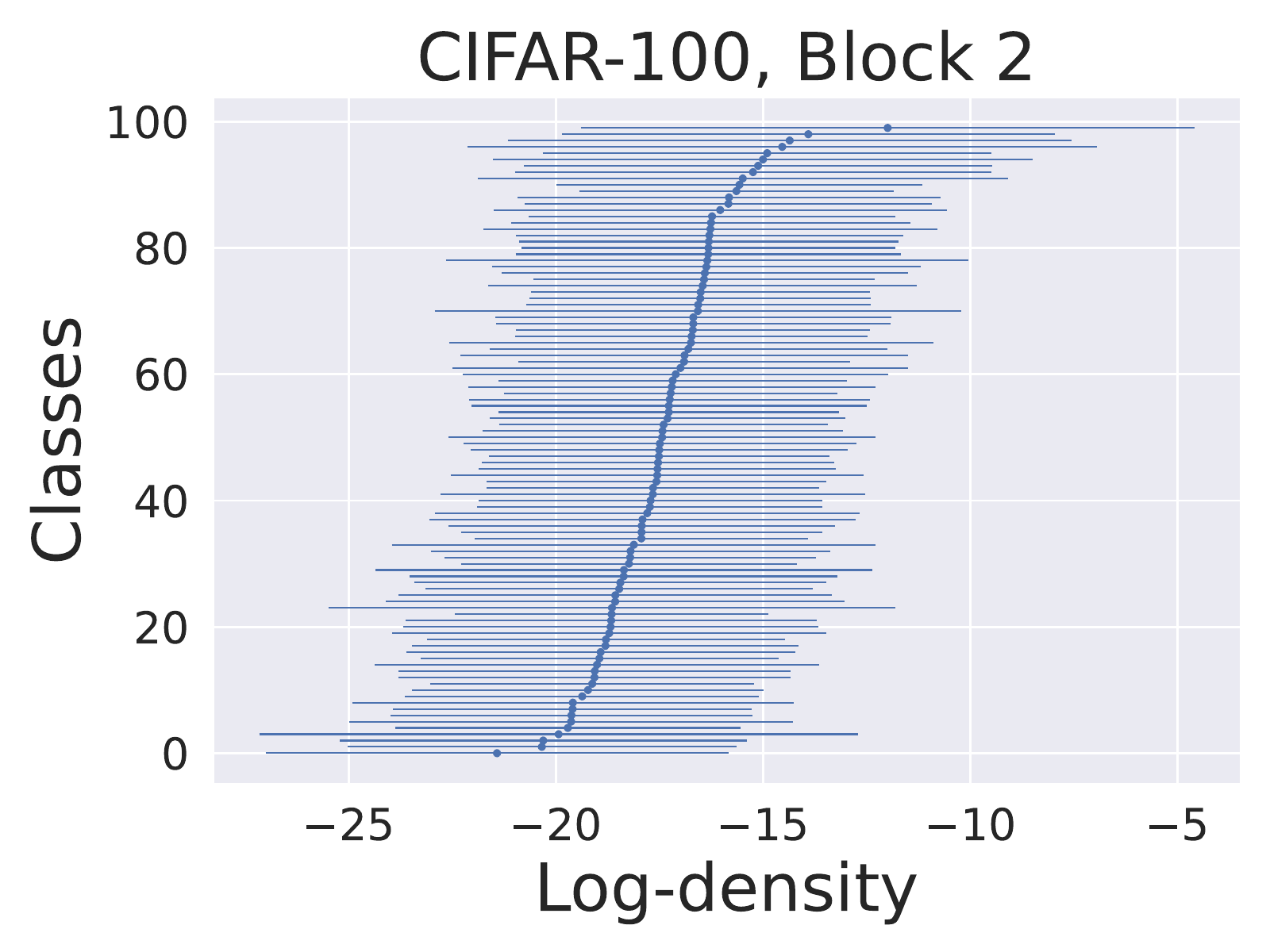}
  \includegraphics[width=0.244\linewidth]{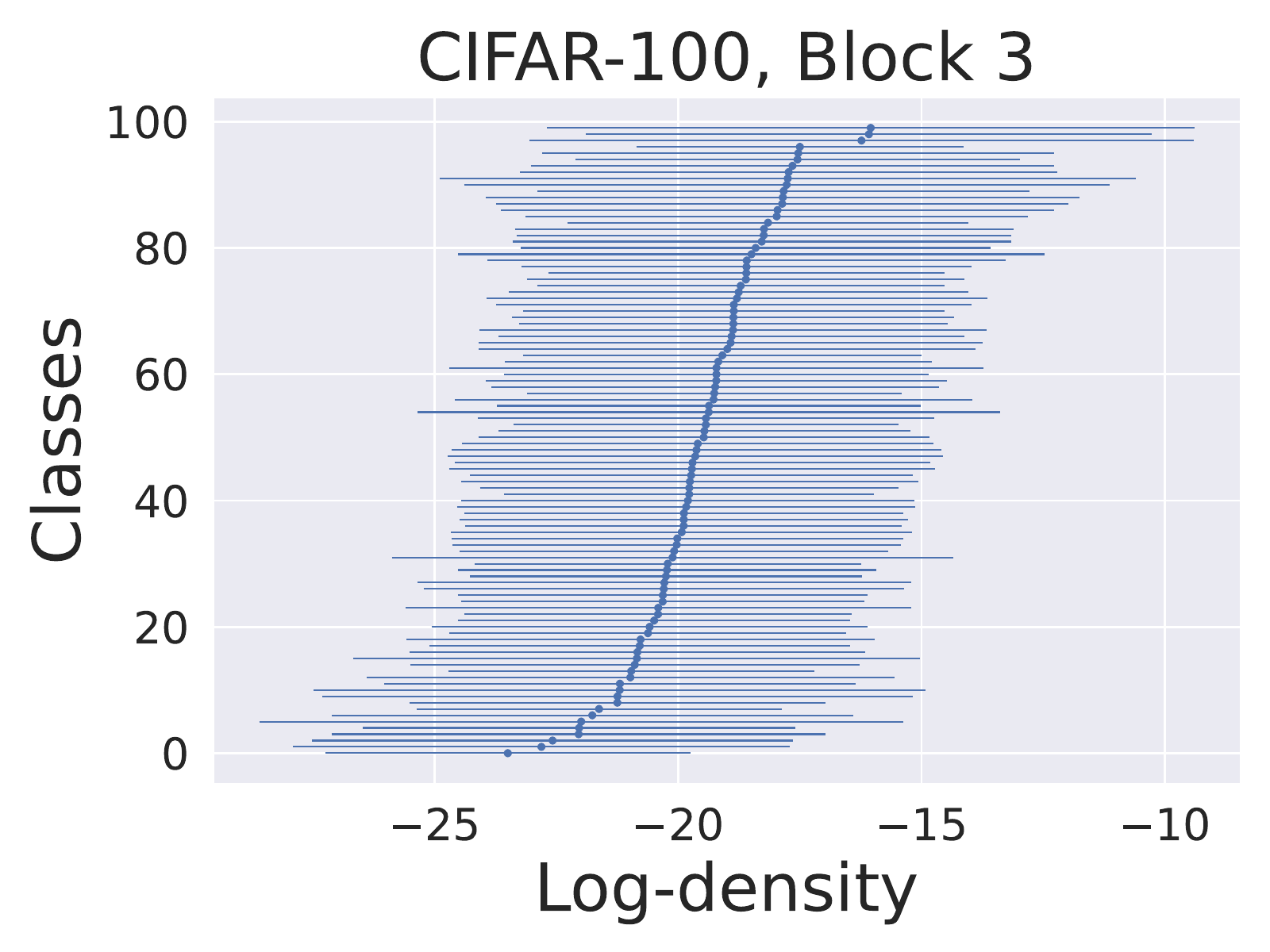}
  \includegraphics[width=0.244\linewidth]{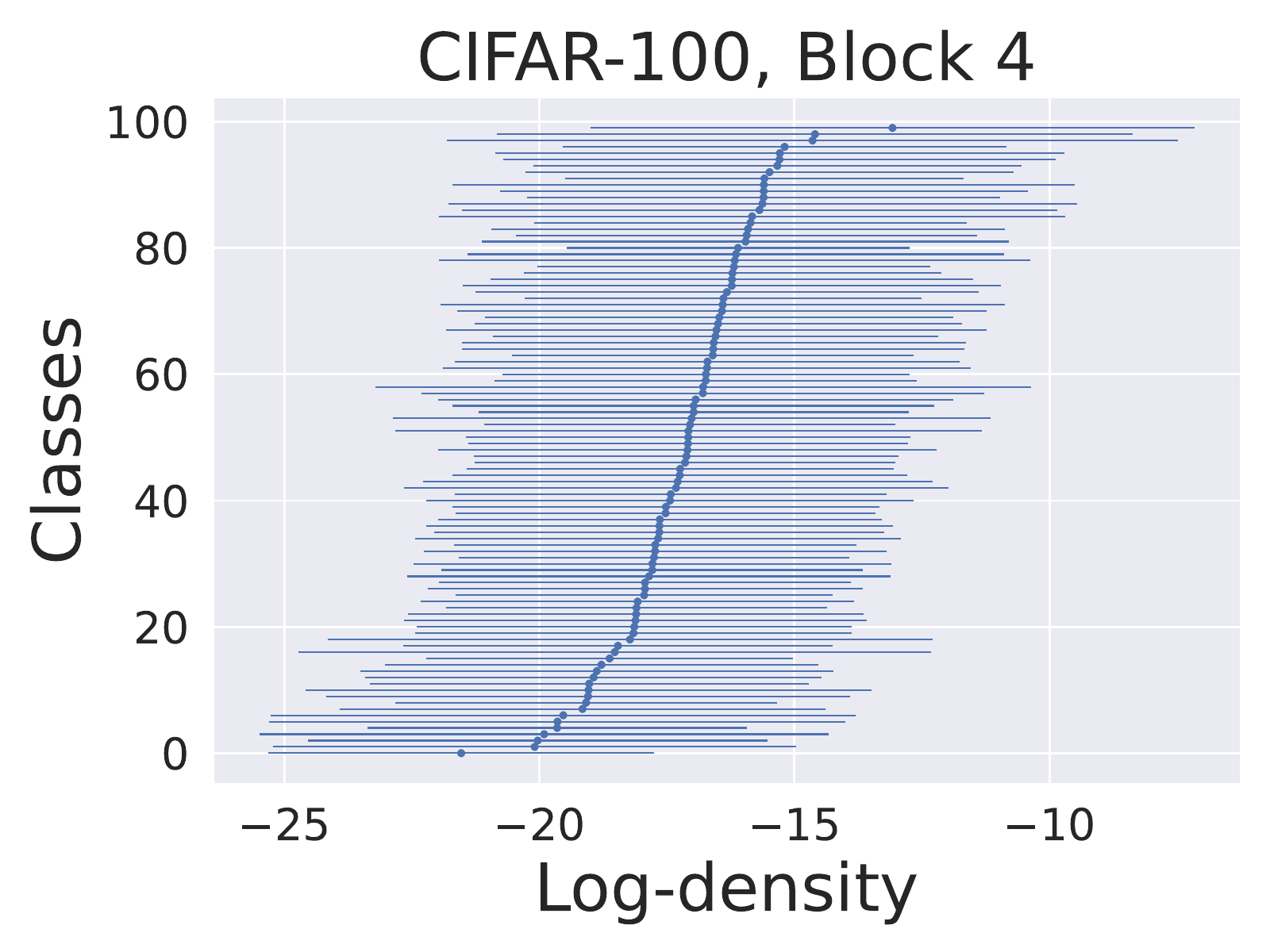}
  \\
  \includegraphics[width=0.244\linewidth]{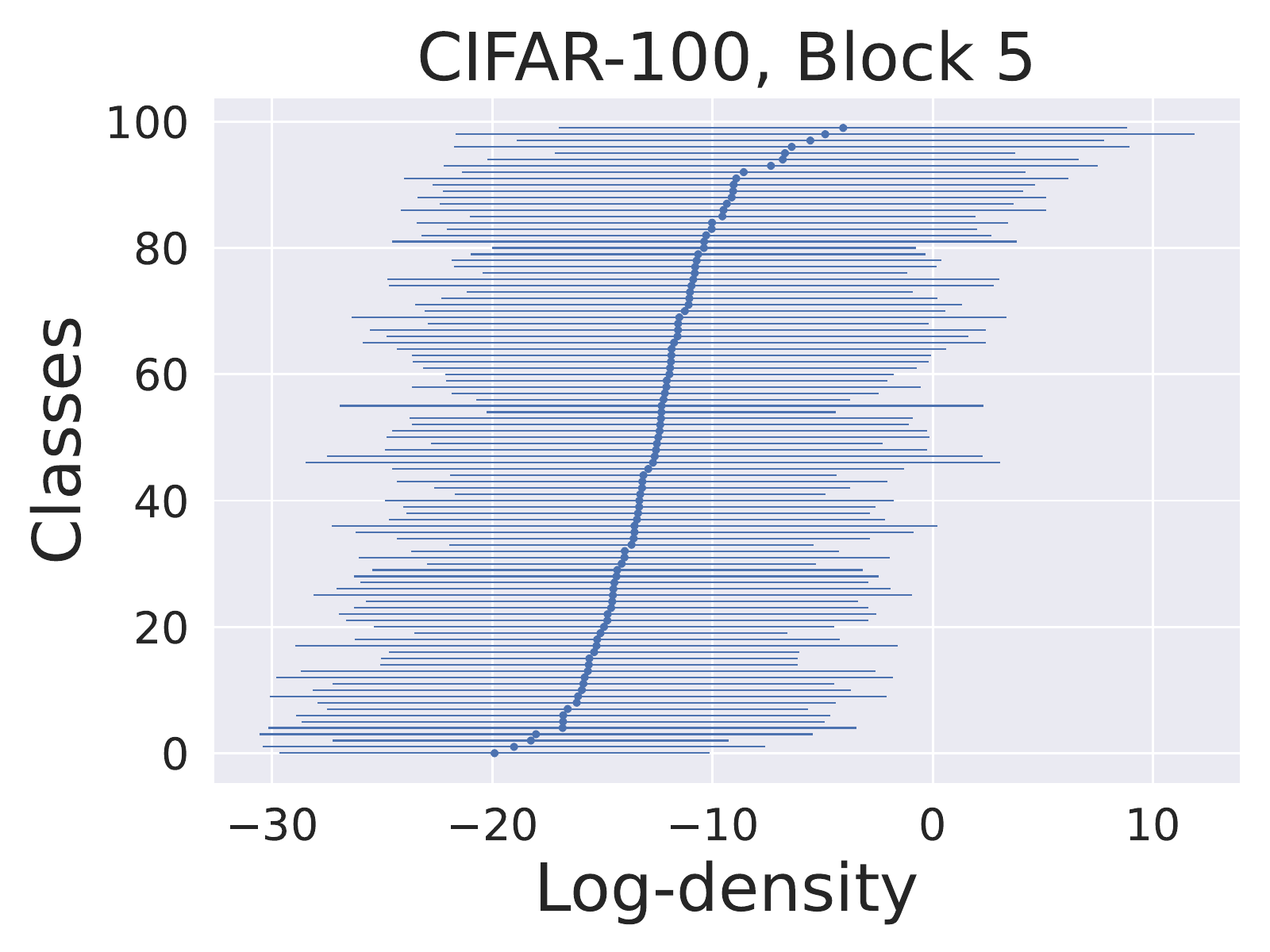}
  \includegraphics[width=0.244\linewidth]{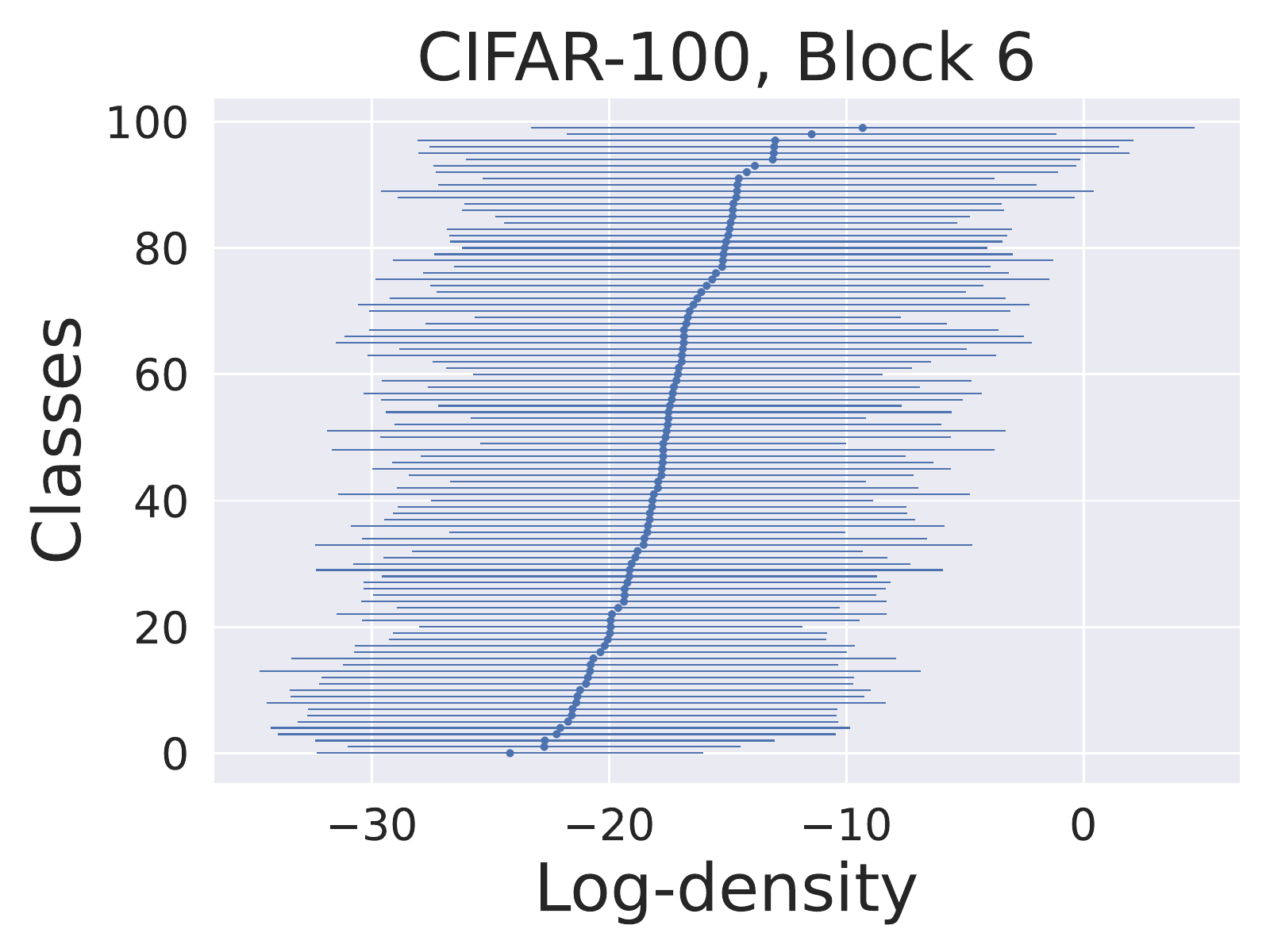}
  \includegraphics[width=0.244\linewidth]{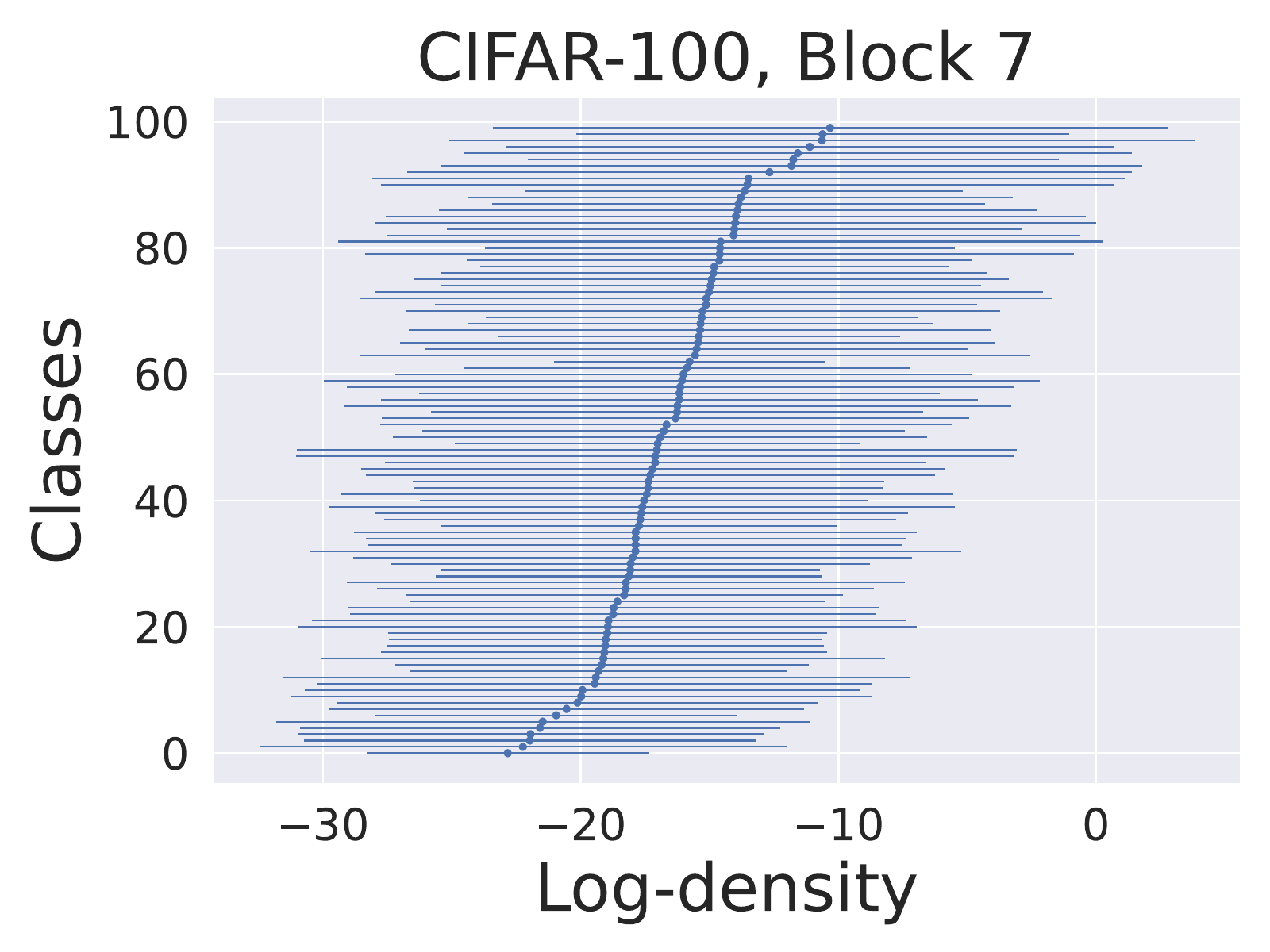}
  \includegraphics[width=0.244\linewidth]{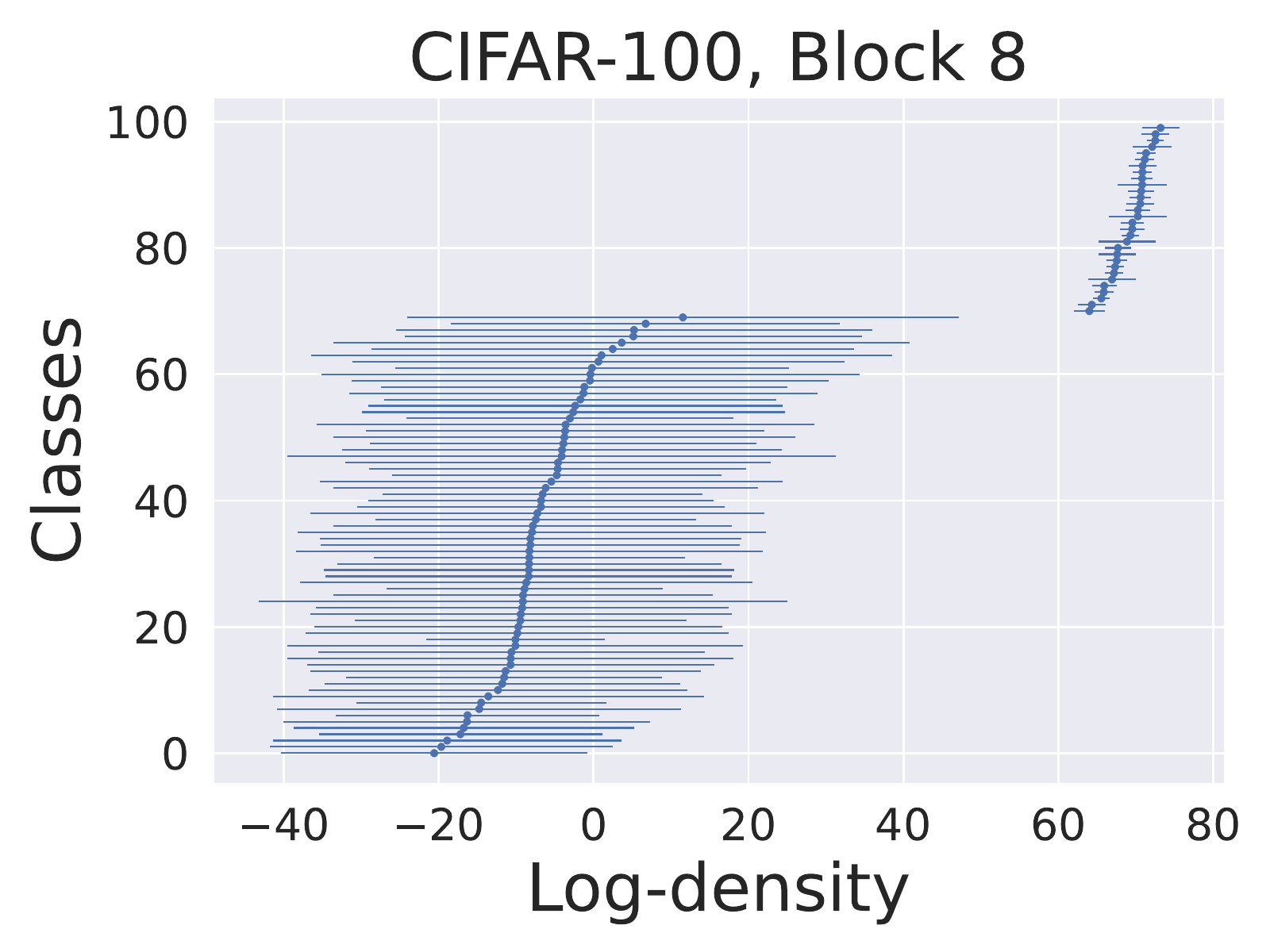}
  \\
  \includegraphics[width=0.244\linewidth]{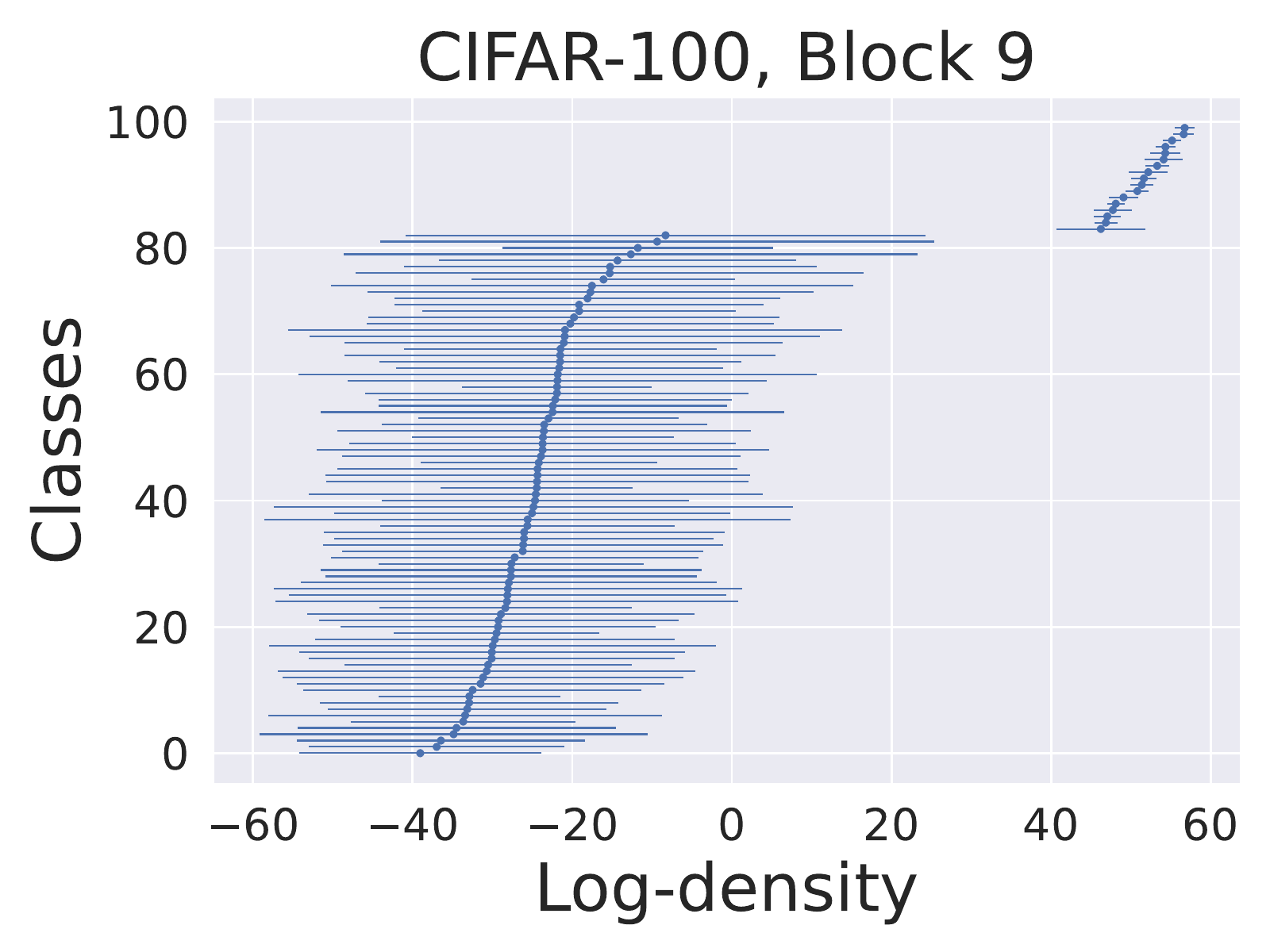}
  \includegraphics[width=0.244\linewidth]{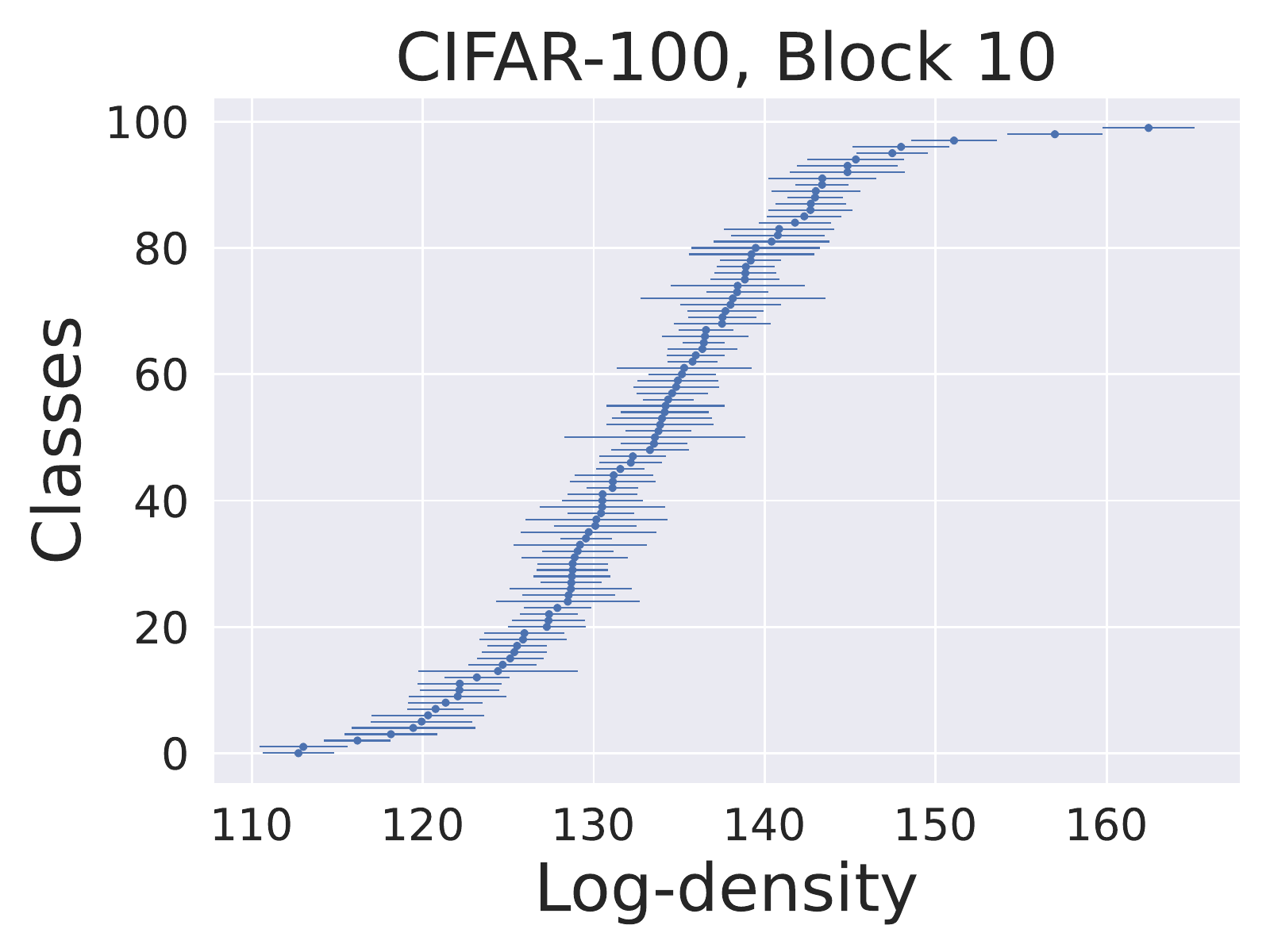}
  \includegraphics[width=0.244\linewidth]{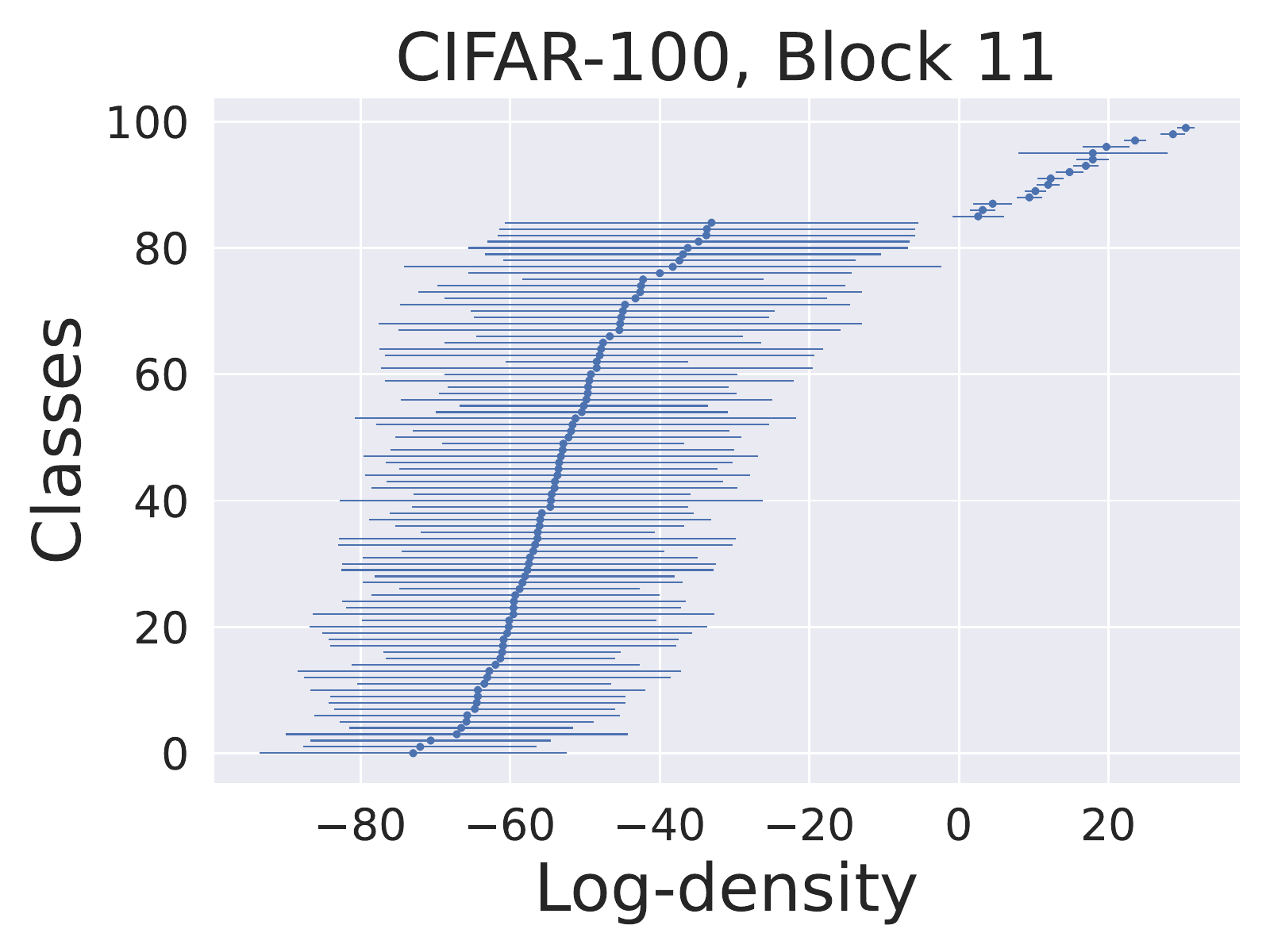}
  \includegraphics[width=0.244\linewidth]{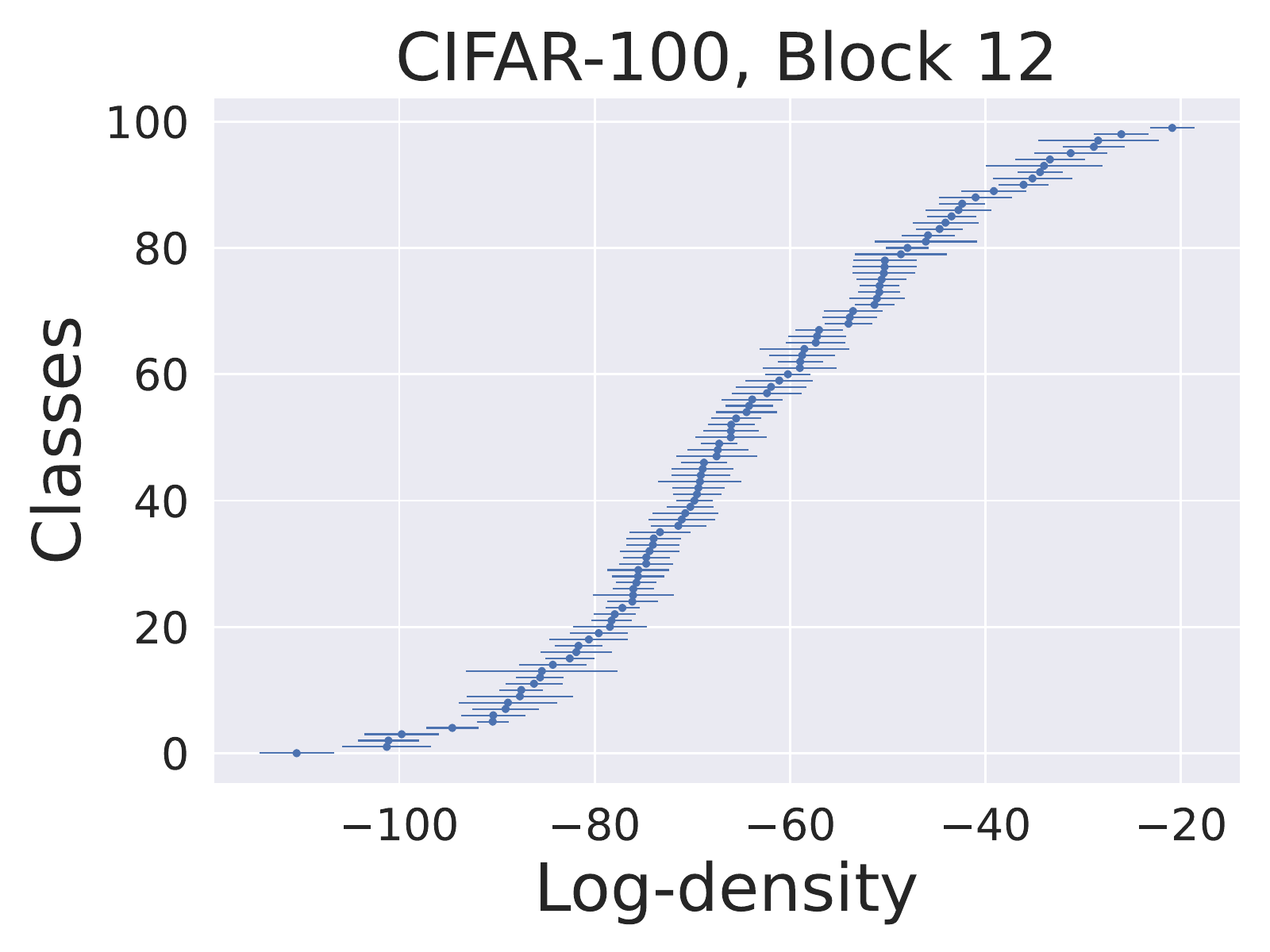}
  \\[1.0em]
  MLP-Mixer, Mini-ImageNet \\[0.25em]
  \includegraphics[width=0.244\linewidth]{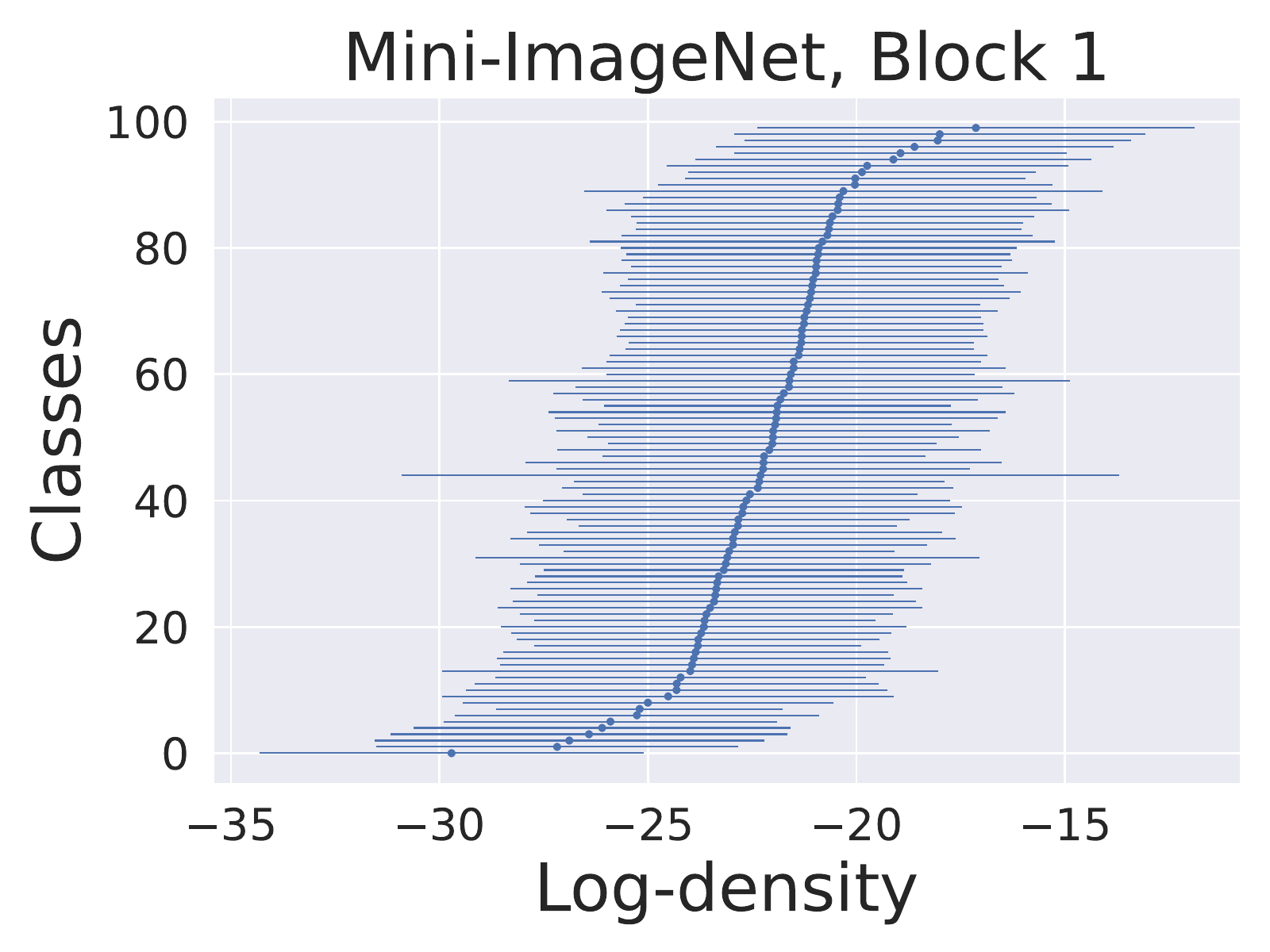}
  \includegraphics[width=0.244\linewidth]{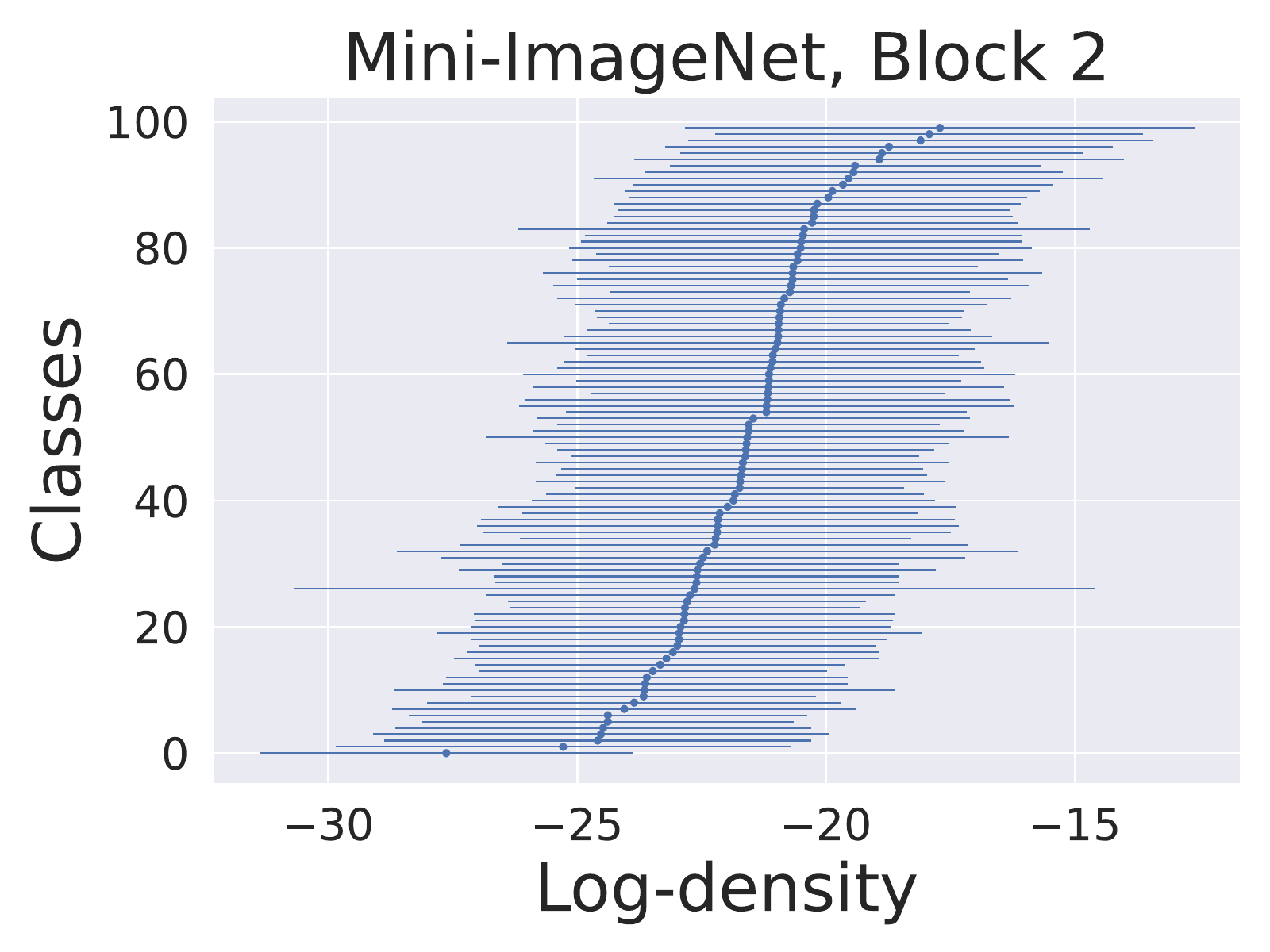}
  \includegraphics[width=0.244\linewidth]{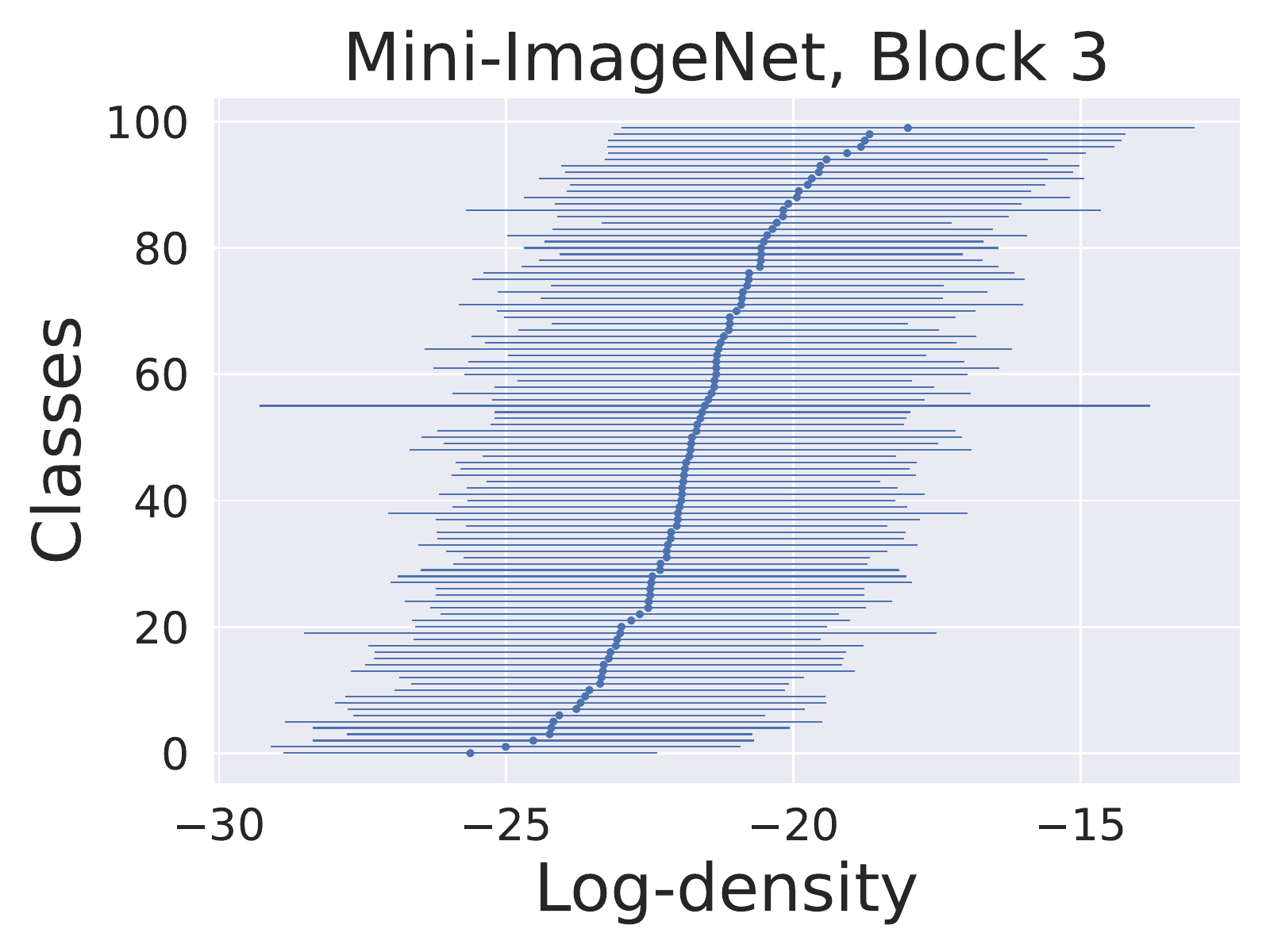}
  \includegraphics[width=0.244\linewidth]{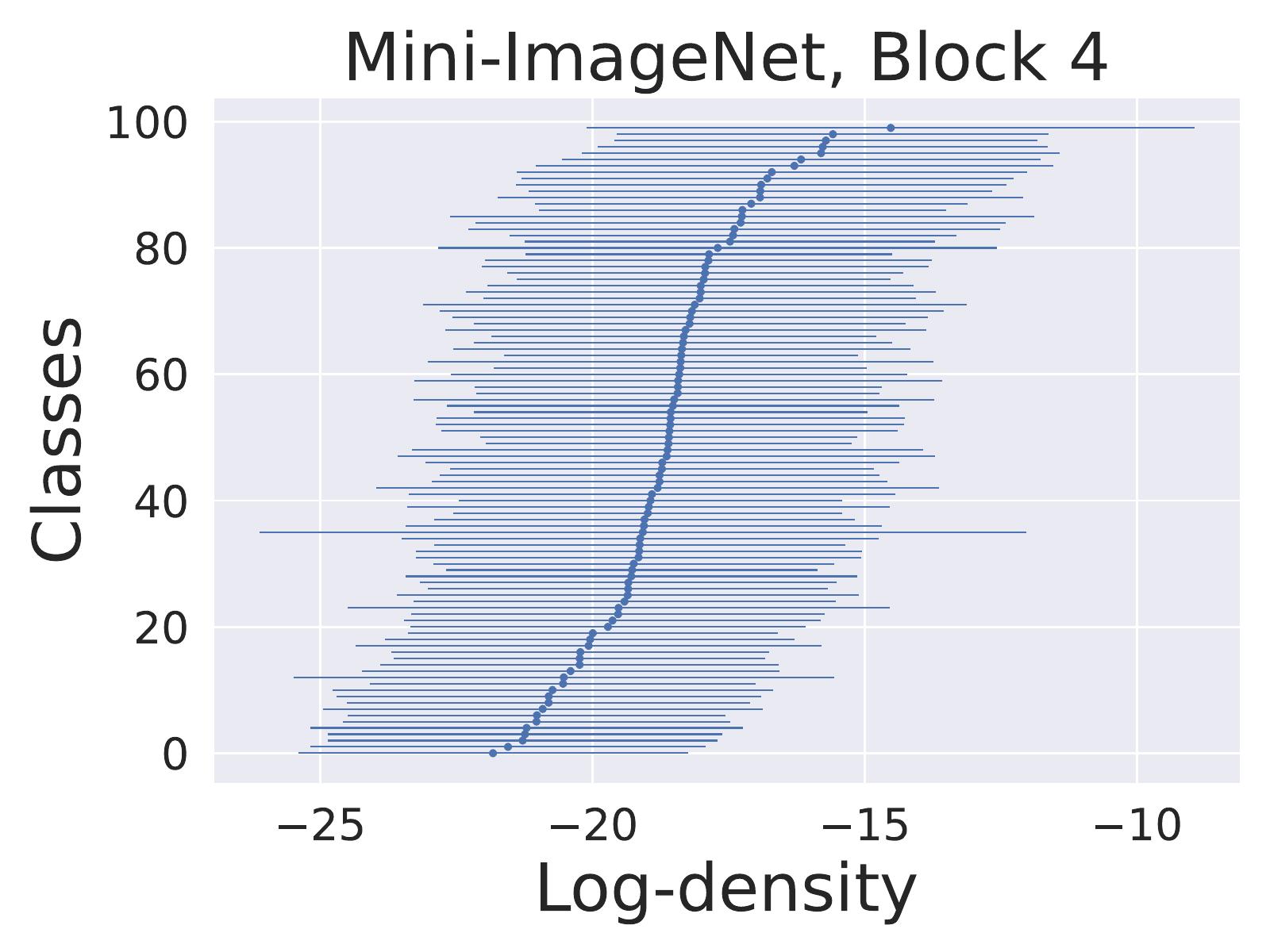}
  \\
  \includegraphics[width=0.244\linewidth]{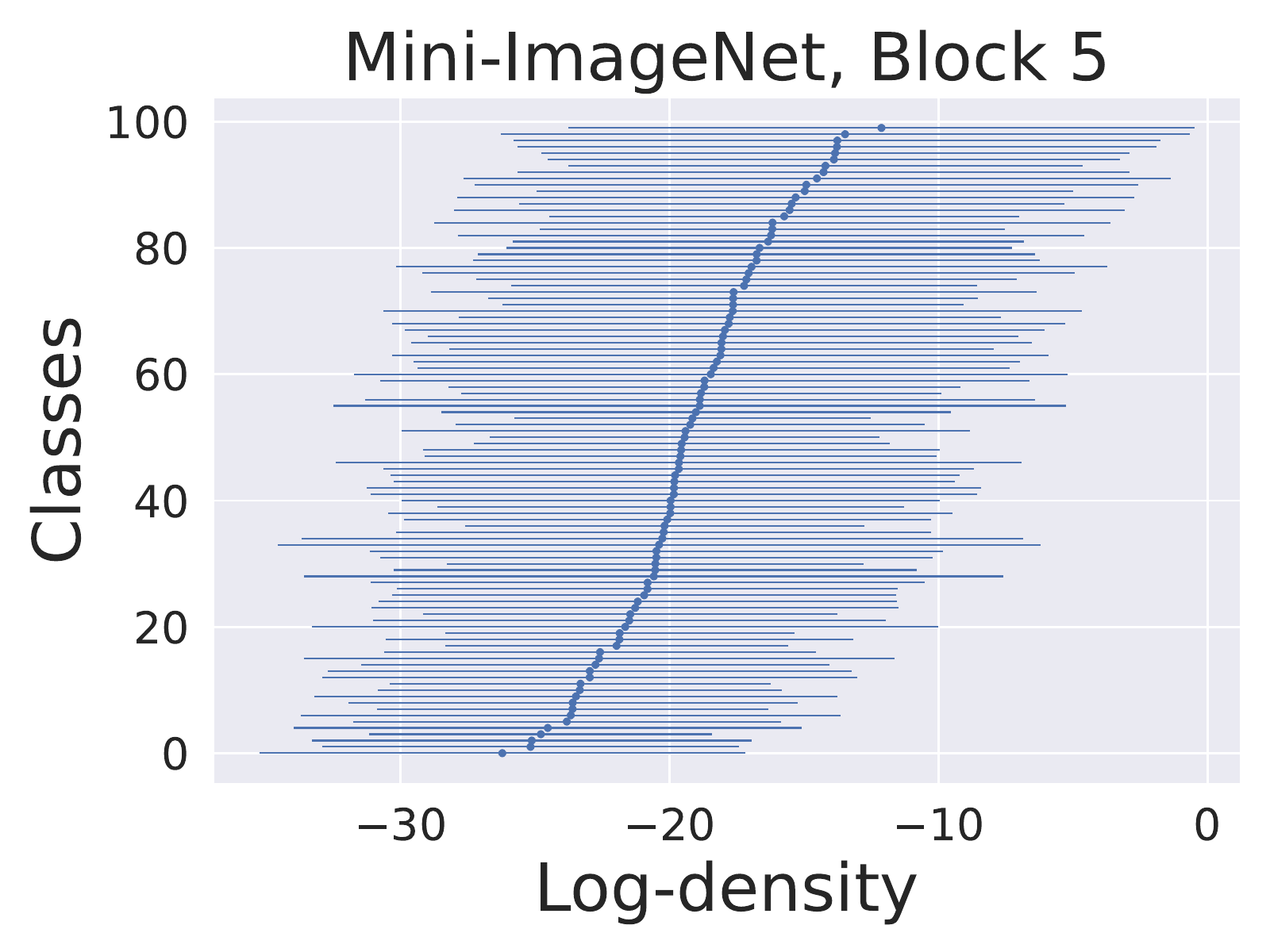}
  \includegraphics[width=0.244\linewidth]{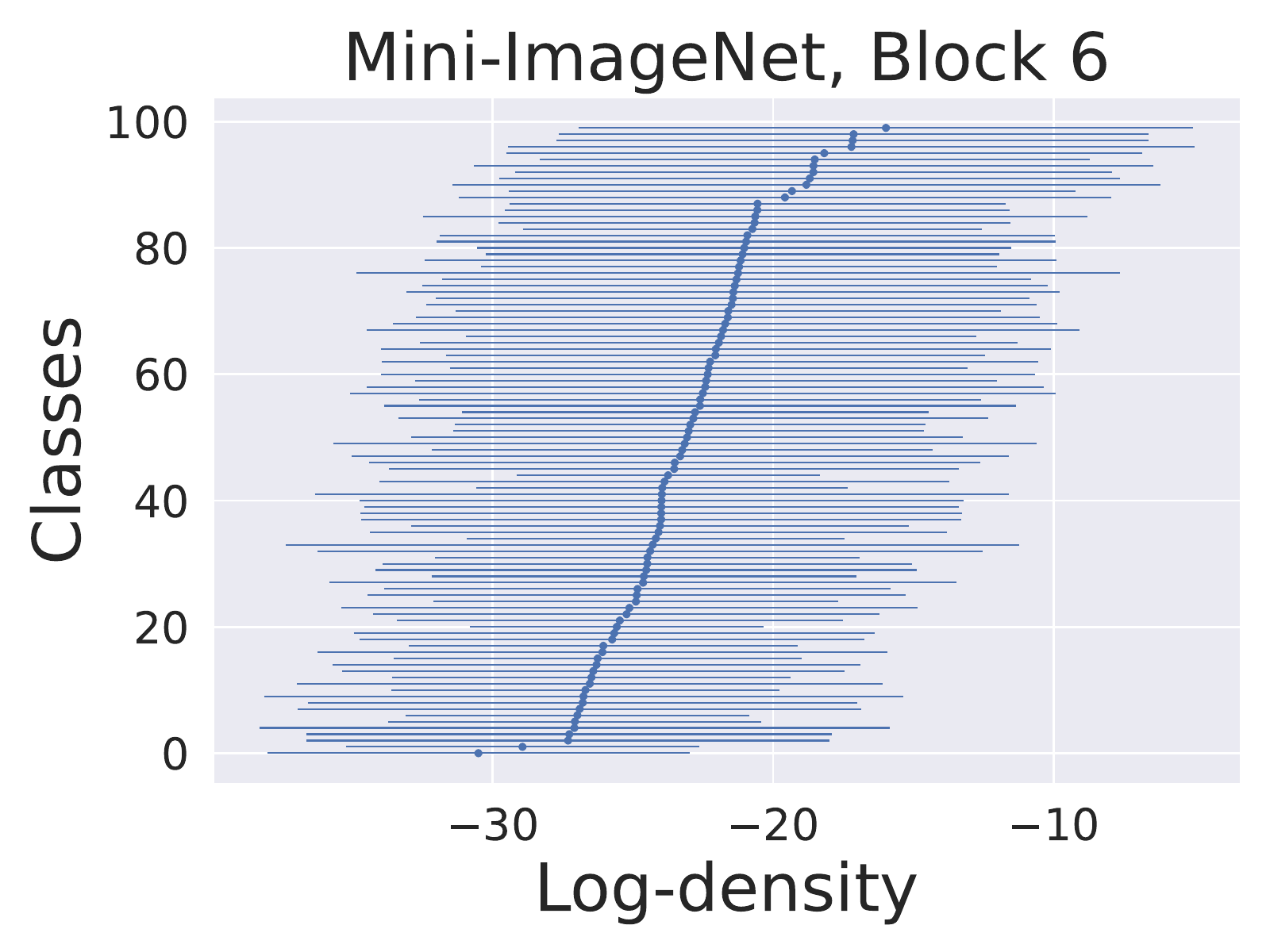}
  \includegraphics[width=0.244\linewidth]{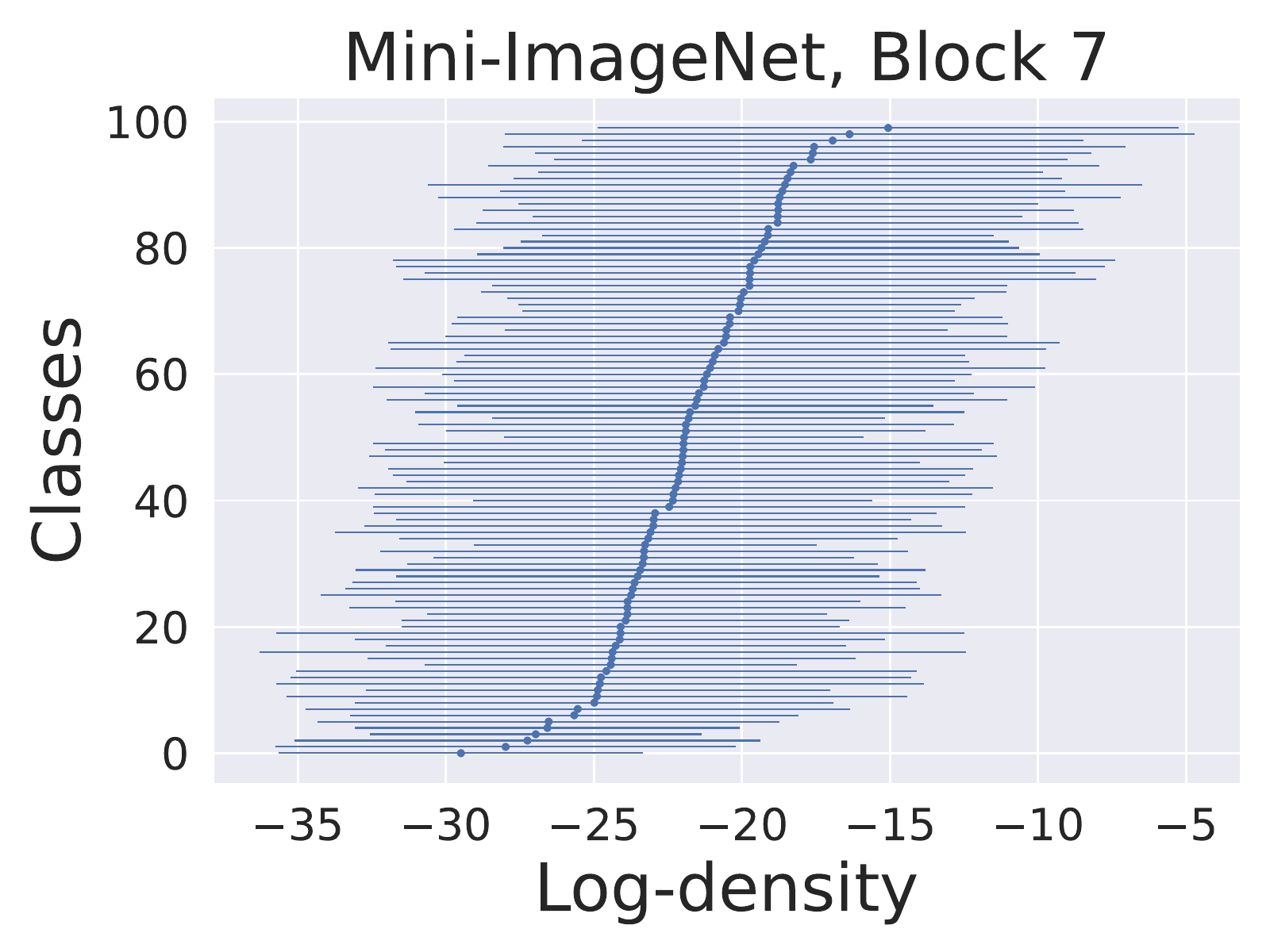}
  \includegraphics[width=0.244\linewidth]{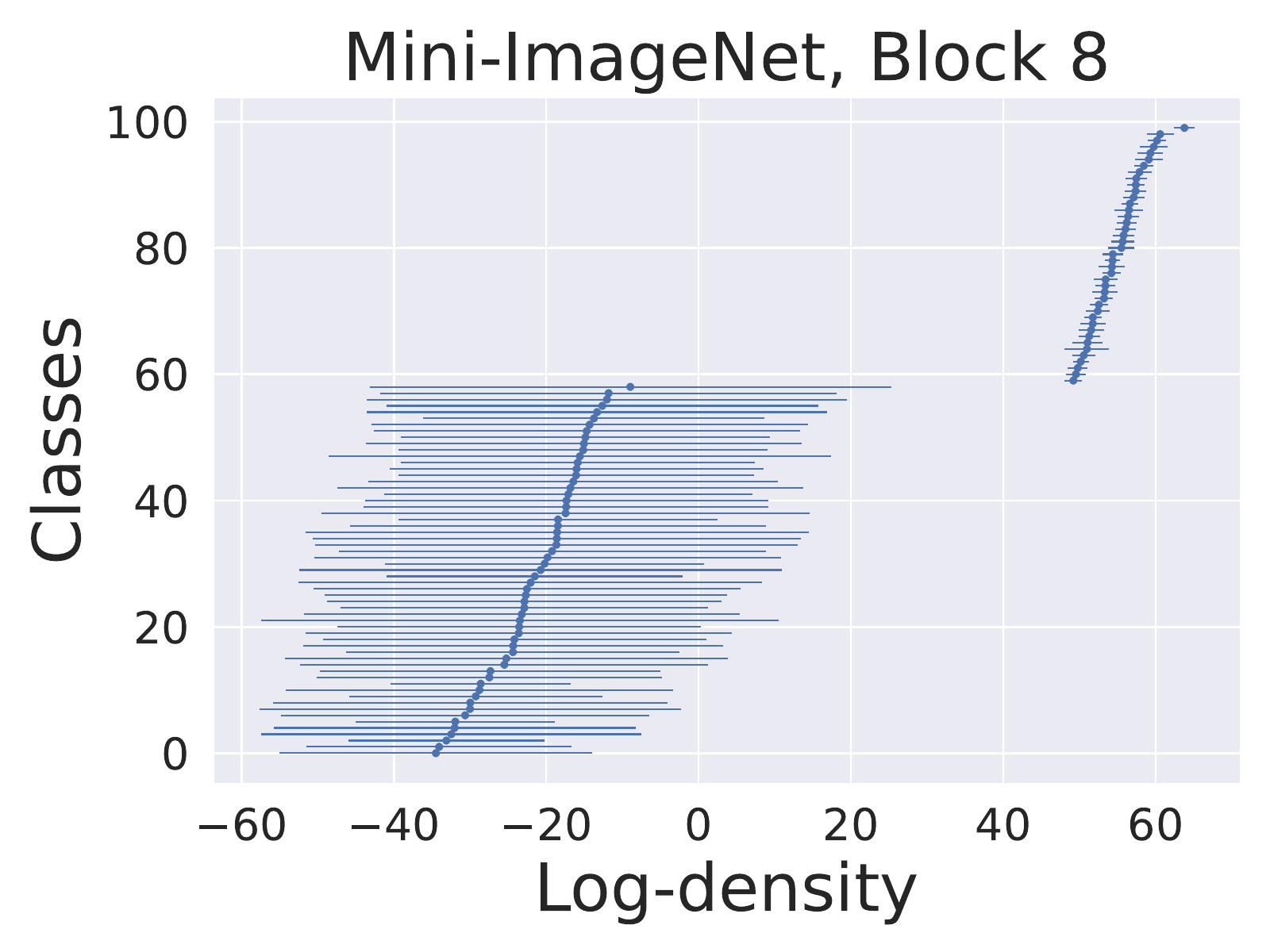}
  \\
  \includegraphics[width=0.244\linewidth]{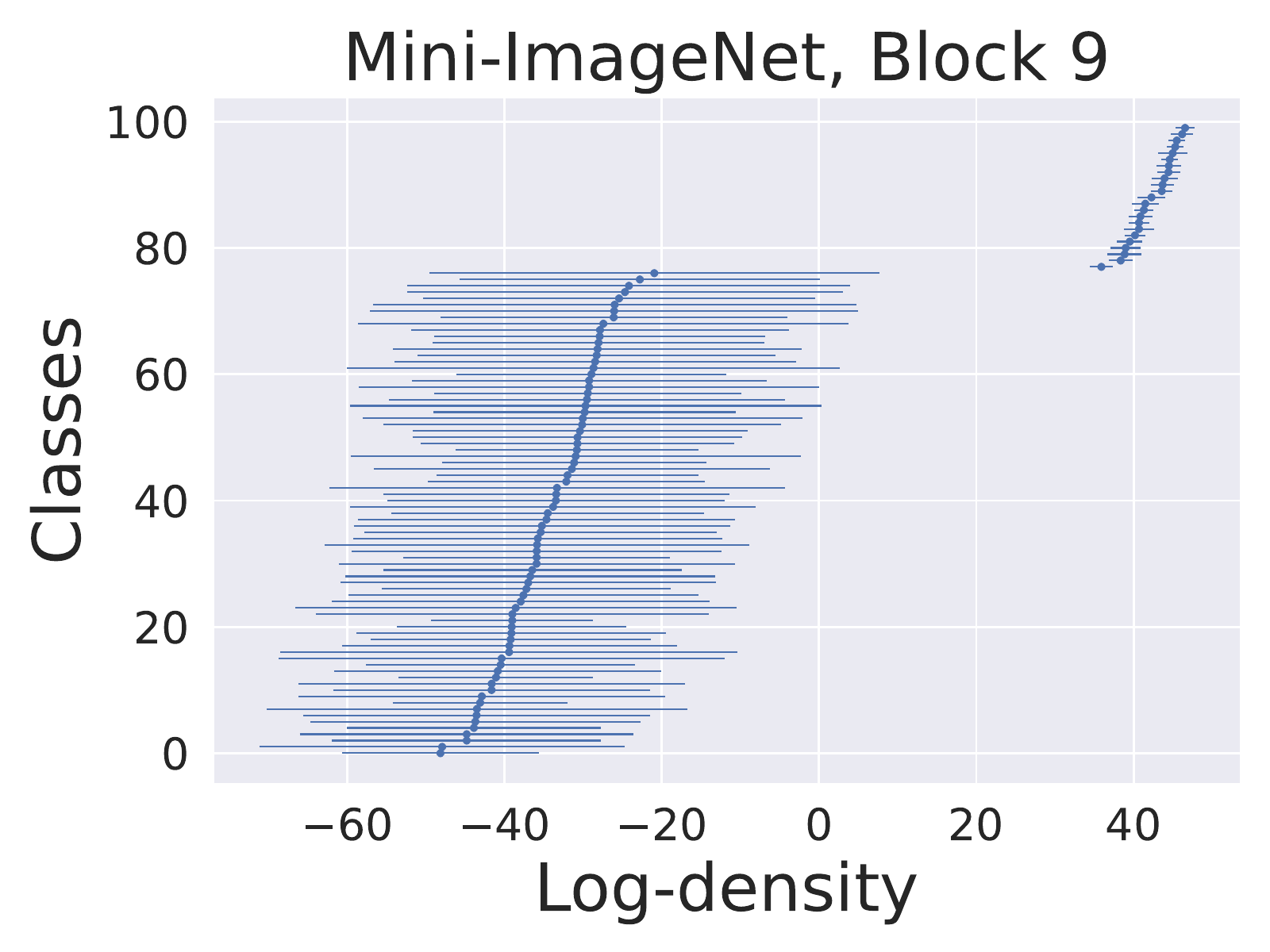}
  \includegraphics[width=0.244\linewidth]{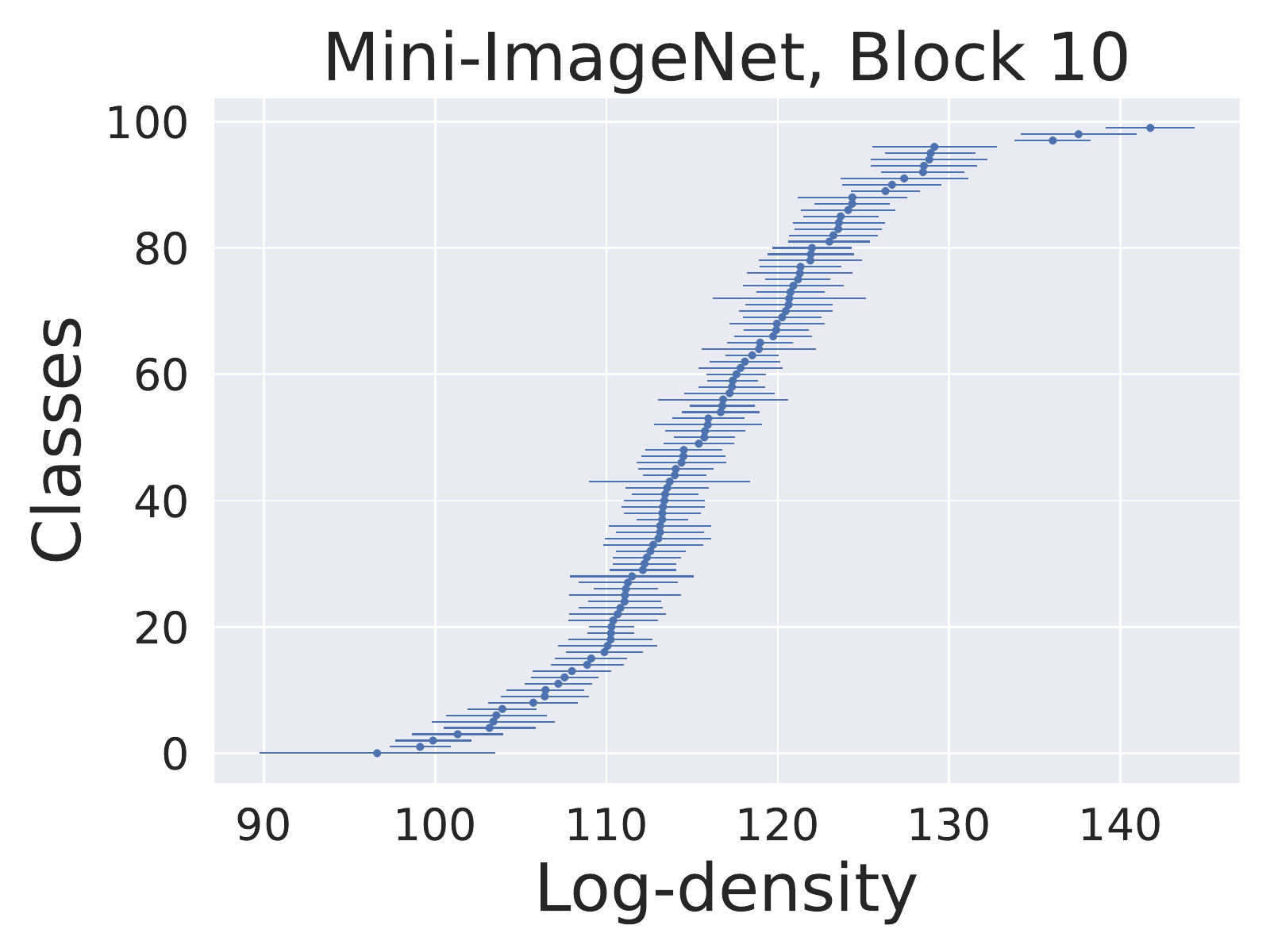}
  \includegraphics[width=0.244\linewidth]{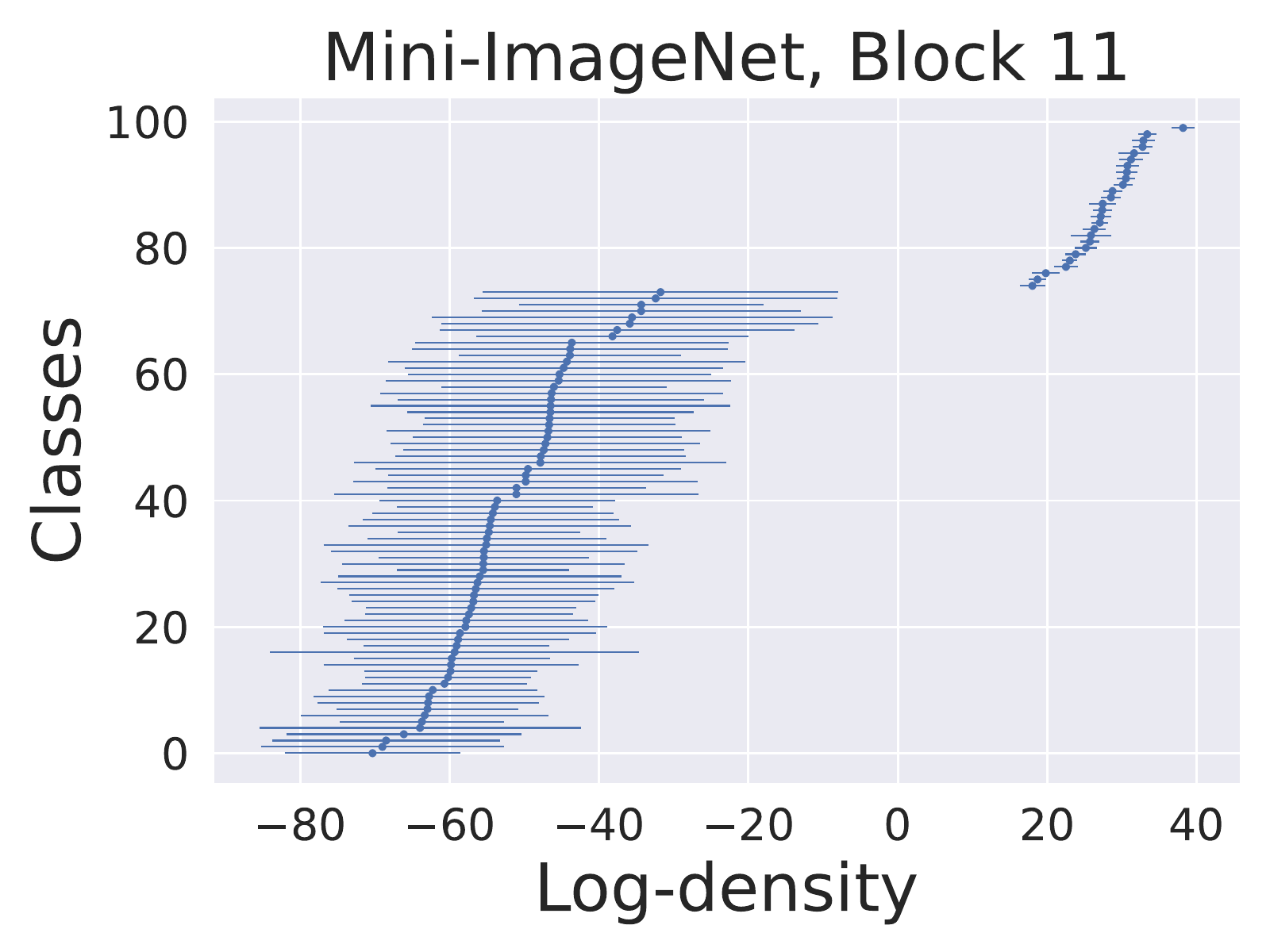}
  \includegraphics[width=0.244\linewidth]{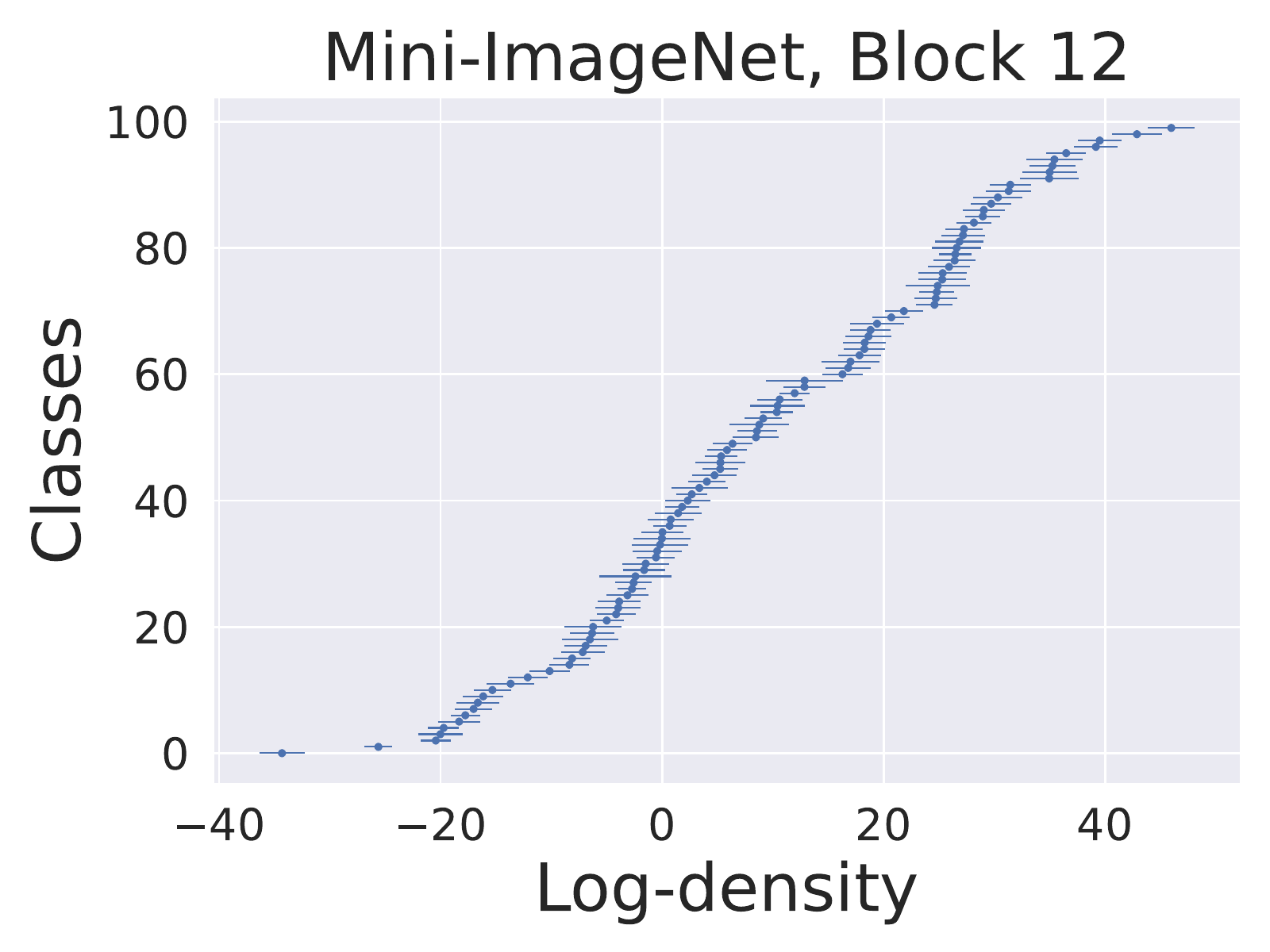}
  \caption{Mean and standard deviation of class-conditional log-densities estimated for representations from a 12-block
           MLP-Mixer model.}
           \label{fig:log_cc_densities_mixer}
\end{figure*}

Similarly to ResNets, initial layers in ViT and Mixer-MLP models do not display any unexpected structure in class
conditional distributions of neural representations (Fig.~\ref{fig:log_cc_densities_vit} and
Fig.~\ref{fig:log_cc_densities_mixer}). The picture changes in deeper layers. First, we observe layers where class
representations display structure similar to the one observed in ResNets, namely low- and high-density modes. However,
we also observe layers where all classes exhibit a relatively low (compared to initial layers) variance in class
conditional densities of neural representations. How this structure influences memorization and adversarial robustness
in these models remains an open question. Specifically, while replication the complete set of our experiments is
conceptually simple, the computational cost of training ViT and Mixer-MLP models makes calculation of required
memorization and adversarial robustness scores challenging.

\clearpage

%% file: appendix_class_divergence.tex
\begin{figure*}[!h]
  \centering
  \begin{tabular}{cc}
    ResNet18 & ResNet50 \\
    \includegraphics[width=0.45\textwidth]{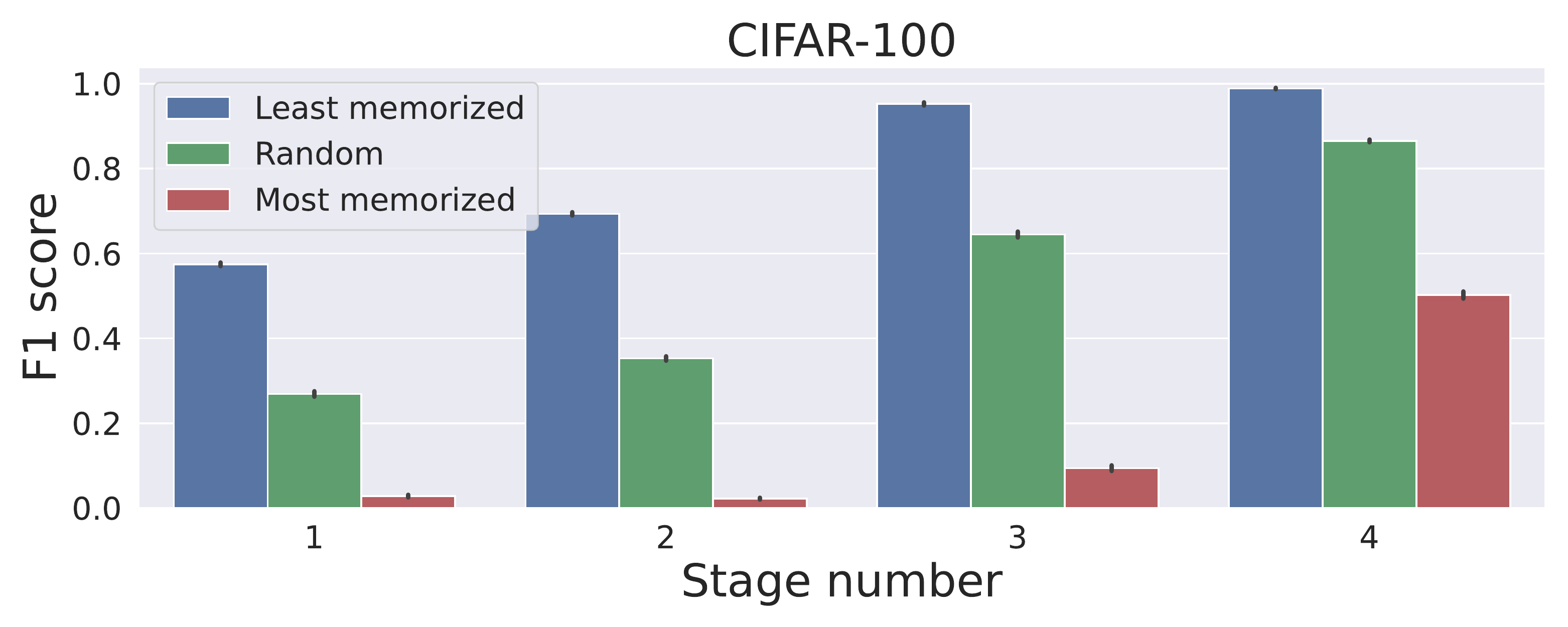} &
    \includegraphics[width=0.45\textwidth]{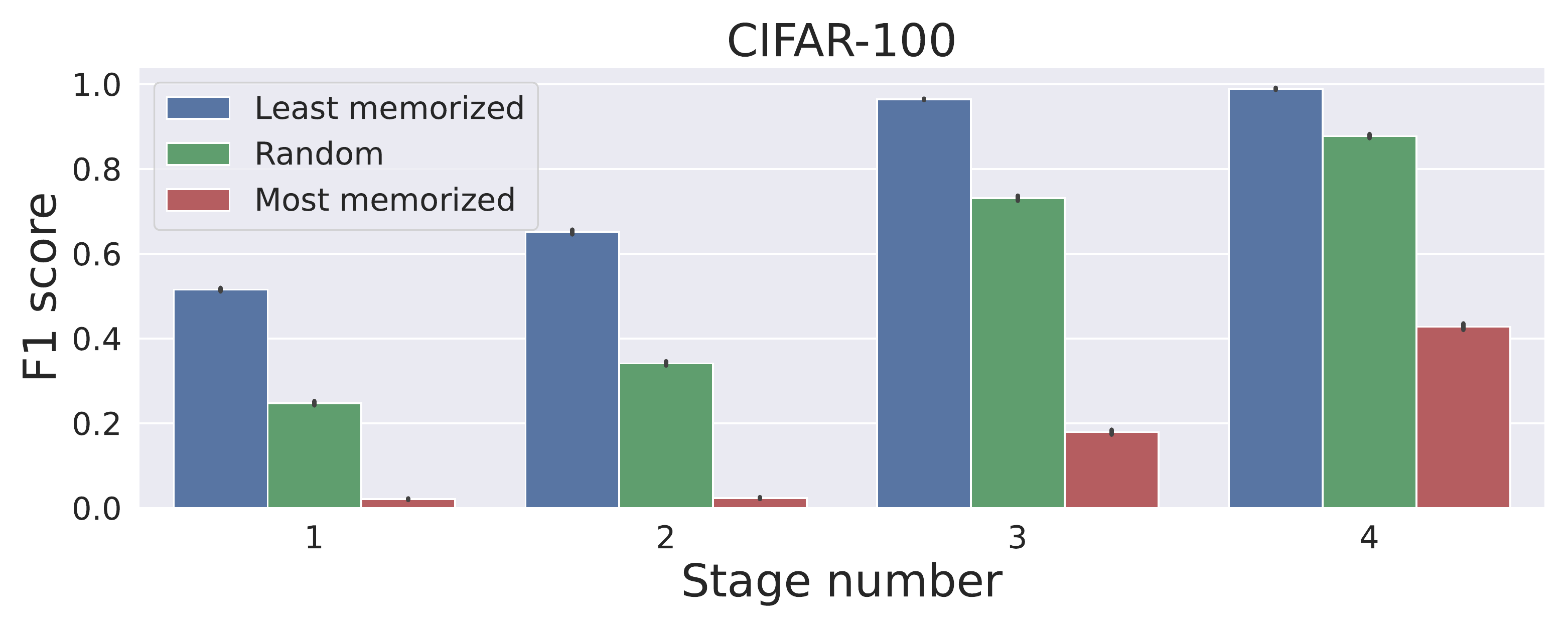}  \\
    \includegraphics[width=0.45\textwidth]{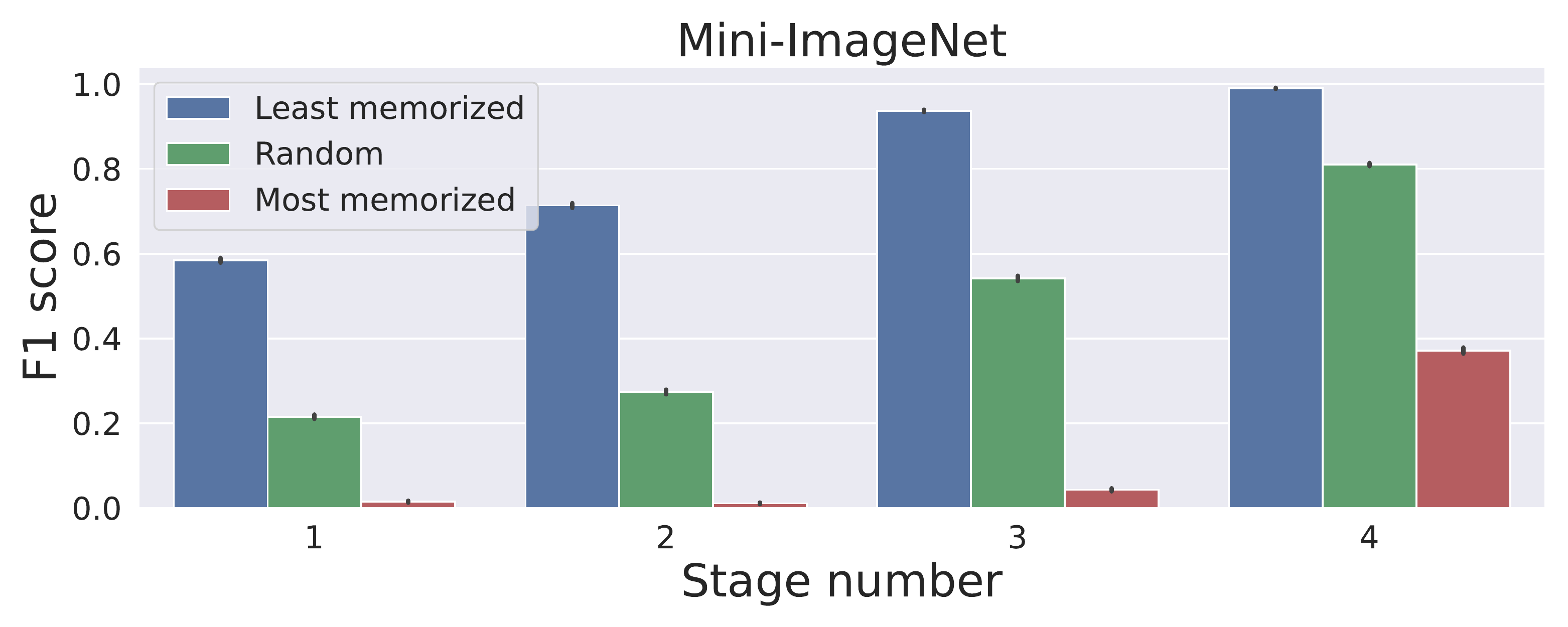} &
    \includegraphics[width=0.45\textwidth]{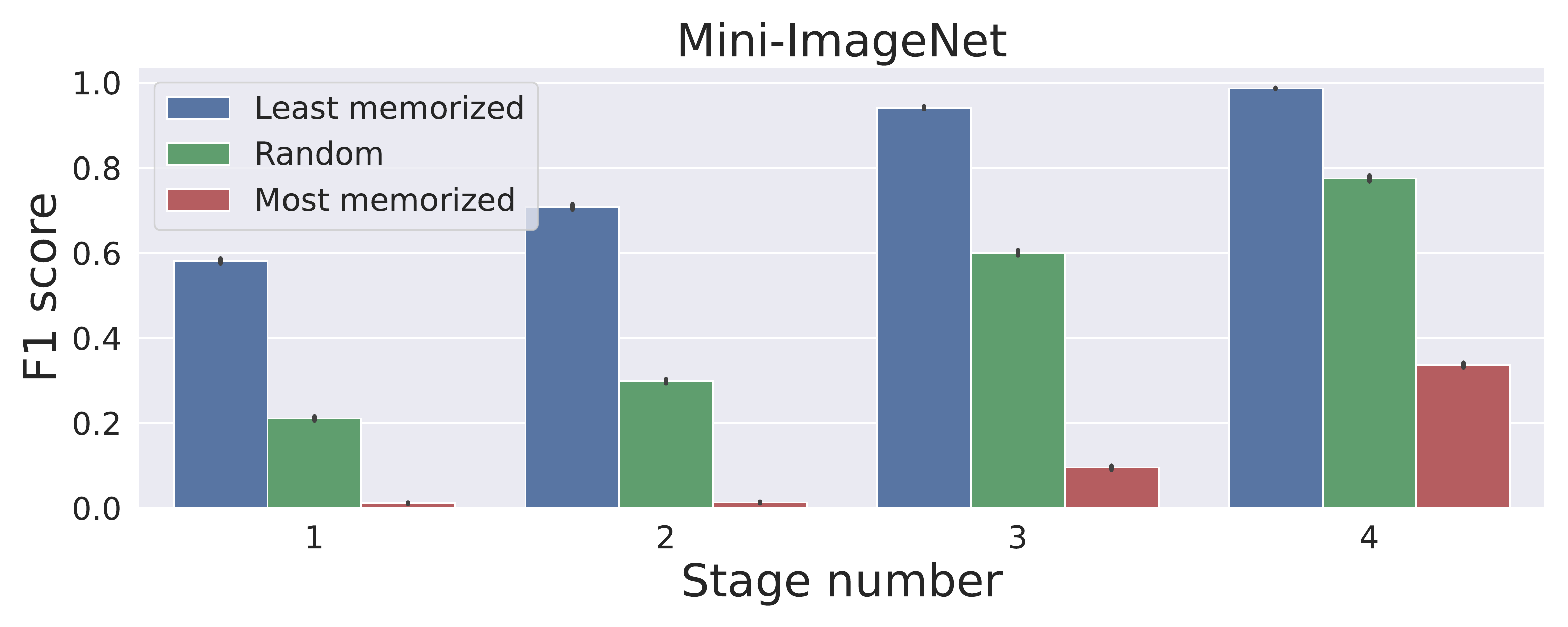}
  \end{tabular}
  \caption{F-scores for generative classification~(Eq.~\eqref{eq:generative_classifier}) of memorized examples, examples
           with no memorization, and randomly selected examples.}
  \label{fig:generative_classifier_f1}
\end{figure*}

Results reported in Fig.~\ref{fig:dkls_between_classes} can be replicated with a generative classifier derived from the
class-conditional densities:
\begin{equation} \label{eq:generative_classifier}
  C = \argmax_C p\left(nn_l\left( \mathbf{x} \right) \mid \mathbf{x} \in C \right) p(C),
\end{equation}
where the prior $p(C)$ comes from empirical class frequencies. More precisely, one can select a set of held-out
examples, fit class-conditional predictive densities to the remaining data points and predict classes for the held-out
examples. Figure~\ref{fig:generative_classifier_f1} reports F-scores for generative classification of $1,000$ most
memorized examples, an equal number of randomly selected examples, and an equal number of randomly selected examples
with no memorization (i.e. memorization estimate equal to $0.0$). Generative classification recover classes of held-out
inputs and its performance is compatible with the between-class divergences reported in
Fig.~\ref{fig:dkls_between_classes}.

\citet{Feldman2020b} reported negative result in an experiment designed to test whether estimation of memorization
scores can be speed-up by sharing a common convolutional backbone across all trained models, and only fitting the
ultimate fully-connected layer. Our results concerning fitting of memorized examples suggest that other strategies of
this kind are also likely to fail.